\documentclass{article} 
\usepackage{iclr2027_conference, times}

\usepackage{amsmath,amsfonts,bm}

\def\eqref#1{equation~\ref{#1}}

\def\1{\bm{1}}

\DeclareMathAlphabet{\mathsfit}{\encodingdefault}{\sfdefault}{m}{sl}
\SetMathAlphabet{\mathsfit}{bold}{\encodingdefault}{\sfdefault}{bx}{n}

\usepackage{amsthm}

\usepackage{pifont}
\usepackage{changepage}
\usepackage[table]{xcolor}
\usepackage{hyperref}
\usepackage{url}
\usepackage{comment}
\usepackage{microtype}
\usepackage{graphicx}
\usepackage{subcaption}
\usepackage{booktabs} 
\usepackage{amssymb}
\usepackage{hyperref}
\usepackage{wrapfig}
\usepackage{stfloats}       
\usepackage{multicol}
\usepackage{multirow}
\usepackage{comment}
\usepackage{enumitem}
\usepackage{bbm}
\usepackage{siunitx}
\usepackage{algorithm}
\usepackage{algpseudocode}
\usepackage{xcolor}
\usepackage[most]{tcolorbox}
\usepackage{mathtools}
\usepackage{longtable}
\usepackage{wrapfig}
\tcbuselibrary{breakable}
\algrenewcommand\algorithmicindent{1.2em}
\algrenewcommand\algorithmiccomment[1]{\hfill$\triangleright$ #1}

\definecolor{clearsky}{RGB}{100,150,255}
\definecolor{palegrey}{gray}{0.97}
\definecolor{darkergrey}{gray}{0.95}

\usepackage{titlesec}
\titlespacing*{\section}{0pt}{1.5ex plus 0.5ex minus 0.2ex}{0.8ex}
\titlespacing*{\subsection}{0pt}{1.2ex plus 0.5ex minus 0.2ex}{0.5ex}
\titlespacing*{\subsubsection}{0pt}{1ex plus 0.5ex minus 0.2ex}{0.3ex}
\tcbset{quietbox/.style={
  enhanced, breakable,
  colback=palegrey, colframe=darkergrey!95!black, boxrule=0.6pt, arc=1pt,
  left=6pt, right=6pt, top=4pt, bottom=4pt,
  before skip=8pt, after skip=8pt,
  fonttitle=\itshape, coltitle=black, 
  colbacktitle=darkergrey, bottomtitle=2pt, toptitle=2pt,
  description delimiters={}{},   
  separator sign={.},       
}}
\newtcbtheorem{definition}{Definition}{quietbox}{def}
\newtcbtheorem{proposition}{Proposition}{quietbox}{prop}
\newtcbtheorem{boxproof}{Proof}{quietbox}{boxproof}

\newtcbtheorem{desideratum}{Desideratum}{
  enhanced, breakable,
  colback=clearsky!8!white, colframe=clearsky!95!black, boxrule=0.6pt, arc=1pt,
  left=6pt, right=6pt, top=4pt, bottom=4pt,
  before skip=8pt, after skip=8pt,
  fonttitle=\bfseries, coltitle=black,
  colbacktitle=clearsky!20!white, bottomtitle=2pt, toptitle=2pt,
  description delimiters={}{},   
  separator sign={.},           
}{des}
      
\title{Q-Shaped Options \\ for Hierarchical Reinforcement Learning}

\author{Clarisse Wibault \\
\textbf{Antoine Gorceix \quad
Antonio Léon Villares \quad
Alexey Zakharov \quad
Evangelos Chatzaroulas} \\
\textbf{Michael Matthews \quad
Eduardo Pignatelli \quad
Jakob Foerster}
\\
BOLD \\
University of Oxford \\
{\footnotesize\texttt{clarisse.wibault@magd.ox.ac.uk}}
}

\iclrfinalcopy 
\begin{document}

\maketitle

\begin{abstract}
Learning to tackle long-horizon, goal-conditioned tasks requires an agent to reason over extended timescales and act across a broad range of states. 
In principle, Hierarchical Reinforcement Learning (HRL) addresses both challenges through the interaction between action (temporal) and state (spatial) abstraction. 
First, using an action abstraction to represent temporally extended behaviour as options reduces the effective decision horizon. 
Second, enabling different state abstractions at each level of the decision process permits greater data aggregation for learning. 
However, realising these two benefits of a hierarchical policy depends on learning an appropriate action abstraction. Current HRL algorithms fail in one of two ways. Some discard distinctions between options needed for optimal control, undermining hierarchy altogether. Others retain unnecessary distinctions, preserving horizon reduction, but forfeiting coarser state abstraction. 
In this work, we characterise three desiderata for an action abstraction. We introduce Q-Shaped Options (QSO) to address all three. QSO builds on an architecture with distinct state-value functions, Q functions and policies at each level of the hierarchy. It learns the action abstraction between consecutive levels as a shared encoder shaped by their respective Q functions. The low-level Q function uses the option as a goal, encouraging the abstraction to retain distinctions necessary for optimal control. The high-level Q function uses it as an action, encouraging unnecessary distinctions to be discarded. Across offline goal-conditioned locomotion and manipulation environments, QSO learns semantically meaningful option spaces and outperforms baselines, achieving non-zero performance in tasks where all other evaluated algorithms fail. Our code is open sourced at \url{https://github.com/CWibault/QSO}. 
\end{abstract}

\section{Introduction}
\label{sec: introduction}
\begin{figure}
    \centering
    \includegraphics[width=1.0\linewidth]{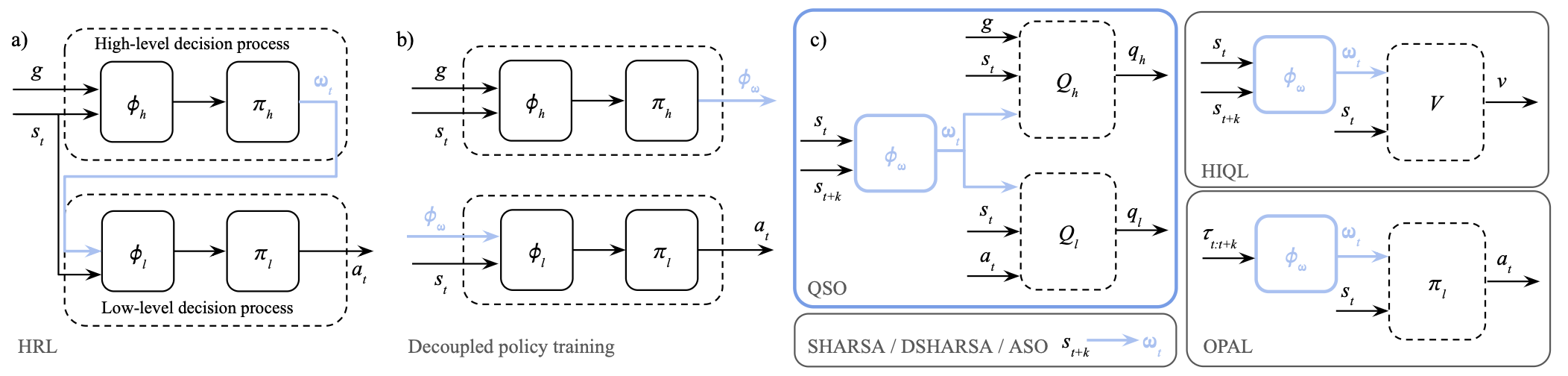}
    \vspace{-1.5em}
    \caption{(\textbf{a}) \textbf{HRL} permits different state abstractions $\phi_h(s_t, g)$ and $\phi_l(s_t, \omega_t)$ at each level of the decision process. (\textbf{b}) \textbf{Decoupled policy training}. (\textbf{c}) Different methods for learning the action abstraction $\phi_\omega(\tau_{t:t+k})$.}
    \label{fig: main}
\end{figure}

Learning to tackle long-horizon, goal-conditioned \citep{kaelbling1993learning, schaul2015universal} tasks requires an agent to reason over extended timescales and act across a broad range of states \citep{sutton1999between, park2026horizon}. In goal-conditioned two-tiered Hierarchical Reinforcement Learning (HRL) a high-level policy outputs a high-level action or option \citep{sutton1999between} conditioned on the current state and goal; a low-level policy then conditions on this option as well as the current state to output a primitive action (Figure \ref{fig: main}a). In maze navigation, for example, the option might be to reach ``a point two metres down'', while the primitive action could be applying torques to an individual joint. 

In principle, HRL \citep{barto2003recent} addresses both challenges of long-horizon, goal-conditioned tasks through the interaction between action (temporal) and state (spatial) abstraction \citep{konidaris2019necessity}. 
First, by using an action abstraction to represent temporally extended behaviour as options, a hierarchical policy reduces the effective decision horizon \citep{park2026horizon}. The high-level policy reasons over options instead of primitive actions, while the low-level policy reasons to pursue the option rather than the distant goal. 
Second, by enabling different state abstractions at each level of the hierarchy (Figure \ref{fig: main}a), a hierarchical policy can achieve coarser state abstraction than a non-hierarchical policy if each level discards different information irrelevant to its part of the decision process \citep{konidaris2019necessity, konidaris2018skills}. 
Intuitively, this permits greater data aggregation for learning. For example, the high-level can aggregate all examples at a certain coordinate irrespective of pose, which is too fine-grained for high-level control, while the low-level can aggregate all examples with a certain pose irrespective of coordinates, since reaching ``a point two metres down'' requires the same actions wherever the agent is.

However, realising these two benefits of a hierarchical policy depends on learning an appropriate action abstraction. 
To improve stability and avoid collapsing to degenerate solutions \citep{bacon2017option}, decoupled approaches \citep{ajay2020opal, park2023hiql, park2026horizon} train each policy separately (Figure \ref{fig: main}b) using fixed-length trajectory segments. 
The action abstraction maps these segments to the option space, serving as actions for the high-level policy and goals for the low-level policy during training. The issue is that current HRL algorithms learn abstractions that fail in one of two ways. 
Some discard distinctions between options that are needed for optimal control \citep{park2023hiql, li2006towards}. 
This limits the quality of the recoverable policy and undermines hierarchy altogether. Others retain distinctions between options that are irrelevant for control \citep{park2026horizon, ajay2020opal, ravindran2002model}. 
This preserves horizon reduction but forfeits coarser state abstraction, as functionally equivalent behaviours remain distinct outputs of the high-level policy and distinct inputs to the low-level one. 

In this work, we characterise three desiderata for an action abstraction to enable horizon reduction and data aggregation. We introduce Q-Shaped Options (QSO) to address all three. 
QSO builds on an architecture \citep{park2026horizon} with separate state-value functions, Q functions and policies at each level of the hierarchy. It learns the action abstraction between consecutive levels as an encoder shared between the two levels and shaped by their respective Q functions. To our knowledge, QSO is the first HRL algorithm to learn an action abstraction through a Q function. 
We encode only the current and target states to represent the outcome of a trajectory segment rather than the exact state-action sequence that produced it.  
We shape the abstraction via the low-level Q function to separate options with different low-level Q values \citep{li2006towards}, retaining distinctions necessary for optimal control.
Finally, we shape the abstraction via the high-level Q function to provide an inductive bias for discarding unnecessary distinctions between functionally equivalent behaviour. Across offline \citep{levine2020offline, lange2012batch} goal-conditioned locomotion and manipulation environments, QSO learns semantically meaningful option spaces (Figures \ref{fig: ant_option_embeddings}, \ref{fig: cube_double_option_embeddings} and \ref{fig: cube_quadruple_option_embeddings}) and outperforms baselines (Table \ref{tab: results_main}), achieving non-zero performance in tasks where all other evaluated algorithms fail (Table \ref{tab: results_appendix_manipulation}). We summarise our contributions as follows.
\begin{enumerate}[nosep, leftmargin=*]
    \item We provide a formal definition of a control-sufficient action abstraction and prove that exact low-level Q-shaping satisfies control sufficiency. 
    \item We extend a Probably Approximately Correct bound on offline error in a finite-state and finite-action MDP to motivate discarding unnecessary distinctions between options.
    \item We develop Q-Shaped Options (QSO), to our knowledge the first HRL algorithm to learn the action abstraction through a Q function, yielding a control-sufficient action abstraction that also discards unnecessary distinctions between options. 
\end{enumerate}

\begin{table}
\centering
\small
\begin{tabular}{lllcc}
\toprule
\textbf{Algorithm} & \textbf{Encodes} & \textbf{Method} &
\textbf{Preserves} &
\textbf{Discards} \\
\midrule
HIQL & $s_t, s_{t+k}$ & $V$ & No & Yes \\
OPAL & $\tau_{t:t+k}$ & $\beta$-VAE (Surrogate Objective) & Yes & No \\
SHARSA & $s_{t+k}$ & -  & Yes & No \\
DSHARSA & $s_{t+k}$ & - & Yes & No \\
\midrule
ASO & $s_{t+k}$ & - & Yes & No \\
QSO  & $s_t, s_{t+k}$ & $Q_h$ and $Q_l$ & Yes & Yes \\
\bottomrule
\end{tabular}
\caption{\textbf{Properties of the action abstraction $\phi_\omega$}: what the abstraction \textbf{encodes}; the \textbf{method} used to learn the abstraction; whether the objective encourages preserving distinctions \textbf{necessary for optimal control} and whether the abstraction empirically exhibits \textbf{discarding unnecessary distinctions}.}
\label{tab: option_embedding_properties}
\end{table} 

\section{Related Works}
\label{sec: related_works}
As summarised in Table \ref{tab: option_embedding_properties} and shown in Figure \ref{fig: main}c, prior algorithms using decoupled training approaches differ substantially both in what the action abstraction encodes, ranging from a single target state $s_{t+k}$ to the full $k$-step state-action trajectory segment $\tau_{t:t+k}$, and the method used to learn the action abstraction. 

HIQL \citep{park2023hiql} learns the action abstraction through its single action-free state-value function $V(s_t, \phi_\omega(s_t, g))$. However, value equivalence does not imply action equivalence \citep{li2006towards}, as two state-goal pairs can share a value, and hence option, while requiring entirely different immediate actions (Figure \ref{fig: ant_option_embeddings}a). This means that the abstraction can discard information necessary for optimal control. 

SHARSA and DSHARSA \citep{park2026horizon} simply bypass learning an action abstraction by using subgoal options $\omega_t = s_{t+k}$ \citep{konidaris2019necessity}, which specify the entire state that the low-level policy should reach. The ``encoder'' is just an identity mapping. While these ``options'' retain the information required for low-level control, they distinguish functionally equivalent behaviour occurring at different points in the state space (Figure \ref{fig: ant_option_embeddings}c), limiting coarser state abstraction and hence data aggregation for learning. 

OPAL \citep{ajay2020opal} uses a surrogate objective to learn an option space $\omega_t \sim \phi_\omega(\omega_t \mid \tau_{t:t+k})$ with a $\beta$-VAE trained to reconstruct the entire action sequence. This means that the option encodes how a segment was executed, instead of its outcome. Similarly to SHARSA and DSHARSA, such an encoding hinders coarser state abstraction as two trajectory segments with the same outcome may be represented differently just because their executions differ (Figure \ref{fig: ant_option_embeddings}b).

QC \citep{li2026reinforcement}, which builds on action chunking \citep{zhao2023learning}, is closely related to our work as it also provides temporal abstraction, but rather than having a multi-level structure, a non-hierarchical policy outputs an action chunk $\pi(a_{t:t+k}\mid s_t, g)$. Because a single level is responsible for the entire decision process, this prevents coarser state abstraction. 

\section{Preliminaries}
\label{sec: preliminaries}

\paragraph{Problem Setting} We consider a goal-conditioned Markov Decision Process (MDP) \citep{sutton1998reinforcement, schaul2015universal} defined by the tuple $\mathcal{M} :=(p_{s_0}, p_{g}, \mathcal{S}, \mathcal{G}, \mathcal{A},  \mathcal{T}, r, \bar{\gamma})$, where $p_{s_0}$ is the initial state distribution, $p_{g}$ is the goal distribution, $\mathcal{S}$, $\mathcal{G}$ and $\mathcal{A}$ respectively denote the state, goal and action spaces with $\mathcal{G} = \mathcal{S}$, and $\mathcal{T}$ is the transition function. 
$r(s, g) := -(1 - \beta(s, g))$ and $\bar{\gamma}(s, g) := \gamma\cdot(1 - \beta(s, g))$, where $\gamma \in [0,1)$ is a constant, are the reward and discount or continuation functions \citep{andrychowicz2017hindsight, sutton2011horde, white2012scaling} defined through a goal-achievement indicator $\beta : \mathcal{S} \times \mathcal{G} \to \{0, 1\}$ (specified per environment in Appendix \ref{app: experiments}): the agent receives $-1$ per step until the goal is reached, after which the return terminates. 
At the beginning of the episode, a state $s_0$ is sampled from $p_{s_0}$, and a goal $g$ is sampled from $p_g$. The goal is fixed for the entire episode. At each timestep $t \ge 0$, an agent takes an action $a_t$ conditioned on its current state $s_t$ and goal state $g$, and deterministically transitions to a new state $s_{t+1} = \mathcal{T}(s_t, a_t)$;  we adopt the fully observed deterministic MDP framework as used in \citet{park2025ogbench} for simplicity. Our aim is to learn a policy $\pi: \mathcal{S} \times \mathcal{G} \to \Delta(\mathcal{A})$, where $\Delta(\mathcal{A})$ denotes the space of probability distributions over $\mathcal{A}$, from a fixed dataset $\mathcal{D}$ of $N$ state-action trajectories of length $H$, collected by an arbitrary policy, that maximises $\mathbb{E}_\pi[\sum_{t=0}^{\infty}(\prod_{i=0}^{t-1} \bar{\gamma}(s_i, g))\, r(s_{t}, g)]$. The return is bounded in $[-1/(1-\gamma), 0]$ and is maximised by minimising the discounted time to reach the goal. In what follows we write the constant $\gamma$ in place of $\bar{\gamma}(s, g)$ and omit goal-termination masking for clarity.

\paragraph{Hierarchical Reinforcement Learning}
In two-tiered goal-conditioned HRL \citep{barto2003recent, park2023hiql} the aim is to learn a high-level policy $\pi_h(\omega_t \mid s_t, g)$ that outputs an option $\omega_t \in \Omega$, where  $\Omega$ denotes the option space, conditioned on the current state $s_t \in \mathcal{S}$ and goal $g \in \mathcal{G}$; a low-level policy $\pi_l(a_t \mid s_t, \omega_t)$ then conditions on this option as well as the current state to output a low-level action $a_t \in \mathcal{A}$ (Figure \ref{fig: main}a), such that the overall policy is defined by the following:
\begin{equation}
    \pi(a_t \mid s_t, g) := \int_{\omega_t \in \Omega}\pi_l(a_t \mid s_t, \omega_t)\pi_h(\omega_t \mid s_t, g) \mathrm{d}\omega_t.
    \notag
\end{equation}

While options may have fixed or variable temporal extent, in this work we focus on fixed-length options, which do not collapse to degenerate solutions \citep{bacon2017option} and simplify the separation of the two decision processes with a fixed temporal and discount relationship. During training, an action abstraction $\phi_\omega$ is learned and used to encode (a subset of) the $k$-step state-action trajectory segment $\tau_{t:t+k}=(s_t, a_t, \cdots, s_{t+k})$ into an option $\omega_t=\phi_\omega(\tau_{t:t+k})$. The high-level policy is trained to output this option, while the low-level policy is trained to condition on this option. At deployment, the action abstraction is no longer required, since the two levels communicate entirely within option space. 

\section{Q-Shaped Options}
\label{sec: method}
Our aim is to learn an action abstraction that represents the outcome of a trajectory segment rather than the exact state-action sequence that produced it, retains distinctions between options necessary for optimal control, and discards unnecessary distinctions. 

\subsection{Desiderata}

\definecolor{clearsky}{RGB}{100,150,255}

\textbf{The action abstraction should represent the outcome of a trajectory segment rather than the exact state-action sequence that produced it.}

We want to prevent potentially suboptimal dataset execution from becoming a part of the option's representation. To enable two different executions with the same outcome to map to the same option, we restrict the abstraction to be a function of the state-achieved-state pair $(s_t, s_{t+k})$, rather than the entire state-action trajectory segment. We include the current state as well as the achieved state to enable representing relative rather than absolute outcome (see Desideratum 3). HIQL also encodes $(s_t, s_{t+k})$, but motivates this choice empirically. 

\begin{desideratum}{Represent outcome rather than exact execution.}{desideratum_1}
\textbf{QSO:} Action abstraction encodes $(s_{t}, s_{t+k})$ instead of $\tau_{t:t+k}$. 
\end{desideratum}

\textbf{The action abstraction should be control-sufficient i.e. preserve distinctions between options necessary for optimal control.}

We want two target states that require different actions from the same current state map to different options, so that a policy conditioning on the representation instead of the target state can still be optimal for both. To address control sufficiency, we shape the abstraction through the low-level Q function. If two admissible target states require different actions from the same current state, the low-level Q function must represent different sets of action values. Minimising the Bellman error therefore pressures the action abstraction to map the current and target state pairs to different points in the option space. Conversely, if two trajectory segments have functionally equivalent relative outcomes but different dataset executions, they can share an option if their sets of action values are compatible, even if one was executed well and the other poorly. This can enable the low-level policy to draw on the better execution at deployment. 
Definition \ref{def:control_sufficiency} formalises this notion of a control-sufficient action abstraction, while Proposition \ref{prop:proposition_1} connects this property to the low-level Q shaping objective. We refer to Appendix \ref{app: control_sufficiency} for the proof. 

\begin{definition}{Control-sufficient action abstraction.}{control_sufficiency}
Let $\mathcal{G}_k(s) \subseteq \mathcal{S}$ denote the set of admissible target states for the $k$-step low-level decision problem from state $s$, and let $\Pi^\star(s, g_s) := \text{argmax}_a Q^\star(s, g_s, a)$ denote the set of optimal actions of the original MDP. An action abstraction $\phi_\omega$ is control-sufficient if, for all $s \in \mathcal{S}$ and $g_s, g_s' \in \mathcal{G}_k(s)$,
\begin{equation}
    \phi_\omega(s, g_s) = \phi_\omega(s, g_s')
    \quad \Longrightarrow \quad
    \Pi^\star(s, g_s) = \Pi^\star(s, g_s').
    \notag
\end{equation}
In particular, there exists a low-level policy $\pi_l$ such that $\pi_l(\cdot \mid s, \phi_\omega(s, g_s))$ is supported on $\Pi^\star(s, g_s)$ for every admissible pair $(s, g_s)$.
\end{definition}

Note that Q-shaping is stronger than control sufficiency, as it distinguishes between options whose action values differ only by an additive offset or in the ranking of suboptimal actions, even if these differences do not affect the optimal action. 
\begin{proposition}{Low-level Q-shaping implies control sufficiency.}{proposition_1}
    Suppose there exists a low-level critic $Q_l$ such that $Q_l\big(s, \phi_\omega(s, g_s), a\big) = Q^\star(s, g_s, a)$ for all $s \in \mathcal{S}$, $g_s \in \mathcal{G}_k(s)$ and $a \in \mathcal{A}$, i.e.\ $\phi_\omega$ is a $Q^\star$-irrelevance abstraction in the sense of \citet{li2006towards}. Then $\phi_\omega$ is control-sufficient.
\end{proposition}

\begin{desideratum}{Preserve distinctions between options necessary for optimal control.}{desideratum_2}
\textbf{QSO:} Learn action abstraction through low-level Q function: $Q_l(s_t, \phi_\omega(s_t, g_s), a_t)$.  
\end{desideratum}

\textbf{The action abstraction should discard unnecessary distinctions between options.}

We want trajectory segments with similar outcomes to map to the same option, so that experience can be aggregated across the state space. This enables conservative extrapolation: aggregating different examples across different levels of the hierarchy to produce an action not observed for a particular state-goal pair. Compressing a finite option space or organising a continuous option space in a semantically meaningful way reduces the effective complexity of the hierarchical decision problem and can reduce the distribution mismatch between the optimal policy and the dataset by aggregating occupancy mass across equivalent state-goal and state-option pairs. We refer to Appendix \ref{app: conservative_extrapolation} for an example, and Appendix \ref{app: compression_bounds} for a proposition that formalises this intuition by deriving and analysing the Probably Approximately Correct bound on offline error for a hierarchical policy with joint state and action abstraction.
 
To encourage discarding unnecessary distinctions between options, we additionally shape the option space through the high-level Q function. Functionally equivalent state-target pairs induce similar high-level action values, so distinguishing between them provides little predictive benefit to $Q_h$. Note that, unlike control sufficiency, this invariance is not explicitly enforced by the objective, instead simply providing an inductive bias towards it under shared function approximation and gradient-based optimisation. Empirically, we find that this is sufficient to form coherent clusters in the option space. Discarding unnecessary distinctions between options also motivates conditioning the action abstraction on the current state, as then the abstraction only needs to separate target states if they require different actions from the same current state rather than if they require different actions from any admissible current state. 

\begin{desideratum}{Discard unnecessary distinctions between options.}{desideratum_3}
\textbf{QSO:} Learn action abstraction $\phi_\omega$ through the high-level Q function: $Q_h(s_t, g, \phi_\omega(s_t, s_{t+k}))$.  
\end{desideratum}

\subsection{Practical Implementation}
\textbf{Joint Action Abstraction and Value Learning}\quad We learn the action abstraction between consecutive levels of the hierarchy as an encoder shared between the two levels and jointly shaped by their respective Q functions:
\begin{equation}
    \underset{Q_l, Q_h, \textcolor{clearsky!95!black}{\phi_\omega}}{\text{min}} \mathcal{L}_{Q_l, \textcolor{clearsky!95!black}{\phi_\omega}} + \mathcal{L}_{Q_h, \textcolor{clearsky!95!black}{\phi_\omega}}
    \notag
\end{equation}
where the critic losses are the following, conservative Implicit Q Learning \citep{kostrikov2021offline} critic losses:
\begin{equation}
    \begin{aligned}
    \mathcal{L}_{Q_l, \textcolor{clearsky!95!black}{\phi_\omega}} &= \mathbb{E}_{\substack{(s_t,a_t, s_{t+1},g_s) \sim \mathcal{D}\\ \omega_{t+1} = \phi_\omega(s_{t+1}, g_s)}} \Bigg[\Big(Q_l \big(s_t, \textcolor{clearsky!95!black}{\phi_\omega(s_t,g_s)},a_t\big) - r_l(s_t, g_s) - \gamma V_l(s_{t+1}, \omega_{t+1})\Big)^2\Bigg],\\
    \mathcal{L}_{Q_h, \textcolor{clearsky!95!black}{\phi_\omega}} &= \mathbb{E}_{(s_t, s_{t+k}, g) \sim \mathcal{D}}
        \Bigg[\Big( Q_h\big(s_t, g, \textcolor{clearsky!95!black}{\phi_\omega(s_t, s_{t+k})}\big) - R_h(s_{t:t+k}, g) - \gamma V_h(s_{t+k}, g) \Big)^2\text{ }\Bigg],
    \notag
    \end{aligned}
\end{equation}
and the normalised $k$-step option return is $R_h(s_{t:t+k}, g) = \frac{1 - \gamma}{1 - \gamma^{k}} \sum_{t'=0}^{k-1} \gamma^{t'}\, r_h(s_{t+t'}, g)$.  

We use a continuous option space, parameterising the action abstraction with an MLP followed by length normalisation \citep{park2023hiql} onto a hypersphere. Unlike a discrete option space, this avoids imposing arbitrary boundaries between semantically similar options, while the length normalisation keeps proposed options in-distribution for the low-level policy. At deployment, the high-level policy resamples an option at every timestep, allowing the agent's intent to slowly shift across the option space. We use the same discount factor at both levels so that each level looks $1 / (1 - \gamma)$ of its own decisions ahead, and normalise the $k$-step option return so that a full-length option has the same cost as a single primitive action. This keeps the two Q functions on comparable scales and hence avoids a hyperparameter to balance their losses. Ideally, $k = 1 / (1 - \gamma)$ so that one option spans exactly the horizon that the low-level reasons over. In practice, $k$ is bounded by the data \citep{park2023hiql, park2026horizon, li2026reinforcement}. 

\begin{algorithm}
{\footnotesize
\caption{Q-Shaped Options (QSO) Training}
\label{alg: qso_training}
\begin{algorithmic}
    {
    \footnotesize
    \State \textbf{Initialise} option representation $\phi_\omega(s_t, g_s)$, high-level value $V_h(s_t, g)$, critic $Q_h(s_t, g, \omega)$ and flow policy $\pi_h(\omega \mid s, g)$, low-level value $V_l(s_t, \omega)$, critic $Q_l(s_t, \omega, a_t)$ and flow policy $\pi_l(a_t \mid s, \omega)$ and target option and critic networks $\tilde\phi_\omega, \tilde{Q}_h, \tilde{Q}_l$.
    \Repeat
        \State $\triangleright$ \textcolor{clearsky!95!black}{Dataset Sampling and Reward Labelling.}
        \State \quad Sample states from $\mathcal{D}$. Determine $a_t, s_{t+1}, s_{t+k}$. 
        \State \quad Sample high-level goals $g$ and low-level targets $g_s$ using Equations \ref{eq: high_level_goals} and \ref{eq: low_level_targets}. 
        \State \quad Determine rewards $r_l(s_{t}, g_s)$ and $r_h(s_{t:t+k},g)$. 
        \State $\triangleright$ \textcolor{clearsky!95!black}{Joint Action Abstraction and Value Learning.}
        \State \quad Encode high-level options $\omega \gets \phi_\omega(s_t, s_{t+k})$ and target options $\tilde\omega \gets\tilde\phi_\omega(s_t, s_{t+k})$.
        \State \quad Encode low-level options $\phi_\omega(s_t, g_s)$ and target options $\tilde\phi_\omega(s_t, g_s)$.
        \State \quad Train $V_h$ and $V_l$ using Equations \ref{eq: high_value} and \ref{eq: low_value}. 
        \State \quad Train $Q_h$ and $Q_l$ and $\phi_\omega$ using Equations \ref{eq: high_q_function} and \ref{eq: low_q_function}.
        \State \quad Update $\tilde\phi_\omega, \tilde{Q}_h, \tilde{Q}_l$ using exponential moving averages. 
        \State $\triangleright$ \textcolor{clearsky!95!black}{Policy Extraction.}
        \State \quad Train $\pi^\text{AWR}_h$ and $\pi^\text{BC}_l$ using Equations \ref{eq: high_policy} and \ref{eq: low_policy}. 
    \Until{convergence}
    }
\end{algorithmic}}
\end{algorithm}

\textbf{Policy Extraction}\quad Given the learned action abstraction and value functions, we extract a policy at each level using conservative policy extraction algorithms \citep{peng2019advantage, ghasemipour2021emaq}. 
The high-level policy uses Rejection Sampling (RS) over an advantage-weighted behaviour cloning (BC) flow policy; the low-level uses BC alone. We use advantage-weighting at the high-level because its dataset contains a large proportion of random goals, needed to cover the otherwise out-of-distribution state-goal pairs faced at deployment. While pure BC learns to ignore the goals, advantage-weighting restores goal-conditioning and leads to better proposals for RS. 
This issue does not arise at the low-level, for which we sample almost no random target states. In fact, we empirically found that advantage-weighting and RS degraded low-level performance by substantially reducing the effective batch size and exploiting errors in the Q function. 
Even though the low-level is trained with BC, we still get conservative extrapolation, as the high-level policy can pair a state with a better option than observed at that state during training; the low-level then pursues the option with behaviour learned elsewhere in the dataset. Figure \ref{fig: ablation_policy_extraction} and Table \ref{tab: goal_sensitivity} in Appendix \ref{app: experiments} show performance and goal sensitivity when ablating high-level policy extraction. 

\begin{algorithm}
{\footnotesize
\caption{Q-Shaped Options (QSO) Deployment}
\label{alg: qso_deployment}
\begin{algorithmic}
    {
    \footnotesize
    \While{episode not terminated}
        \State $\triangleright$ \textcolor{clearsky!95!black}{High-level Rejection Sampling.} 
        \State \quad Sample $\omega^1 \cdots \omega^N \sim \pi^\text{AWR}_h (\cdot \mid s_t, g)$. 
        \State \quad Select $\omega \gets \text{argmax}_{\omega^1 \cdots \omega^N} Q_h(s_t, g, \omega^i)$.
        \State $\triangleright$ \textcolor{clearsky!95!black}{Low-level Behaviour Cloning.} 
        \State \quad Sample $a_t \sim \pi^\text{BC}_l(\cdot \mid s_t, \omega)$.
    \EndWhile
    }
\end{algorithmic}}
\end{algorithm}

We provide pseudocode for QSO training and deployment in Algorithms \ref{alg: qso_training} and \ref{alg: qso_deployment}, and full equations and dataset sampling methods in Appendix \ref{app: method}. We refer to Appendix \ref{app: extended_preliminaries} for our choice of IQL and policy extraction algorithms.

\section{Experiments}
\label{sec: experiments}
We first evaluate QSO’s downstream performance from a fixed offline dataset to isolate policy learning from data collection. To assess our second and third desiderata, we introduce ASO, which shares QSO’s policy extraction but, like SHARSA and DSHARSA, uses absolute subgoal options instead of a learned action abstraction. This isolates the effect of QSO’s action abstraction: discarding distinctions necessary for control should yield a worse policy than ASO, whereas discarding only unnecessary ones should improve performance through greater data aggregation. We then analyse the learned option space directly, testing whether functionally equivalent behaviours map to similar points while functionally distinct ones remain separated. Finally, we ablate QSO’s components to identify their individual contributions.

\textbf{Experimental Protocol}\quad We evaluate on the most challenging OGBench \citep{park2025ogbench} environments using the default datasets, as OGBench is a standard offline RL benchmark with complex, long-horizon, goal-conditioned tasks. We compare against hierarchical algorithms from Section~\ref{sec: related_works}. As non-hierarchical baselines, we include IQL with DDPGBC \citep{fujimoto2021minimalist} policy extraction, the strongest non-hierarchical baseline \citep{park2025ogbench, park2024value}, and QC. All baselines use the hyperparameters from their original implementations. Apart from the option dimension, for which we use the OGBench default, QSO introduces no additional hyperparameters, so we use those of SHARSA and DSHARSA. Environment, baseline and hyperparameter details are in Appendix~\ref{app: experiments}.
\begin{figure}[t]
    \centering
    \includegraphics[width=\linewidth]{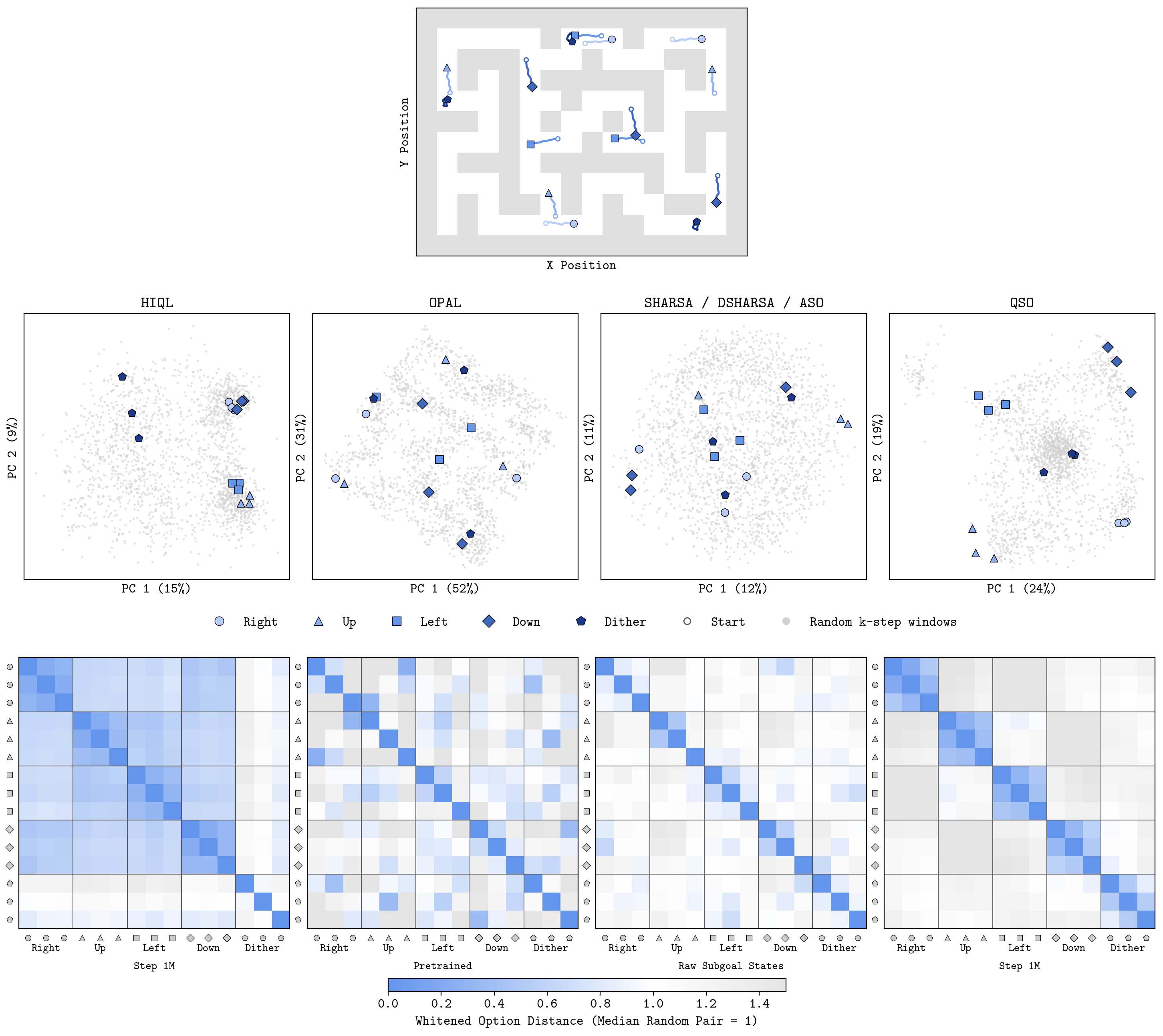}
    \caption{\textbf{PCA} (\textbf{top}) of the action abstraction for antmaze-giant fitted on random segments, showing different groups of segments with semantically similar behaviour (randomly selected from the validation dataset).  \textbf{L2 distances} (\textbf{bottom}) in units of median distance between two random segments. QSO is the only algorithm that consistently groups functionally equivalent behaviours (\textbf{Desideratum 3}) while separating behaviour groups (\textbf{Desideratum 2}).}
    \label{fig: ant_option_embeddings}
\end{figure}

\begin{table}[H]
\centering
\providecommand{\resultcell}[1]{\rule{0pt}{3.2ex}#1\rule[-1.4ex]{0pt}{0pt}}
\renewcommand{\arraystretch}{1.0}
\setlength{\tabcolsep}{4pt}
\resizebox{\textwidth}{!}{\ttfamily
\begin{tabular}{lcc>{\columncolor[HTML]{F2F2F2}}c>{\columncolor[HTML]{F2F2F2}}c>{\columncolor[HTML]{F2F2F2}}c>{\columncolor[HTML]{F2F2F2}}c>{\columncolor[HTML]{F2F2F2}}c>{\columncolor[HTML]{EDF2FE}}c|c}
\toprule
 & \multicolumn{2}{c}{Non-Hierarchical} & \multicolumn{6}{c}{Hierarchical} & \multirow{2}{*}{\begin{tabular}{@{}c@{}}\textcolor[HTML]{6495ED}{QSO} /\\ Best Mean\end{tabular}} \\
\cmidrule(lr){2-3} \cmidrule(lr){4-9}
Environment & IQL & QC & HIQL & OPAL & SHARSA & DSHARSA & ASO & \textcolor[HTML]{6495ED}{QSO} &  \\
\midrule
antmaze-giant-stitch-v0 & \resultcell{0 {\footnotesize [0, 0]}} & \resultcell{0 {\footnotesize [0, 0]}} & \resultcell{54 {\footnotesize [48, 60]}} & \resultcell{53 {\footnotesize [52, 55]}} & \resultcell{68 {\footnotesize [61, 76]}} & \resultcell{\textcolor[HTML]{6495ED}{\textbf{72}} {\footnotesize [67, 77]}} & \resultcell{68 {\footnotesize [66, 70]}} & \resultcell{68 {\footnotesize [66, 70]}} & \resultcell{94\%} \\
antmaze-teleport-stitch-v0 & \resultcell{47 {\footnotesize [44, 51]}} & \resultcell{37 {\footnotesize [32, 40]}} & \resultcell{36 {\footnotesize [34, 39]}} & \resultcell{\textcolor[HTML]{6495ED}{\textbf{69}} {\footnotesize [64, 74]}} & \resultcell{53 {\footnotesize [52, 55]}} & \resultcell{58 {\footnotesize [55, 62]}} & \resultcell{66 {\footnotesize [62, 69]}} & \resultcell{54 {\footnotesize [52, 57]}} & \resultcell{79\%} \\
humanoidmaze-giant-stitch-v0 & \resultcell{1 {\footnotesize [1, 1]}} & \resultcell{0 {\footnotesize [0, 0]}} & \resultcell{6 {\footnotesize [4, 8]}} & \resultcell{82 {\footnotesize [78, 85]}} & \resultcell{30 {\footnotesize [27, 33]}} & \resultcell{30 {\footnotesize [27, 32]}} & \resultcell{44 {\footnotesize [38, 48]}} & \resultcell{\textcolor[HTML]{6495ED}{\textbf{90}} {\footnotesize [88, 92]}} & \resultcell{100\%} \\
\midrule
cube-double-play-v0 & \resultcell{20 {\footnotesize [17, 23]}} & \resultcell{72 {\footnotesize [70, 74]}} & \resultcell{7 {\footnotesize [5, 9]}} & \resultcell{29 {\footnotesize [26, 33]}} & \resultcell{72 {\footnotesize [68, 77]}} & \resultcell{\textcolor[HTML]{6495ED}{\textbf{80}} {\footnotesize [75, 84]}} & \resultcell{68 {\footnotesize [65, 70]}} & \resultcell{70 {\footnotesize [65, 74]}} & \resultcell{87\%} \\
cube-triple-play-v0 & \resultcell{6 {\footnotesize [6, 7]}} & \resultcell{\textcolor[HTML]{6495ED}{\textbf{23}} {\footnotesize [19, 27]}} & \resultcell{16 {\footnotesize [14, 18]}} & \resultcell{3 {\footnotesize [2, 4]}} & \resultcell{14 {\footnotesize [10, 18]}} & \resultcell{14 {\footnotesize [13, 16]}} & \resultcell{4 {\footnotesize [4, 4]}} & \resultcell{22 {\footnotesize [20, 23]}} & \resultcell{92\%} \\
cube-quadruple-play-v0 & \resultcell{0 {\footnotesize [0, 0]}} & \resultcell{2 {\footnotesize [1, 2]}} & \resultcell{3 {\footnotesize [1, 5]}} & \resultcell{0 {\footnotesize [0, 0]}} & \resultcell{0 {\footnotesize [0, 1]}} & \resultcell{0 {\footnotesize [0, 1]}} & \resultcell{0 {\footnotesize [0, 0]}} & \resultcell{\textcolor[HTML]{6495ED}{\textbf{4}} {\footnotesize [4, 5]}} & \resultcell{100\%} \\
scene-play-v0 & \resultcell{48 {\footnotesize [45, 51]}} & \resultcell{82 {\footnotesize [81, 83]}} & \resultcell{41 {\footnotesize [35, 46]}} & \resultcell{84 {\footnotesize [78, 87]}} & \resultcell{92 {\footnotesize [90, 93]}} & \resultcell{89 {\footnotesize [87, 91]}} & \resultcell{\textcolor[HTML]{6495ED}{\textbf{94}} {\footnotesize [91, 95]}} & \resultcell{85 {\footnotesize [82, 90]}} & \resultcell{91\%} \\
puzzle-3x3-play-v0 & \resultcell{\textcolor[HTML]{6495ED}{\textbf{100}} {\footnotesize [99, 100]}} & \resultcell{81 {\footnotesize [46, 99]}} & \resultcell{48 {\footnotesize [46, 49]}} & \resultcell{70 {\footnotesize [65, 74]}} & \resultcell{78 {\footnotesize [71, 87]}} & \resultcell{85 {\footnotesize [81, 90]}} & \resultcell{76 {\footnotesize [72, 80]}} & \resultcell{72 {\footnotesize [69, 75]}} & \resultcell{72\%} \\
puzzle-4x4-play-v0 & \resultcell{71 {\footnotesize [61, 78]}} & \resultcell{25 {\footnotesize [23, 27]}} & \resultcell{0 {\footnotesize [0, 1]}} & \resultcell{64 {\footnotesize [61, 68]}} & \resultcell{74 {\footnotesize [67, 79]}} & \resultcell{72 {\footnotesize [69, 75]}} & \resultcell{69 {\footnotesize [65, 75]}} & \resultcell{\textcolor[HTML]{6495ED}{\textbf{92}} {\footnotesize [81, 100]}} & \resultcell{100\%} \\
puzzle-4x5-play-v0 & \resultcell{14 {\footnotesize [11, 15]}} & \resultcell{16 {\footnotesize [13, 19]}} & \resultcell{4 {\footnotesize [3, 5]}} & \resultcell{\textcolor[HTML]{6495ED}{\textbf{24}} {\footnotesize [22, 25]}} & \resultcell{17 {\footnotesize [15, 20]}} & \resultcell{18 {\footnotesize [18, 19]}} & \resultcell{14 {\footnotesize [11, 18]}} & \resultcell{15 {\footnotesize [11, 20]}} & \resultcell{65\%} \\
puzzle-4x6-play-v0 & \resultcell{8 {\footnotesize [5, 10]}} & \resultcell{16 {\footnotesize [14, 19]}} & \resultcell{5 {\footnotesize [4, 6]}} & \resultcell{17 {\footnotesize [15, 18]}} & \resultcell{14 {\footnotesize [10, 16]}} & \resultcell{16 {\footnotesize [14, 17]}} & \resultcell{11 {\footnotesize [8, 15]}} & \resultcell{\textcolor[HTML]{6495ED}{\textbf{18}} {\footnotesize [15, 22]}} & \resultcell{100\%} \\
\midrule
Average over 11 Environments & \resultcell{29 {\footnotesize [27, 30]}} & \resultcell{32 {\footnotesize [29, 34]}} & \cellcolor[HTML]{E6E6E6}\resultcell{20 {\footnotesize [20, 20]}} & \cellcolor[HTML]{E6E6E6}\resultcell{45 {\footnotesize [44, 46]}} & \cellcolor[HTML]{E6E6E6}\resultcell{46 {\footnotesize [44, 49]}} & \cellcolor[HTML]{E6E6E6}\resultcell{49 {\footnotesize [48, 49]}} & \cellcolor[HTML]{E6E6E6}\resultcell{47 {\footnotesize [46, 48]}} & \cellcolor[HTML]{DCE6FC}\resultcell{\textcolor[HTML]{6495ED}{\textbf{54}} {\footnotesize [51, 56]}} & \resultcell{100\%} \\
\bottomrule
\end{tabular}}
\caption{\textbf{Summary of results on selected OGBench environments using the default datasets.} Success rate averaged over 5 evaluation tasks after 1M training steps. Mean and 95\% bootstrap confidence interval over 4 seeds.}
\label{tab: results_main}
\end{table}

\textbf{Results}\quad Table \ref{tab: results_main} shows that QSO achieves the highest average success rate across the 11 environments, 54\% against 49\% for the next best method (DSHARSA). Particularly notable is its performance in humanoidmaze, which has the largest action space dimension (21, against at most 8 in the other environments) where it achieves 90\% against 82\% for the next best method and three times that of DSHARSA. In puzzle-4x4 it achieves 92\% against at most 74\% for any baseline. In cube-quadruple, the hardest manipulation environment, QSO is the only method with non-zero success on two of the five tasks (Table~\ref{tab: results_appendix_manipulation} in Appendix \ref{app: experiments}). QSO performs consistently across environments, achieving at least 65\% of the best success rate in all environments, while every other algorithm drops below 35\% of the best in at least one.

\textbf{Is the action abstraction semantically meaningful?}\quad We test whether QSO maps functionally equivalent behaviours to similar parts of the option space by identifying groups of semantically similar behaviours in antmaze-giant, cube-double, and cube-quadruple (Figures \ref{fig: ant_option_embeddings}, \ref{fig: cube_double_option_embeddings}, and \ref{fig: cube_quadruple_option_embeddings}). For each behaviour, we randomly sample trajectory sequences from the validation dataset and compare PCA projections and whitened distances of their encodings. An invariant encoder should produce a clear block-diagonal similarity structure. QSO exhibits this structure, whereas HIQL also assigns high similarity between behaviour groups, and OPAL and absolute subgoal options produce low similarity within behaviour groups. 

\begin{figure}[H]
    \centering
    \includegraphics[width=0.75\linewidth]{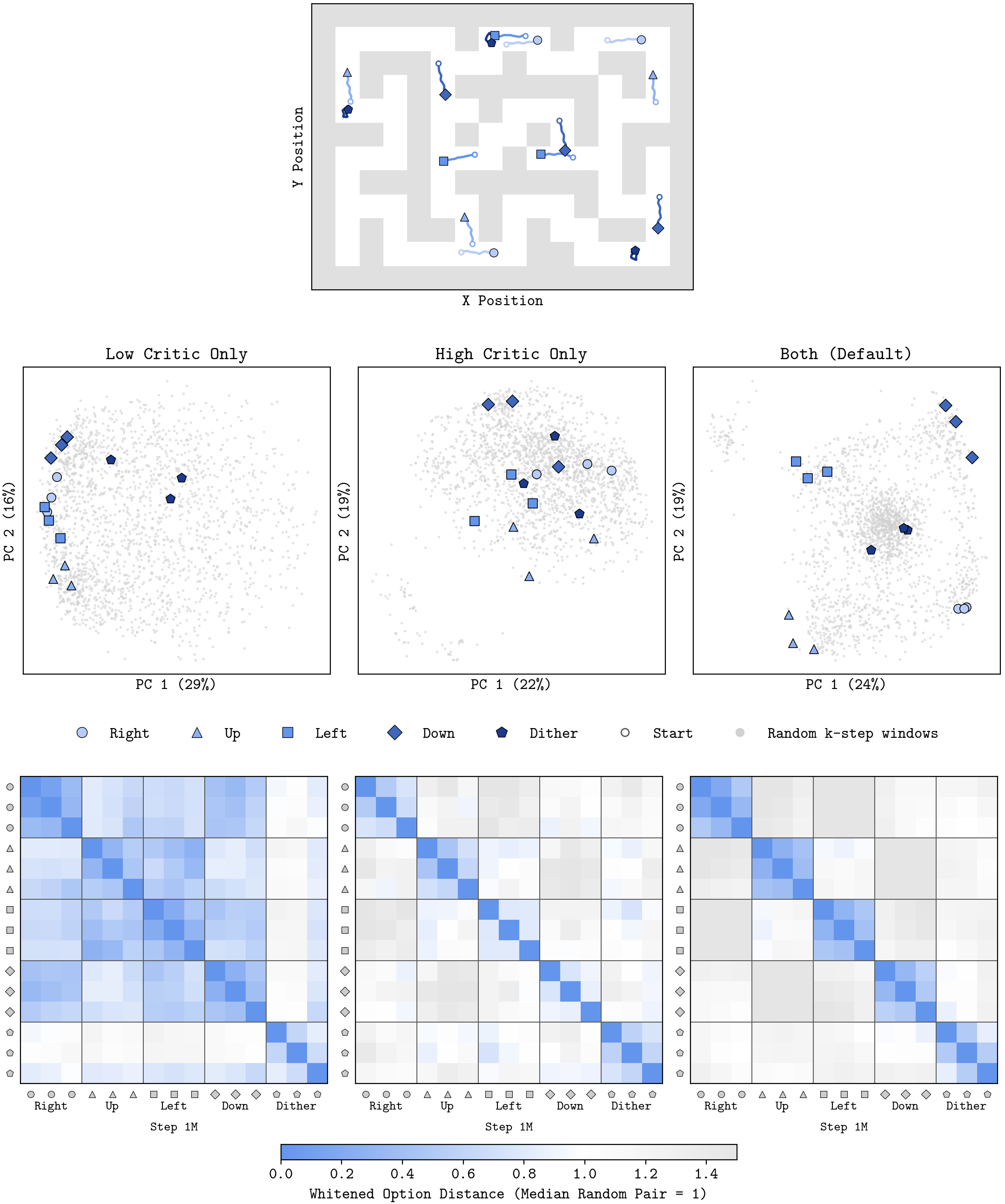}
    \caption{\textbf{PCA} (\textbf{top}) of QSO's action abstractions for gradient flow ablations for antmaze-giant fitted on random trajectory sequences, showing different groups of three k-step trajectory sequences with semantically similar behaviour (randomly selected from the validation dataset).  \textbf{L2 distances} (\textbf{bottom}) for those groups in units of median distance between two random trajectory sequences. Both high and low-level gradient flows are required to group functionally equivalent behaviours while separating the behaviour groups.}
    \label{fig: ant_critic_option_embeddings}
\end{figure}

\textbf{How necessary is joint option shaping by both the high and low-level Q functions?}\quad Table \ref{tab: qso_ablation} and Figures \ref{fig: ant_critic_option_embeddings} and \ref{fig: cube_critic_option_embeddings} indicate that joint shaping is important for both the learned action abstraction and downstream performance. Neither Q function alone recovers the clean semantic structure obtained when both are used. Figure \ref{fig: ablation_gradient_flow} shows that shaping with the high-level only performs well in antmaze-giant but degrades significantly in cube-double. We hypothesise that for cube-double continued high-level shaping increasingly removes distinctions that are irrelevant to high-level value but necessary for fine-grained control; without the low-level Q function to preserve this information, the resulting abstractions become insufficient for the low-level policy. 

\textbf{Do the decoupled levels form distinct state abstractions?}\quad For the high and low-level decision processes, we measure how much each network's output varies with slow and fast moving dimensions; a larger percentage share of the output variance being associated to a particular dimension means that the network is more dependent on that dimension. We would expect the low-level decision process to focus more on high frequency (fast-moving) dimensions, and the high-level decision process on low frequency (slow-moving) ones. 
As expected, the two levels focus on different states (Table \ref{tab: state_abstraction} in Appendix \ref{app: experiments}). For example, the high-level state-value function depends almost entirely on slow-moving dimensions (88\% for antmaze-giant and 74\% for cube-double), while the low-level policy depends much more on fast-moving ones (59\% and 77\% respectively). We refer to Appendix \ref{app: experiments} for further discussion. 

\begin{table}[H]
\centering
\providecommand{\resultcell}[1]{\rule{0pt}{3.2ex}#1\rule[-1.4ex]{0pt}{0pt}}
\renewcommand{\arraystretch}{1.0}
\resizebox{\textwidth}{!}{\ttfamily\setlength{\tabcolsep}{\dimexpr 4.0pt+0.3000em+0.9750pt\relax}%
\begin{tabular}{l>{\columncolor[HTML]{EDF2FE}}c|>{\columncolor[HTML]{F2F2F2}}c>{\columncolor[HTML]{F2F2F2}}c|cccc}
\toprule
 & \multicolumn{1}{>{\columncolor[HTML]{EDF2FE}}c|}{Default} & \multicolumn{2}{>{\columncolor[HTML]{F2F2F2}}c|}{Q Shaping Ablation} & \multicolumn{4}{c}{Option Resampling Ablation} \\
\cmidrule(lr){2-2} \cmidrule(lr){3-4} \cmidrule(lr){5-8}
Environment & \textcolor[HTML]{6495ED}{QSO} & Low-Level Only & High-Level Only & Every Step & 5\% (2 Steps) & 25\% (7 Steps) & 50\% (13 Steps)\\
\midrule
antmaze-giant-stitch-v0 & \resultcell{68 {\footnotesize [66, 70]}} & \resultcell{38 {\footnotesize [38, 40]}} & \resultcell{69 {\footnotesize [67, 72]}} & \resultcell{68 {\footnotesize [66, 70]}} & \resultcell{68 {\footnotesize [66, 70]}} & \resultcell{62 {\footnotesize [61, 64]}} & \resultcell{54 {\footnotesize [47, 60]}} \\
cube-double-play-v0 & \resultcell{70 {\footnotesize [65, 74]}} & \resultcell{70 {\footnotesize [67, 73]}} & \resultcell{47 {\footnotesize [42, 50]}} & \resultcell{70 {\footnotesize [65, 74]}} & \resultcell{69 {\footnotesize [63, 73]}} & \resultcell{50 {\footnotesize [44, 55]}} & \resultcell{14 {\footnotesize [10, 18]}} \\
\bottomrule
\end{tabular}}
\caption{\textbf{Q shaping and option resampling ablations of QSO.} Success rate averaged over 5 evaluation tasks after 1M training steps. Mean and 95\% bootstrap confidence interval over 4 seeds.}
\label{tab: qso_ablation}
\end{table}

\begin{figure}
    \centering
    \includegraphics[width=0.6\linewidth]{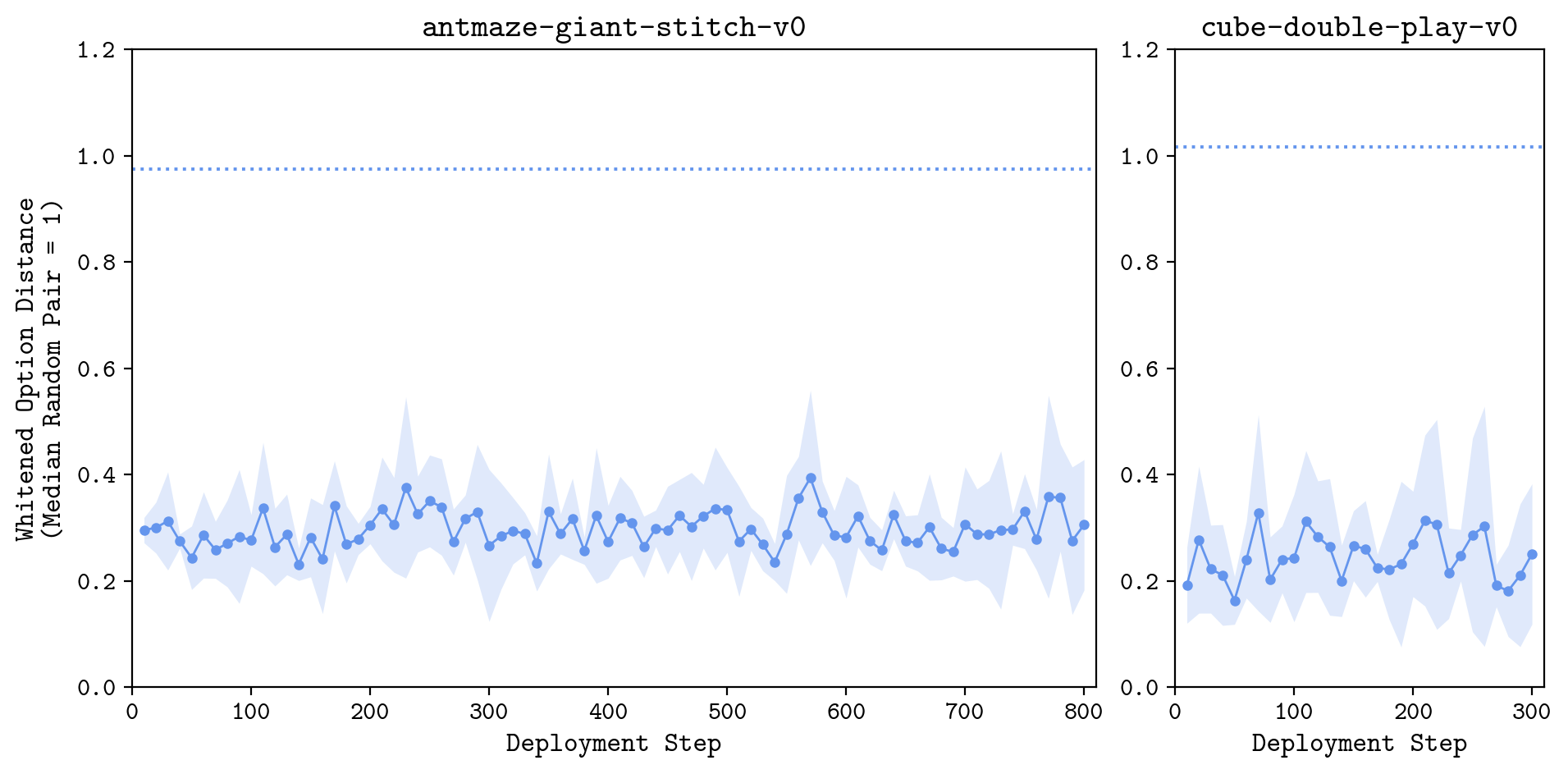}
    \caption{\textbf{QSO's sampled options gradually shift across the option space during deployment.} L2 distance between two consecutively sampled options (every 10 steps), whitened against the distance between two randomly sampled options. \textbf{Dotted blue} shows the distance between two random trajectory sequence embeddings whitened against the distance between two randomly sampled options. Mean and 95\% confidence interval over 4 seeds, cut off where the third quartile of successful episodes has ended.}
    \label{fig: option_deployment}
\end{figure}

\textbf{How does the option vary during deployment?}\quad Figure \ref{fig: option_deployment} shows that sampled options gradually shift across the option space during deployment: the distance between consecutively sampled options is roughly 30\% of that between two randomly sampled options in antmaze-giant, and 25\% of that in cube-double. Although consecutive options may partly be similar due to smoothness of the high-level policy, the above analysis showing that option-space proximity reflects semantic similarity suggests that it also corresponds to gradual changes in intent. The distance between two random trajectory sequence embeddings whitened against the distance between two randomly sampled options (dotted blue) being approximately 1 confirms that the small consecutive distances reflect genuinely slow variation rather than a collapsed policy. Table \ref{tab: qso_ablation} ablates how long the option is held for at deployment: the rapid decrease in performance for cube-double suggests that the option encodes significant information about the immediate next action.

We provide ablations on the option representation input and dimension, goal sampling, reward labelling, policy extraction, and temporal horizon in Appendix~\ref{app: experiments}.

\section{Discussion and Future Work}
We introduce QSO, which, unlike prior algorithms, learns a control-sufficient action abstraction that discards unnecessary distinctions between options. Future work will extend QSO to do targeted exploration at the option-level in online RL. A limitation of QSO arises in stochastic environments: by representing an option retrospectively, the high-level critic evaluates the option conditioned on its realised outcome, which can lead to optimistic bias. Although we note that this issue is shared by all hindsight-based hierarchical methods including HIQL, OPAL, SHARSA and DSHARSA, future work could address this through progress-adaptive high-level control.

\section*{Ethics Statement}
This work aims to advance machine learning, specifically hierarchical reinforcement learning, and is evaluated only in simulated environments using public datasets. It may have broader societal consequences, but none of them must be specifically highlighted here.

\section*{Acknowledgements}
The authors would like to thank \textbf{Darius Muglich} for helpful advice and comments on the project.

\textbf{Clarisse Wibault} is funded by EPSRC grant EP/W524311/1. \textbf{Antoine Gorceix} is funded by the EPSRC Centre for Doctoral Training in Autonomous Intelligent Machines grant EP/Y035070/1 and Iconic Interactive Ltd. \textbf{Alexey Zakharov} is supported by the Clarendon Fund in partnership with the Oxford-Basil Reeve Graduate Scholarship at the University of Oxford. \textbf{Eduardo Pignatelli} is funded by University College London. \textbf{Jakob Foerster} is partially funded by the UKRI grant EP/Y028481/1, as well as being supported by the JPMC and Amazon Research Awards.

\section*{Reproducibility Statement}
Our code is open sourced at \url{https://github.com/CWibault/QSO}. Hyperparameters and exact implementation details can be found in Appendices \ref{app: method} and \ref{app: experiments}. 

\section*{AI Disclosure}
We used AI to render figures, debug code, and check spelling and grammar, but all ideas, analysis, and written content are our own. We take full responsibility for the final submission.

\bibliography{oak}

@article{kingma2014adam,
  title={Adam: A method for stochastic optimization},
  author={Kingma, Diederik P and Ba, Jimmy},
  journal={arXiv preprint arXiv:1412.6980},
  year={2014}
}

@article{hendrycks2016gaussian,
  title={Gaussian error linear units (gelus)},
  author={Hendrycks, Dan and Gimpel, Kevin},
  journal={arXiv preprint arXiv:1606.08415},
  year={2016}
}

@article{zhang2020learning,
  title={Learning invariant representations for reinforcement learning without reconstruction},
  author={Zhang, Amy and McAllister, Rowan and Calandra, Roberto and Gal, Yarin and Levine, Sergey},
  journal={arXiv preprint arXiv:2006.10742},
  year={2020}
}

@inproceedings{abel2016near,
  title={Near optimal behavior via approximate state abstraction},
  author={Abel, David and Hershkowitz, David and Littman, Michael},
  booktitle={International Conference on Machine Learning},
  pages={2915--2923},
  year={2016},
  organization={PMLR}
}

@inproceedings{chen2019information,
  title={Information-theoretic considerations in batch reinforcement learning},
  author={Chen, Jinglin and Jiang, Nan},
  booktitle={International conference on machine learning},
  pages={1042--1051},
  year={2019},
  organization={PMLR}
}

@inproceedings{fujimoto2018addressing,
  title={Addressing function approximation error in actor-critic methods},
  author={Fujimoto, Scott and Hoof, Herke and Meger, David},
  booktitle={International conference on machine learning},
  pages={1587--1596},
  year={2018},
  organization={Pmlr}
}

@article{lillicrap2015continuous,
  title={Continuous control with deep reinforcement learning},
  author={Lillicrap, Timothy P and Hunt, Jonathan J and Pritzel, Alexander and Heess, Nicolas and Erez, Tom and Tassa, Yuval and Silver, David and Wierstra, Daan},
  journal={arXiv preprint arXiv:1509.02971},
  year={2015}
}

@inproceedings{silver2014deterministic,
  title={Deterministic policy gradient algorithms},
  author={Silver, David and Lever, Guy and Heess, Nicolas and Degris, Thomas and Wierstra, Daan and Riedmiller, Martin},
  booktitle={International conference on machine learning},
  pages={387--395},
  year={2014},
  organization={Pmlr}
}

@inproceedings{sutton2011horde,
  title={Horde: A scalable real-time architecture for learning knowledge from unsupervised sensorimotor interaction},
  author={Sutton, Richard S and Modayil, Joseph and Delp, Michael and Degris, Thomas and Pilarski, Patrick M and White, Adam and Precup, Doina},
  booktitle={The 10th international conference on autonomous agents and multiagent systems-volume 2},
  pages={761--768},
  year={2011}
}

@article{mnih2015human,
  title={Human-level control through deep reinforcement learning},
  author={Mnih, Volodymyr and Kavukcuoglu, Koray and Silver, David and Rusu, Andrei A and Veness, Joel and Bellemare, Marc G and Graves, Alex and Riedmiller, Martin and Fidjeland, Andreas K and Ostrovski, Georg and others},
  journal={nature},
  volume={518},
  number={7540},
  pages={529--533},
  year={2015},
  publisher={Nature Publishing Group UK London}
}

@article{watkins1992q,
  title={Q-learning},
  author={Watkins, Christopher JCH and Dayan, Peter},
  journal={Machine learning},
  volume={8},
  number={3},
  pages={279--292},
  year={1992},
  publisher={Springer}
}

@article{park2024value,
  title={Is value learning really the main bottleneck in offline RL?},
  author={Park, Seohong and Frans, Kevin and Levine, Sergey and Kumar, Aviral},
  journal={Advances in Neural Information Processing Systems},
  volume={37},
  pages={79029--79056},
  year={2024}
}

@book{kakade2003sample,
  title={On the sample complexity of reinforcement learning},
  author={Kakade, Sham Machandranath},
  year={2003},
  publisher={University of London, University College London (United Kingdom)}
}

@article{robert2023sample,
  title={Sample complexity of goal-conditioned hierarchical reinforcement learning},
  author={Robert, Arnaud and Pike-Burke, Ciara and Faisal, Aldo},
  journal={Advances in Neural Information Processing Systems},
  volume={36},
  pages={62696--62712},
  year={2023}
}

@article{munos2008finite,
  title={Finite-Time Bounds for Fitted Value Iteration.},
  author={Munos, R{\'e}mi and Szepesv{\'a}ri, Csaba},
  journal={Journal of Machine Learning Research},
  volume={9},
  number={5},
  year={2008}
}

@article{li2006towards,
  title={Towards a unified theory of state abstraction for MDPs.},
  author={Li, Lihong and Walsh, Thomas J and Littman, Michael L},
  journal={AI\&M},
  volume={1},
  number={2},
  pages={3},
  year={2006}
}

@inproceedings{lattimore2012pac,
  title={PAC bounds for discounted MDPs},
  author={Lattimore, Tor and Hutter, Marcus},
  booktitle={International Conference on Algorithmic Learning Theory},
  pages={320--334},
  year={2012},
  organization={Springer}
}

@inproceedings{white2012scaling,
  title={Scaling life-long off-policy learning},
  author={White, Adam and Modayil, Joseph and Sutton, Richard S},
  booktitle={2012 ieee international conference on development and learning and epigenetic robotics (icdl)},
  pages={1--6},
  year={2012},
  organization={IEEE}
}

@inproceedings{ghasemipour2021emaq,
  title={Emaq: Expected-max q-learning operator for simple yet effective offline and online rl},
  author={Ghasemipour, Seyed Kamyar Seyed and Schuurmans, Dale and Gu, Shixiang Shane},
  booktitle={International Conference on Machine Learning},
  pages={3682--3691},
  year={2021},
  organization={PMLR}
}

@article{peng2019advantage,
  title={Advantage-weighted regression: Simple and scalable off-policy reinforcement learning},
  author={Peng, Xue Bin and Kumar, Aviral and Zhang, Grace and Levine, Sergey},
  journal={arXiv preprint arXiv:1910.00177},
  year={2019}
}

@article{fujimoto2021minimalist,
  title={A minimalist approach to offline reinforcement learning},
  author={Fujimoto, Scott and Gu, Shixiang Shane},
  journal={Advances in neural information processing systems},
  volume={34},
  pages={20132--20145},
  year={2021}
}

@article{kostrikov2021offline,
  title={Offline reinforcement learning with implicit q-learning},
  author={Kostrikov, Ilya and Nair, Ashvin and Levine, Sergey},
  journal={arXiv preprint arXiv:2110.06169},
  year={2021}
}

@article{park2026horizon,
  title={Horizon reduction makes rl scalable},
  author={Park, Seohong and Frans, Kevin and Mann, Deepinder and Eysenbach, Benjamin and Kumar, Aviral and Levine, Sergey},
  journal={Advances in Neural Information Processing Systems},
  volume={38},
  pages={8350--8389},
  year={2026}
}

@article{ajay2020opal,
  title={Opal: Offline primitive discovery for accelerating offline reinforcement learning},
  author={Ajay, Anurag and Kumar, Aviral and Agrawal, Pulkit and Levine, Sergey and Nachum, Ofir},
  journal={arXiv preprint arXiv:2010.13611},
  year={2020}
}

@inproceedings{vezhnevets2017feudal,
  title={Feudal networks for hierarchical reinforcement learning},
  author={Vezhnevets, Alexander Sasha and Osindero, Simon and Schaul, Tom and Heess, Nicolas and Jaderberg, Max and Silver, David and Kavukcuoglu, Koray},
  booktitle={International conference on machine learning},
  pages={3540--3549},
  year={2017},
  organization={PMLR}
}

@article{nachum2018data,
  title={Data-efficient hierarchical reinforcement learning},
  author={Nachum, Ofir and Gu, Shixiang Shane and Lee, Honglak and Levine, Sergey},
  journal={Advances in neural information processing systems},
  volume={31},
  year={2018}
}

@article{zhao2023learning,
  title={Learning fine-grained bimanual manipulation with low-cost hardware},
  author={Zhao, Tony Z and Kumar, Vikash and Levine, Sergey and Finn, Chelsea},
  journal={arXiv preprint arXiv:2304.13705},
  year={2023}
}

@article{li2026reinforcement,
  title={Reinforcement learning with action chunking},
  author={Li, Qiyang and Zhou, Zhiyuan Paul and Levine, Sergey},
  journal={Advances in Neural Information Processing Systems},
  volume={38},
  pages={55518--55553},
  year={2026}
}

@inproceedings{bacon2017option,
  title={The option-critic architecture},
  author={Bacon, Pierre-Luc and Harb, Jean and Precup, Doina},
  booktitle={Proceedings of the AAAI conference on artificial intelligence},
  volume={31},
  year={2017}
}

@book{sutton1998reinforcement,
  title={Reinforcement learning: An introduction},
  author={Sutton, Richard S and Barto, Andrew G and Barto, Andrew},
  volume={1},
  year={1998},
  publisher={MIT press Cambridge}
}

@article{barto2003recent,
  title={Recent advances in hierarchical reinforcement learning},
  author={Barto, Andrew G and Mahadevan, Sridhar},
  journal={Discrete event dynamic systems},
  volume={13},
  number={4},
  pages={341--379},
  year={2003},
  publisher={Springer}
}

@inproceedings{park2025ogbench,
  title={Ogbench: Benchmarking offline goal-conditioned rl},
  author={Park, Seohong and Frans, Kevin and Eysenbach, Benjamin and Levine, Sergey},
  booktitle={International Conference on Learning Representations},
  volume={2025},
  pages={94937--94982},
  year={2025}
}

@article{andrychowicz2017hindsight,
  title={Hindsight experience replay},
  author={Andrychowicz, Marcin and Wolski, Filip and Ray, Alex and Schneider, Jonas and Fong, Rachel and Welinder, Peter and McGrew, Bob and Tobin, Josh and Pieter Abbeel, OpenAI and Zaremba, Wojciech},
  journal={Advances in neural information processing systems},
  volume={30},
  year={2017}
}

@incollection{lange2012batch,
  title={Batch reinforcement learning},
  author={Lange, Sascha and Gabel, Thomas and Riedmiller, Martin},
  booktitle={Reinforcement learning: State-of-the-art},
  pages={45--73},
  year={2012},
  publisher={Springer}
}

@article{levine2020offline,
  title={Offline reinforcement learning: Tutorial, review, and perspectives on open problems},
  author={Levine, Sergey and Kumar, Aviral and Tucker, George and Fu, Justin},
  journal={arXiv preprint arXiv:2005.01643},
  year={2020}
}

@inproceedings{ravindran2002model,
  title={Model minimization in hierarchical reinforcement learning},
  author={Ravindran, Balaraman and Barto, Andrew G},
  booktitle={International symposium on abstraction, reformulation, and approximation},
  pages={196--211},
  year={2002},
  organization={Springer}
}

@article{konidaris2018skills,
  title={From skills to symbols: Learning symbolic representations for abstract high-level planning},
  author={Konidaris, George and Kaelbling, Leslie Pack and Lozano-Perez, Tomas},
  journal={Journal of Artificial Intelligence Research},
  volume={61},
  pages={215--289},
  year={2018}
}

@misc{sutton2025oakvideo,
  author       = {Richard S. Sutton},
  title        = {The {OaK} Architecture: A Vision of Superintelligence from Experience},
  howpublished = {YouTube Video},
  month        = {August},
  year         = {2025},
  url          = {},
  note         = {Accessed: August 2026}
}

@inproceedings{schaul2015universal,
  title={Universal value function approximators},
  author={Schaul, Tom and Horgan, Daniel and Gregor, Karol and Silver, David},
  booktitle={International conference on machine learning},
  pages={1312--1320},
  year={2015},
  organization={PMLR}
}

@inproceedings{kaelbling1993learning,
  title={Learning to achieve goals},
  author={Kaelbling, Leslie Pack},
  booktitle={Ijcai},
  volume={2},
  pages={1094--8},
  year={1993}
}

@article{ferns2012metrics,
  title={Metrics for finite Markov decision processes},
  author={Ferns, Norman and Panangaden, Prakash and Precup, Doina},
  journal={arXiv preprint arXiv:1207.4114},
  year={2012}
}

@article{konidaris2019necessity,
  title={On the necessity of abstraction},
  author={Konidaris, George},
  journal={Current opinion in behavioral sciences},
  volume={29},
  pages={1--7},
  year={2019},
  publisher={Elsevier}
}

@article{park2023hiql,
  title={Hiql: Offline goal-conditioned rl with latent states as actions},
  author={Park, Seohong and Ghosh, Dibya and Eysenbach, Benjamin and Levine, Sergey},
  journal={Advances in Neural Information Processing Systems},
  volume={36},
  pages={34866--34891},
  year={2023}
}

@article{sutton1999between,
  title={Between MDPs and semi-MDPs: A framework for temporal abstraction in reinforcement learning},
  author={Sutton, Richard S and Precup, Doina and Singh, Satinder},
  journal={Artificial intelligence},
  volume={112},
  number={1-2},
  pages={181--211},
  year={1999},
  publisher={Elsevier}
}
\bibliographystyle{unsrtnat}
\newpage
\appendix

\newpage
\section{Extended Preliminaries}
\label{app: extended_preliminaries}

\paragraph{Value Learning} 
To learn our state-value and Q functions, we adopt Implicit Q-Learning \citep{kostrikov2021offline}, a conservative value learning algorithm that evaluates the state-value function under the dataset action that leads to the maximum value rather than evaluating the Q-value under the current policy or using the maximum over actions, as is typically done in online RL \citep{watkins1992q, mnih2015human}. This avoids querying the Q function on out-of-distribution actions, to which the Q function can assign erroneously high values \citep{levine2020offline}. The state-value function and critic are trained with the following losses: 
\begin{equation}
    \begin{aligned}
        \mathcal{L}_V &= \mathbb{E}_{(s_t,a_t, g) \sim \mathcal{D}} \left[\ell^2_\nu\left(V(s_t, g) - \tilde{Q}(s_t, g, a_t) \right)\right], \\
    \mathcal{L}_{Q} &= \mathbb{E}_{(s_t,a_t,s_{t+1},g) \sim \mathcal{D}} \left[\left(Q(s_t,g,a_t) - r(s_t,g) - \gamma V(s_{t+1}, g)\right)^2\right], 
    \end{aligned}
    \notag
\end{equation}
where $\ell^2_\nu$ denotes the expectile loss $\ell^2_\nu(x):= \mid \nu - \mathbf{1}\cdot (x < 0)\mid \cdot x^2$. For $\nu >0.5$, underestimating a high Q value is penalised more strongly than overestimating a low one, so $V(s_t, g)$ is pulled to the upper end of the dataset $Q$ values. 

\paragraph{Policy Extraction} 
To extract our policies, we use Advantage Weighted Regression (AWR) \citep{peng2019advantage} and Rejection Sampling (RS)  \citep{ghasemipour2021emaq}, which are conservative policy extraction algorithms that constrain the policy to sample the highest quality actions within the dataset.
As the name suggests, AWR does advantage-weighted behaviour cloning of dataset actions:
\begin{equation}
        \mathcal{L}_\pi = -\mathbb{E}_{(s_t,a_t,g) \sim \mathcal{D}} \left[\exp\left(\frac{Q(s_t,g, a_t)- V(s_t,g)}{\alpha}\right) \cdot \log \pi(a_t \mid s_t,g)\right],
        \notag
\end{equation}
while RS learns a behaviour-cloned policy to imitate the dataset and then defines the policy by choosing the action that maximises the Q function:
\begin{equation}
    \pi := \text{argmax}_{a_t \in \{a_t^i \}, a_t^i \sim \pi^\text{BC}(\cdot \mid s_t,g)} Q(s_t,g,a_t^i).
    \notag
\end{equation}
Unlike algorithms that optimise the policy by differentiating through the learned Q-function \citep{silver2014deterministic, lillicrap2015continuous, fujimoto2018addressing, fujimoto2021minimalist}, these conservative policy extraction algorithms avoid a direct source of Q function exploitation, whereby errors in the learned Q values are propagated into the policy. However, a non-hierarchical policy extracted with AWR or RS is bounded by the quality of the data at that state: where a region of the state space contains only low-quality actions, there is nothing ``better'' among the proposals to upweight.

\newpage
\section{Low-level Q-shaping implies control sufficiency}
\label{app: control_sufficiency}
\begin{boxproof}{Low-level Q-shaping implies control sufficiency.}{proof_1}
    If $\phi_\omega(s, g_s) = \phi_\omega(s, g_s') =: \omega$, then for every $a$, $Q^\star(s, g_s, a) = Q_l(s, \omega, a) = Q^\star(s, g_s', a)$, so the two targets induce the same action-value function at $s$ and hence the same set of maximisers. Taking $\pi_l(\cdot \mid s, \omega)$ to be any distribution supported on $\text{argmax}_a Q_l(s, \omega, a)$ gives the policy required by  Definition~\ref{def:control_sufficiency}.
\end{boxproof}

\newpage
\section{Conservative Extrapolation}
\label{app: conservative_extrapolation}
To provide intuition on how joint state and action abstractions can enable conservative extrapolation, consider an agent learning a locomotion task, where its state can be decomposed into the agent’s pose and coordinate $s = (p, c)$. Imagine three state-goal pairs: 
\begin{equation}
    (s_1, g, \textcolor{red!90!black}{a_1}) = \big(\underbrace{(p, c)}_{s_1}, \; g, \textcolor{red!90!black}{a_1}\big), \quad
    (s_2, g, \textcolor{green!65!black}{a_2}) = \big(\underbrace{(p', c)}_{s_2}, \; g, \textcolor{green!65!black}{a_2}\big), \quad
    (s_3, g', \textcolor{green!65!black}{a_3}) = \big(\underbrace{(p, c')}_{s_3}, \; g', \textcolor{green!65!black}{a_3}\big).
    \notag
\end{equation}
The first state-goal pair is the one that the agent must act at, and its dataset action is poor (it might just dither on the spot), while the dataset actions at the second and third pair are good. Imagine that in all situations, the agent must move two metres down to reach its respective goal because the goals lie in the same direction from their respective coordinates. 

A non-hierarchical policy extracted with AWR or RS never learns a useful action at $(s_1, g)$ because the only action the dataset offers there is the poor one. We now show how a hierarchical policy with abstract relative options (such as stay ``on the spot'' or reach ``a point two metres down'') can output a good action for the first state-goal pair using conservative extrapolation (where each level only selects among actions its own data supports).

The second state-goal pair shares the coordinate and goal of the first, so the high-level abstraction (which ignores low-level pose information) groups them. This means that the high-level decision is rescued by the second state-goal pair: conservative extraction has the good option to reach ``a point two metres down'' available from the second state-goal pair, rather than only to stay ``on the spot'' from the first: 
\begin{equation}
    \begin{aligned}
        \phi_h(s_1, g) = \phi_h(s_2, g) = (c, g)\\
        \pi_h(\omega \mid s_1, g) = \pi_h(\omega \mid s_2, g) = \delta_{\omega = \omega_{\text{down}}}
        \notag
    \end{aligned}
\end{equation}
Note that the second pair supplies the right option, but its action is executed from a different pose, so it cannot be copied directly.

Similarly, the third state-goal pair then shares the pose and option of the first, so the low-level abstraction (which ignores high-level coordinate information) groups them. The low-level decision is then rescued by the third state-goal pair: conservative extraction has the good action $a_3$ available from the third pair, rather than only $a_1$ from the first:
\begin{equation}
    \begin{aligned}
        \phi_l(s_1, \omega_{\text{down}}) = \phi_l(s_3, \omega_{\text{down}}) = (p, \omega_{\text{down}})\\
        \pi_l(a \mid s_1, \omega_{\text{down}}) = \pi_l(a \mid s_3, \omega_{\text{down}}) = \delta_{a = a_3}
        \notag
    \end{aligned}
\end{equation}
Under the hierarchical policy, the agent therefore takes $a_3$ at $(s_1, g)$, an action never observed for that state-goal pair. 

Note that this example hinges on the learning of abstract relative options so that reach ``a point two metres down'' is a single option rather than, for example, two coordinate-specific ones.

\newpage
\section{Option Space Compression bounds Offline Error}
\label{app: compression_bounds}

\begin{proposition}{Offline error under joint state and action abstraction.}{proposition_2}
For a finite-state, finite-action MDP with at most two possible next-state transitions, with probability of at least $1 - \delta$, the suboptimality in expected return, $\epsilon := \max_{s \in \mathcal{S},g \in \mathcal{G}} \big( V^\star(s,g) - V^{\pi}(s,g) \big)$ of a goal-conditioned hierarchical policy learned from a dataset of $N$ transitions satisfies
\begin{align}
    \epsilon \propto
    \sqrt{\frac{|\mathcal{C}_h| \, |\Omega| \cdot \rho^\text{rep}_h}{(1-\gamma^k)^3 N}}
    +
    \sqrt{\frac{|\mathcal{C}_l| \, |\mathcal{A}| \, k^3 \cdot \rho^\text{rep}_l}{N}},
    \notag
\end{align}
where $\phi_h : \mathcal{S} \times \mathcal{G} \to \mathcal{C}_h$ aggregates state-goal pairs requiring the same option such that $\mathcal{C}_h$ denotes the state-abstracted high-level space, and $\phi_l : \mathcal{S} \times \Omega \to \mathcal{C}_l$ aggregates state-option pairs requiring the same action such that $\mathcal{C}_l$ denotes the state-abstracted low-level space. Constants and logarithmic factors are suppressed and $\rho^\text{rep}_h$ and $\rho^\text{rep}_l$ are the concentrability coefficients at each level:
\begin{align}
    \rho^\text{rep}_h = \sup_{c_h \in \mathcal{C}_h,\, \omega \in \Omega}
    \frac{d^{\pi_h^\star}(c_h, \omega)}{d^{\pi_h^{\mathrm{BC}}}(c_h, \omega)},
    \qquad
    \rho^\text{rep}_l = \sup_{c_l \in \mathcal{C}_l,\, a \in \mathcal{A}}
    \frac{d^{\pi_l^\star}(c_l, a)}{d^{\pi_l^{\mathrm{BC}}}(c_l, a)},
    \notag
\end{align}
where $d^\pi$ is underlying discounted occupancy measure of a policy $\pi$ and occupancy measures are aggregated over each abstraction:
\begin{align}
    d^{\pi_h}(c_h, \omega) = \sum_{(s,g) \,\in\, \phi_h^{-1}(c_h)} d^{\pi_h}(s, g, \omega),
    \qquad
    d^{\pi_l}(c_l, a) = \sum_{(s,\omega) \,\in\, \phi_l^{-1}(c_l)} d^{\pi_l}(s, \omega, a).
    \notag
\end{align}
\end{proposition}
We refer to Appendices \ref{app: sample_complexity} and \ref{app: offline_error} for full definitions and derivations. 
The two terms respectively bound the high and low-level error terms; the concentrability coefficients $\rho^\text{rep}$ measure the distribution shift between the occupancy of the optimal policy and that of the dataset: as suprema of occupancy ratios, a single state-goal or state-option pair that the optimal policy visits but the dataset does not can cause the term to go to infinity, unless aggregation lets that pair borrow mass from equivalent ones. Compressing the option space by discarding unnecessary distinctions between options reduces the bound in three ways. It reduces the high-level action space $|\Omega|$, tightens the bound $|\mathcal{C}_l| \leq |\mathcal{S}||\Omega|$ on the low-level state-abstraction space, and aggregates occupancy mass across a larger number of equivalent state-goal and state-option pairs, thereby reducing both concentrability coefficients (``conservative extrapolation''). In a continuous space, rather than making $|\Omega|$ as small as possible, this equates to mapping functionally equivalent behaviour to similar points in the option space \citep{abel2016near, ferns2012metrics, zhang2020learning}, which reduces the effective complexities of both error terms \citep{munos2008finite, chen2019information}.

\newpage
\section{State and Action Abstraction Sample Complexity in Online Goal-Conditioned RL}
\label{app: sample_complexity}
In this section, we show how joint state and action abstraction can reduce the sample complexity for finding an optimal goal-conditioned policy such that the error in expected return is smaller than a fixed constant $\epsilon$ from 
\begin{equation}
    N \propto \frac{|\mathcal{S}||\mathcal{G}||\mathcal{A}|}{(1-\gamma)^3\epsilon^2 }, 
    \notag
\end{equation}
for a policy with no state or action abstraction, to 
\begin{equation}
    N \propto \frac{|\mathcal{C}_h||\Omega|}{(1-\gamma^k)^3\epsilon^2} + \frac{|\mathcal{C}_l||\mathcal{A}|k^3}{\epsilon^2 },
    \notag
\end{equation}
for a policy with both state and action abstraction. We build on the works of \citet{kakade2003sample, munos2008finite, lattimore2012pac, robert2023sample,  li2006towards}.

\paragraph{Sample Complexity in Online GCRL} In online GCRL, the aim is to find the optimal policy with the smallest number of samples or online interactions, $N$, such that the error in optimal return is smaller than a fixed constant $\epsilon$. Consider the case of a discrete, discounted horizon MDP with a finite state space $\mathcal{S}$, action space $\mathcal{A}$ and discount factor $\gamma \in [0, 1)$. The Probably-Approximately Correct (PAC) Learning \citep{kakade2003sample} upper-bound sample complexity to find an $\epsilon$ optimal policy reaching a unique goal-state optimally (assuming at most two possible next-states for each state/action pair) with constant probability of $1 - \delta$ is given by \cite{lattimore2012pac}: 
\begin{equation}
    N \propto \frac{|\mathcal{S}||\mathcal{A}|}{(1-\gamma)^3\epsilon^2 }. 
    \notag
\end{equation}
Hence, the minimax sample complexity required to find an $\epsilon$ optimal policy reaching any given state (such that we have $|\mathcal{G}|$ unique goal-states) is given by: 
\begin{equation}
    N \propto \frac{|\mathcal{S}||\mathcal{G}||\mathcal{A}|}{(1-\gamma)^3\epsilon^2 }. 
    \label{eq: gc_sc_infinite}
\end{equation}

Consider the case of a discrete finite horizon (of length $H$) MDP with a finite state space $\mathcal{S}$, action space $\mathcal{A}$ and discount factor $\gamma \in [0, 1)$. The minimax sample complexity to find an $\epsilon$ optimal policy reaching a unique goal-state optimally is given by: 
\begin{equation}
    N \propto \frac{|\mathcal{S}||\mathcal{G}||\mathcal{A}|H^3}{\epsilon^2 }. 
    \notag
\end{equation}

\paragraph{Action Abstraction Sample Complexity in Online GCRL} Using a hierarchical policy, we can break the distant goal $g$ from our current state $s$ into options defined over an option space $\Omega$. Hence, the high-level policy becomes
\begin{equation}
    N\propto \frac{|\mathcal{S}||\mathcal{G}||\Omega|}{(1-\gamma^k)^3\epsilon^2 },
    \notag
\end{equation}
where we have substituted the option-space to Equation \ref{eq: gc_sc_infinite}, and use the environment's discount factor raised to a factor of $k$, assuming that the high-level policy acts every $k$ steps.

The low-level policy has a sample complexity of 
\begin{equation}
    N \propto \frac{|\mathcal{S}||\Omega||\mathcal{A}|k^3}{\epsilon^2 }, 
    \notag
\end{equation}
since $|\Omega|$ options can be executed. 
Hence, the overall sample complexity of the hierarchical policy is \citep{robert2023sample}: 
\begin{equation}
    N \propto \frac{|\mathcal{S}||\mathcal{G}||\Omega|}{(1-\gamma^k)^3\epsilon^2} + \frac{|\mathcal{S}||\Omega||\mathcal{A}|k^3}{\epsilon^2 }. 
    \label{eq: sc_hrl}
\end{equation}

Note that in the case of no temporal abstraction, when $k = 1$, we approximately recover the original sample complexity of a non-hierarchical policy. Since the option just becomes a single primitive action, Equation \ref{eq: sc_hrl} becomes:
\begin{equation}
    N\propto \frac{|\mathcal{S}||\mathcal{G}||\mathcal{A}|}{(1-\gamma)^3\epsilon^2} + \frac{|\mathcal{S}||\mathcal{A}|^2}{\epsilon^2 }
    \approx \frac{|\mathcal{S}||\mathcal{G}||\mathcal{A}|}{(1-\gamma)^3\epsilon^2},
    \notag
\end{equation}
where the approximation follows due to the dominating $1/(1 - \gamma)^3$ factor in the first term, assuming long-horizon problems such that $\gamma \to 1$ and that the goal-space is larger or equal to the size of the action space $|\mathcal{G}| \ge |\mathcal{A}|$. Even though this might not be the case, usually the goal-space is unknown, so we train the policy to reach any state within the state-space i.e. such that $\mathcal{G} = \mathcal{S}$. Assuming that $|\mathcal{S}| \gg |\mathcal{A}|$ is a standard assumption. 

In most real-world settings we expect $|\Omega| \ll |\mathcal{A}|^k$. The option space does not have to perfectly recreate the sequence of low-level actions, only contain the sufficient information to approximate the high-level value and choose the low-level action. 


\paragraph{State and Action Abstraction Sample Complexity in Online GCRL} Finally, since the high-level decision process can discard irrelevant low-level information, and the low-level decision process can discard irrelevant high-level information, state abstraction reduces the effective complexity of the state space at each level of the hierarchy such that Equation \ref{eq: sc_hrl} becomes: 
\begin{equation}
    N \propto \frac{|\mathcal{C}_h||\Omega|}{(1-\gamma^k)^3\epsilon^2} + \frac{|\mathcal{C}_l||\mathcal{A}|k^3}{\epsilon^2 },
    \label{eq: gc_sc_hrl}
\end{equation}
where $|\mathcal{C}_h|$ is the effective state-space required for the high-level decision process and $|\mathcal{C}_l|$ is the effective state-space for the low-level decision process. By definition, $|\mathcal{C}_h|\le |\mathcal{S}||\mathcal{G}|$ and $|\mathcal{C}_l|\le |\mathcal{S}||\Omega|$. 

\newpage
\section{State and Action Abstraction Error in Offline Goal-Conditioned RL}
\label{app: offline_error}
In offline GCRL, the aim is to bound the error $\epsilon$ given an offline dataset $\mathcal{D}$ of fixed size $N$.  

Unlike in the online setting, where it is assumed that the agent can sample any state-action pair to learn the environment's dynamics, in offline RL, a concentrability coefficient $\rho$ is incorporated to account for the distribution shift in data collected by the policy $\pi^{\text{BC}}$, and the data that would have been collected induced under the optimal policy $\pi^\star$. Intuitively, if the dataset does not include the states required to learn the optimal policy, the algorithm may never learn that optimal policy. 
The concentrability coefficient $\rho$ is defined as: 
\begin{equation}
    \rho = \sup_{s \in \mathcal{S},g \in \mathcal{G},a \in \mathcal{A}} \frac{d^{\pi^\star}(s, g,a)}{d^{\pi^{\text{BC}}}(s,g,a)},
    \notag
\end{equation}
where $d^\pi(s,g,a)$ is the discounted occupancy measure (i.e. stationary distribution) of the policy $\pi$:
\begin{equation}
    d^\pi(s,g,a) = (1-\gamma) \mathbb{E}_{\tau \sim\pi, s_0, g_0 \sim \text{Unif}(\mathcal{G})} \left[ \sum_{t=0}^{\infty} \gamma^t \mathbbm{1}(s_t = s, a_t = a, g_0 = g) \right],
    \notag
\end{equation}
and where the trajectory is induced by the MDP and following the policy $\pi( a \mid s, g)$. If the dataset is highly exploratory and covers the optimal paths well, $\rho$ will be small. If the dataset is narrow, or misses critical regions of the state-action space, $\rho$ will be large. 

Rearranging Equation \ref{eq: gc_sc_infinite} and incorporating the concentrability coefficient, the offline bound for a goal-conditioned non-hierarchical policy is given by: 
\begin{equation}
    \epsilon \propto \sqrt{\frac{|\mathcal{S}||\mathcal{G}||\mathcal{A}| \cdot \rho}{(1-\gamma)^3 N}}.
    \notag
\end{equation}

\paragraph{Action Abstraction Error in Offline GCRL} Since the size of the offline dataset is fixed, one way to reduce the error $\epsilon$ is to use a hierarchical policy with distinct state-goal representations. 

Using a hierarchical policy, the error term becomes: 
\begin{equation}
    \epsilon \propto \sqrt{\frac{|\mathcal{S}||\mathcal{G}||\Omega| \cdot \rho_h}{(1-\gamma^k)^3 N}} +  \sqrt{\frac{|\mathcal{S}||\Omega||\mathcal{A}|k^3 \cdot \rho_l}{N}},
    \notag
\end{equation}
where the concentrability coefficients are defined as: 
\begin{equation}
    \rho_h = \sup_{s \in \mathcal{S}, g \in \mathcal{G}, \omega \in \Omega} \frac{d^{\pi_h^\star}(s, g, \omega)}{d^{\pi_h^{\text{BC}}}(s,g,\omega)}     \quad \text{and} \quad 
    \rho_l = \sup_{s \in \mathcal{S}, \omega \in \Omega, a \in \mathcal{A}} \frac{d^{\pi_l^\star}(s,\omega,a)}{d^{\pi_l^{\text{BC}}}(s,\omega,a)}.
    \notag
\end{equation}

\paragraph{State and Action Abstraction Error in Offline GCRL} Then, ignoring irrelevant low-level information for the high-level decision process and high-level information for the low-level decision process, the error term is bounded by: 
\begin{equation}
    \epsilon \propto \sqrt{\frac{|\mathcal{C}_h||\Omega| \cdot \rho^{\text{rep}}_h}{(1-\gamma^k)^3 N}} +  \sqrt{\frac{|\mathcal{C}_l||\mathcal{A}|k^3 \cdot \rho^{\text{rep}}_l}{N}}. 
    \notag
\end{equation}
Again, $|\mathcal{C}_h|\le |\mathcal{S}||\mathcal{G}|$ and $|\mathcal{C}_l|\le |\mathcal{S}||\Omega|$.

Because the concentrability coefficients are now defined over the constrained spaces $\mathcal{C}_h$ and $\mathcal{C}_l$, i.e.
\begin{align}
    \rho^{\text{rep}}_h = \sup_{c_h \in \mathcal{C}_h,\, \omega \in \Omega}
    \frac{d^{\pi_h^\star}(c_h, \omega)}{d^{\pi_h^{\mathrm{BC}}}(c_h, \omega)},
    \qquad
    \rho^{\text{rep}}_l = \sup_{c_l \in \mathcal{C}_l,\, a \in \mathcal{A}}
    \frac{d^{\pi_l^\star}(c_l, a)}{d^{\pi_l^{\mathrm{BC}}}(c_l, a)},
    \notag
\end{align}
the probability mass of the offline dataset is aggregated across state-goal or state-option pairs that are equivalent under this information loss (i.e. for which $c_h = \phi_h(s,g)$ and $c_l = \phi_l(s, \omega)$):
\begin{align}
    d^{\pi_h}(c_h, \omega) = \sum_{(s,g) \,\in\, \phi_h^{-1}(c_h)} d^{\pi_h}(s, g, \omega),
    \qquad
    d^{\pi_l}(c_l, a) = \sum_{(s,\omega) \,\in\, \phi_l^{-1}(c_l)} d^{\pi_l}(s, \omega, a).
    \notag
\end{align}
This reduces the likelihood of a support mismatch, where the optimal policy requires a state transition on which the dataset places zero mass. Hence, 
\begin{equation}
    \rho^{\text{rep}}_h \le \rho_h \quad \text{and} \quad \rho^{\text{rep}}_l \le \rho_l. \notag
\end{equation}

\newpage
\section{Q-Shaped Options}
\label{app: method}

\subsection{Action Abstraction}

\paragraph{Joint Action Abstraction and Value Learning} The action abstractions are learned jointly with the high-level and low-level Q functions, acting as a goal-space for the low-level Q function and an action space for the high-level Q function.  
The high-level value and critic losses are given by the following IQL loss: 
\begin{align}
    \mathcal{L}_{V_h} &= \mathbb{E}_{(s_t, s_{t+k}, g) \sim \mathcal{D}}
    \Big[\ell^2_\nu\Big(V_h(s_t, g) - \tilde{Q}_h\big(s_t, g, \tilde\phi_\omega(s_t, s_{t+k})\big) \Big)\Big] \label{eq: high_value} \\[4pt]
    \mathcal{L}_{Q_h, \phi_\omega} &= \mathbb{E}_{(s_t, s_{t+k}, g) \sim \mathcal{D}}
    \Bigg[\Big( Q_h\big(s_t, g, \phi_\omega(s_t, s_{t+k})\big) - R_h(s_{t:t+k}, g) - \gamma V_h(s_{t+k}, g) \Big)^2\text{ }\Bigg], \label{eq: high_q_function}
\end{align}
where the normalised $k$-step option return is
\begin{equation}
    R_h(s_{t:t+k}, g) = \frac{1 - \gamma}{1 - \gamma^{k}} \sum_{t'=0}^{k-1} \gamma^{t'}\, r_h(s_{t+t'}, g).
    \notag
\end{equation}

The low-level value and critic losses are given by the following IQL loss: 
\begin{align}
        \mathcal{L}_{V_l} &= \mathbb{E}_{(s_t,a_t, g_s) \sim \mathcal{D}} \Big[\ell^2_\nu\Big(V_l(s_t, \phi_\omega(s_t,g_s)) - \tilde{Q}_l \big(s_t, \tilde\phi_\omega(s_t, g_s), a_t \big) \Big)\Big] \label{eq: low_value}\\
    \mathcal{L}_{Q_l, \phi_\omega} &= \mathbb{E}_{\substack{(s_t,a_t, s_{t+1},g_s) \sim \mathcal{D}\\ \omega_{t+1} = \phi_\omega(s_{t+1}, g_s)}} \Bigg[\Big(Q_l \big(s_t, \phi_\omega(s_t,g_s),a_t\big) - r_l(s_t, g_s) - \gamma V_l(s_{t+1}, \omega_{t+1})\Big)^2\Bigg]. \label{eq: low_q_function}
\end{align}

\paragraph{Continuous Option Space} 
We use a continuous option space encoded using a feed-forward MLP followed by length-normalisation so that the options lie on a hypersphere of radius $\sqrt{d_\omega}$ in $\mathbb{R}^{d_\omega}$, preventing any proposed option from being significantly out-of-distribution for the low-level policy: 
\begin{align}
    \phi_\omega(s, g) = \sqrt{d_\omega}\;\frac{f_\omega(s, g)}{\lVert f_\omega(s, g) \rVert_2},
    \notag
\end{align}
where $f_\omega : \mathcal{S} \times \mathcal{G} \to \mathbb{R}^{d_\omega}$ denotes the feed-forward MLP encoder. At deployment, the options are continually resampled at every timestep.

\paragraph{Discounting} 
The same discount factor $\gamma$ is used at each level of the hierarchy, so that each level has the
same effective horizon. For a two-level hierarchy, this means that the low-level decision process looks $1/(1-\gamma)$ actions ahead, and the higher-level decision process looks $1/(1-\gamma)$ options ahead. Since one high-level option corresponds to $k$ low-level actions, we would ideally set
\begin{align}
    k = \frac{1}{1 - \gamma},
    \notag
\end{align}
so that the low-level decision process reasons over exactly the span of a single option. The effective horizons then compose multiplicatively: in other words, the higher level decision process covers $k^2$ actions, while only reasoning over $1/(1-\gamma)$ decisions.

In practice (and like prior work on hierarchical RL and QC) we find that $k$ is bounded by the data. Since options are sampled as $k$-steps within a single trajectory, $k$ must be small relative to the trajectory length $H$ for the dataset to contain enough distinct options. For example, with our target discount factor of $0.99$, we would want to use $100$-step options, but antmaze-giant's trajectories are only $200$ steps. 

\paragraph{Reward Calibration}
Accumulated option rewards are normalised so that a full-length option costs the same as a single action in the low-level decision process. This means that the Q functions work on similar scales, preventing one Q function from dominating the other in magnitude (bypassing the need to introduce any hyperparameters), while options terminating early incur a proportionally smaller cost. The accumulated option reward is normalised by \begin{equation}
    \frac{1 - \gamma}{1 - \gamma^{k}},
    \notag
\end{equation} 
so that a full-length option (i.e. $\beta(s_{t+t'},g) = 0$ for $t' \in \{0, \cdots, k-1\}$) costs exactly $-1$, identical to the cost of a single action in the low-level decision process. 

\newpage
\subsection{State Abstractions}

Rather than introducing an explicit representation-learning objective, we induce meaningful state abstractions through level-specific datasets. At the high-level, we sample a broad mixture of local, trajectory, and random goals, encouraging it to generalise across a broad combinatorial coverage of state-goal pairs; at the low-level, we predominantly sample nearby trajectory target states, encouraging precise control over the option horizon. We replace the stricter reward-labelling criterion used by prior work \citep{park2025ogbench}, which requires an exact match with the goal on every state dimension, with the environment success criterion at the high-level, avoiding distinctions between states that are equivalent for the task. Note that, to isolate the effect of action abstraction and because we find that it improves all hierarchical baselines (Figure \ref{fig: ablation_reward_label_all} in Appendix \ref{app: experiments}) we apply this change consistently in our experiments. At the low-level, where we want to focus on fine-grained control, we retain the stricter criterion. We ablate using the environment criterion at both levels: in Appendix \ref{app: experiments}, Figure \ref{fig: ablation_dataset_sampling} shows performance ablating high-level goal sampling, while Figures \ref{fig: ablation_reward_label}, \ref{fig: ant_reward_option_embeddings} and \ref{fig: cube_reward_option_embeddings} show performance and option embeddings when ablating reward labelling. Table \ref{tab: state_abstraction} quantifies state abstraction.

\paragraph{Goal Sampling}
During training, goals are sampled according to the following, where $\tau = (s_0, \dots, s_{H})$ denotes the $H$-length trajectory containing $s_t$ and $\mathrm{Unif}_{\mathcal{D}}$ denotes a uniform distribution over all states in the dataset, and the weights $w$ of each of the three sampling components sum to $1$. As usual, $k$ denotes the fixed option length.

\emph{High-level Decision Process}

For the high-level decision process, goals are sampled according to
\begin{align}
    p_h(g \mid s_t)
    =
    w_h^{\mathrm{opt}}\, p^{\mathrm{opt}}_h(g \mid s_t)
    + 
    w_h^{\mathrm{traj}}\, p^{\mathrm{traj}}_h(g \mid s_t)
    +
    w_h^{\mathrm{rand}}\, \mathrm{Unif}_{\mathcal{D}}(g),
    \label{eq: high_level_goals}
\end{align}
where the first component samples a goal uniformly from within the option window,
\begin{align}
    p^{\mathrm{opt}}(g \mid s_t) = \mathcal{U}\big(\{\,\min(t+1,\,H),\; \dots,\; \min(t+k-1,\,H)\,\}\big),
    \notag
\end{align}
the second component samples the goal uniformly from the remainder of the trajectory beyond the
option window,
\begin{align}
    p^{\mathrm{traj}}_h(g \mid s_t) = \mathcal{U}\big(\{\min(t+k,\,H),\; \dots,\; H\,\}\big),
    \notag
\end{align}
and the third samples goals uniformly from the entire dataset. 

The first component ensures that a fraction of every batch returns success, the second component provides goals that are distant but reachable by dataset stitching, and the third component provides examples of the out-of-distribution combinations of state-goal pairs that the high-level policy will face at deployment.

\emph{Low-level Decision Process}
For the low-level decision process, goals are sampled according to
\begin{align}
    p_l(g_s \mid s_t)
    =
    w_l^{\mathrm{cur}}\, \delta_{s_t}(g_s)
    +
    w_l^{\mathrm{traj}}\, p^{\mathrm{traj}}_l(g_s \mid s_t)
    +
    w_l^{\mathrm{rand}}\, \mathrm{Unif}_{\mathcal{D}}(g_s),
    \label{eq: low_level_targets}
\end{align}
where the first component places all mass on the current state, the second geometrically samples a state a number of steps ahead,
\begin{align}
    p^{\mathrm{traj}}_l(g_s \mid s_t): \quad g_s = s_{\min(t + \kappa,\; H)},
    \qquad \kappa \sim \mathrm{Geom}\!\left(\tfrac{1}{k}\right),
    \notag
\end{align}
so that the expected look-ahead is $k$ steps and the third samples uniformly from the entire dataset. 

The first component again ensures that a fraction of every batch returns a success, the second component acts in accordance with the option length of the high-level decision process, and the third component calibrates the value functions across the state space, making the learned values neater to visualise, but is unnecessary for algorithm functioning, since, even at deployment, the low-level value functions are never queried at such states. 

\paragraph{Reward Labelling}
We use the environment success criterion (goal-achievement indicator) for the high-level
\begin{align}
    \beta^{h}(s,g) = \beta(s,g),
    \notag
\end{align}
specified per-environment in Appendix \ref{app: experiments}. For the low-level, we follow prior work and use a stricter success criterion that requires an exact match on every state dimension:
\begin{align}
    \beta^{l}(s, g_s) = \mathbbm{1} \{s = g_s\}.
    \notag
\end{align}

\newpage
\subsection{Policy Extraction}

The high-level policy is defined using Rejection Sampling of an advantage-weighted behaviour cloning policy, and the low-level simply as a behaviour cloning policy: 
\begin{equation}
    \begin{aligned}
    \pi_h(\omega \mid s_t,g) &:= \text{argmax}_{\omega \in \{\omega^i \}, \omega^i \sim \pi_h^\text{AWR}(\cdot \mid s_t,g)} Q_h(s_t,g,\omega^i),\\
    \pi_l(a_t \mid s_t, \omega)&:= \pi_l^\text{BC}(a_t \mid s_t, \omega)
    \end{aligned}
    \notag
\end{equation}

\paragraph{High-Level Policy Extraction} 
The high-level policy is fitted using the following advantage-weighted flow-matching loss:
\begin{equation}
    \begin{aligned}
        \mathcal{L}_{\pi_h^{\mathrm{AWR}}} &= \mathbb{E}_{\substack{(s_t, s_{t+k}, g) \sim \mathcal{D},\; z \sim \mathcal{N}(0, I_{d_\omega}) \\ u \sim \mathrm{Unif}([0,1]),\; \omega = \phi_\omega(s_t, s_{t+k}),\; \omega^u = (1-u)z + u\,\omega}} 
        \\ & \quad \quad \quad \quad \quad \quad 
        \Bigg[ \exp\!\Bigg( \frac{Q_h(s_t, g, \omega) - V_h(s_t, g)}{\alpha} \Bigg) \cdot \left\lVert v_{\pi^\text{AWR}_h}\big(s_t, g, u, \omega^u\big) - (\omega - z) \right\rVert_2^2 \Bigg].
    \end{aligned}
    \label{eq: high_policy}
\end{equation}

\paragraph{Low-Level Policy Extraction}
The low-level policy is fitted using the following behaviour-cloning flow-matching loss: 
\begin{equation}
    \mathcal{L}_{\pi^\text{BC}_l} = \mathbb{E}_{\substack{(s_t, a_t, s_{t+k}) \sim \mathcal{D},\; z \sim \mathcal{N}(0, I_d) \\ u \sim \mathrm{Unif}([0,1]),\; a^u = (1-u)z + u\,a_t}}
    \left[\left\lVert v^\text{BC}_{\pi_l}\big(s_t, \phi_\omega(s_t, s_{t+k}), u, a^u\big) - (a_t - z) \right\rVert_2^2 \right].
    \label{eq: low_policy}
\end{equation}

\newpage
\section{Experiments}
\label{app: experiments}

\subsection{Environments}

\begin{table}[H]
\centering
\setlength{\tabcolsep}{6pt}
\resizebox{\textwidth}{!}{
\begin{tabular}{lccccc}
\toprule
\textbf{Environment} & \textbf{State Dim.} & \textbf{Action Dim.} & \textbf{Data Trajectory Length} & \textbf{Dataset Size} & \textbf{Max. Episode Length}\\
\midrule
\texttt{antmaze-giant-stitch-v0} & 29 & 8 & 200 & 1M & 1000 \\
\texttt{antmaze-teleport-stitch-v0} & 29 & 8 & 200 & 1M  & 1000\\
\texttt{humanoidmaze-giant-stitch-v0} & 69 & 21 & 400 & 4M & 4000 \\
\midrule
\texttt{cube-double-play-v0} & 37 & 5 & 1000 & 1M & 500\\
\texttt{cube-triple-play-v0} & 46 & 5 & 1000 & 3M & 1000\\
\texttt{cube-quadruple-play-v0} & 55 & 5 & 1000 & 5M & 1000\\
\texttt{puzzle-3x3-play-v0} & 55 & 5 & 1000 & 1M & 500\\
\texttt{puzzle-4x4-play-v0} & 83 & 5 & 1000 & 1M & 500\\
\texttt{puzzle-4x5-play-v0} & 99 & 5 & 1000 & 3M & 1000\\
\texttt{puzzle-4x6-play-v0} & 115 & 5 & 1000 & 5M & 1000\\
\texttt{scene-play-v0} & 40 & 5 & 1000 & 1M & 750 \\
\bottomrule
\end{tabular}}
\caption{\textbf{OGBench environment and dataset characteristics.}}
\label{tab: env_properties}
\end{table}

\emph{Maze Environments}

For antmaze-giant, antmaze-teleport and humanoidmaze-giant, high-level success is defined as the agent lying within $\epsilon$ of the goal's xy coordinates, where $x(s) \in \mathbb{R}^2$ denotes a state $s$'s xy coordinates:
\begin{align}
    \beta(s,g) = \mathbbm{1}\left\{ \left\lVert x(s) - x(g) \right\rVert_2 \le \epsilon \right\}.
    \notag
\end{align}

\emph{Cube Environments}

For cube-double, cube-triple and cube-quadruple, high-level success is defined as every cube being within $\epsilon$ of its goal xyz position, where $c_i(s) \in \mathbb{R}^3$ denotes cube $i$'s xyz coordinates in state $s$, and $n$ is the number of cubes in the environment:
\begin{align}
    \beta(s,g) = \prod_{i=1}^{n} \mathbbm{1}\left\{ \left\lVert c_i(s) - c_i(g) \right\rVert_2 \le \epsilon \right\}.
    \notag
\end{align}

\emph{Puzzle Environments}

For puzzle-3x3, puzzle-4x4, puzzle-4x5 and puzzle-4x6, high-level success is defined as every button matching its goal configuration, where $b_i(s) \in \mathbb{R}^3$ denotes button $i$'s configuration in state $s$, and $n$ is the number of buttons in the environment:
\begin{align}
    \beta(s,g) = \prod_{i=1}^{n} \mathbbm{1}\left\{ b_i(s) = b_i(g) \right\}.
    \notag
\end{align}

\emph{Scene Environment}

For scene-play, high-level success is defined as the $n_c$ cubes, drawer and window being within $\epsilon_c$, $\epsilon_d$ and $\epsilon_w$ of their goal positions and the $n_b$ buttons matching their goal configurations:
\begin{align}
    \beta(s,g) =
    \prod_{i=1}^{n_c} \mathbbm{1}\left\{ \left\lVert c_i(s) - c_i(g) \right\rVert_2 \le \epsilon_c \right\}
    \cdot
    \prod_{i=1}^{n_b} \mathbbm{1}\left\{ b_i(s) =b_i(g) \right\}
    \notag \\
    \cdot\;
    \mathbbm{1}\left\{ \left\lvert d(s) - d(g) \right\rvert \le \epsilon_{\mathrm{d}} \right\}
    \cdot
    \mathbbm{1}\left\{ \left\lvert w(s) - w(g) \right\rvert \le \epsilon_{\mathrm{w}} \right\}.
    \notag
\end{align}

\newpage
\subsection{Baselines}
\subsubsection{Algorithms}
\paragraph{Absolute Subgoal Options: DSHARSA}
For DSHARSA \citep{park2026horizon}, both the high and low-level value and critic functions are learned using IQL:
\begin{equation}
    \begin{aligned}
        \mathcal{L}_{V_h} &= \mathbb{E}_{(s_t,s_{t+k}, g) \sim \mathcal{D}} \left[\ell^2_\nu\left(V_h(s_{t}, g) - \tilde{Q}_h(s_t, g, s_{t+k}) \right)\right], \\
    \mathcal{L}_{Q_h} &= \mathbb{E}_{(s_t,s_{t+k},g) \sim \mathcal{D}} \left[\left(Q_h(s_t,g,s_{t+k}) - \sum_{t'=0}^{k-1}\gamma^{t'}r_h(s_{t+t'}, g)  - \gamma^k V_h(s_{t+k}, g)\right)^2\right],
    \end{aligned}
    \notag
\end{equation}
and
\begin{equation}
    \begin{aligned}
        \mathcal{L}_{V_l} &= \mathbb{E}_{(s_t,a_t, g_s) \sim \mathcal{D}} \left[\ell^2_\nu\left(V_l(s_t,g_s) - \tilde{Q}_l(s_t, g_s, a_t) \right)\right], \\
    \mathcal{L}_{Q_l} &= \mathbb{E}_{(s_t,a_t, s_{t+1},g_s) \sim \mathcal{D}} \left[\left(Q_l(s_t,g_s,a_t) - r_l(s_t, g_s) - \gamma V_l(s_{t+1}, g_s)\right)^2\right],
    \end{aligned}
    \notag
\end{equation}
The high-level and low-level policies are fitted using behaviour-cloning flow-matching losses: 
\begin{equation}
    \begin{aligned}
        \mathcal{L}_{\pi_h^{\mathrm{BC}}} &= \mathbb{E}_{\substack{(s_t, s_{t+k}, g) \sim \mathcal{D},\; z \sim \mathcal{N}(0, I_{d_s}) \\ u \sim \mathrm{Unif}([0,1]),\; g_s = s_{t+k},\; g_s^u = (1-u)z + u\,g_s}} 
        \Bigg[  \left\lVert v_{\pi^\text{BC}_h}\big(s_t, g, u, g_s^u\big) - (g_s - z) \right\rVert_2^2 \Bigg],
    \end{aligned}
    \notag
\end{equation}
and 
\begin{equation}
    \mathcal{L}_{\pi^\text{BC}_l} = \mathbb{E}_{\substack{(s_t, a_t, s_{t+k}) \sim \mathcal{D},\; z \sim \mathcal{N}(0, I_d) \\ u \sim \mathrm{Unif}([0,1]),\; a^u = (1-u)z + u\,a_t}}
    \Bigg[\left\lVert v_{\pi_l}\big(s_t, s_{t+k}, u, a^u\big) - (a_t - z) \right\rVert_2^2 \Bigg].
    \notag
\end{equation}
At deployment, both the high and low-level policies are defined with Rejection Sampling: 
\begin{equation}
    \begin{aligned}
        \pi_h(g_s \mid s_t, g) := \text{argmax}_{g_s \in \{g_s^i \}, g_s^i \sim \pi^{\text{BC}}_h(\cdot \mid s_t,g)} Q_h(s_t, g, g_s^i),\\
        \pi_l(a_t \mid s_t, g_s) :=  \text{argmax}_{a_t \in \{a_t^i \}, a_t^i \sim \pi^{\text{BC}}_l(\cdot \mid s_t,g_s)} Q_l(s_t, g_s, a_t^i).
    \end{aligned}
    \notag
\end{equation}

\paragraph{Absolute Subgoal Options: SHARSA}
SHARSA \citep{park2026horizon} is almost identical to DSHARSA, but only uses Rejection Sampling for the high-level policy. The low-level policy is defined using pure behaviour cloning:
\begin{equation}
    \begin{aligned}
        \pi_h(g_s \mid s_t, g) &:= \text{argmax}_{g_s \in \{g_s^i \}, g_s^i \sim \pi^{\text{BC}}_h(\cdot \mid s_t,g)} Q_h(s_t, g, g_s^i),\\
        \pi_l(a_t \mid s_t, g_s) &:=  \pi^{\text{BC}}_l(a_t \mid s_t,g_s).
    \end{aligned}
    \notag
\end{equation}
Note that conservative extrapolation is still possible as the low-level policy might be queried at an absolute subgoal option (proposed by the high-level policy) that was unseen during training. 

\paragraph{Absolute Subgoal Options: ASO}
As explained in Section \ref{sec: method}, for QSO we use advantage-weighting to train the high-level policy. Therefore, as a baseline, we introduce a new variant of SHARSA that, like QSO, uses AWR to train the high-level policy:
\begin{equation}
    \begin{aligned}
        \mathcal{L}_{\pi_h^{\mathrm{AWR}}} &= \mathbb{E}_{\substack{(s_t, s_{t+k}, g) \sim \mathcal{D},\; z \sim \mathcal{N}(0, I_{d_s}) \\ u \sim \mathrm{Unif}([0,1]),\; g_s = s_{t+k},\; g_s^u = (1-u)z + u\,g_s}}
        \\ & \quad \quad \quad \quad \quad \quad 
        \Bigg[ \exp\!\Bigg( \frac{Q_h(s_t, g, g_s) - V_h(s_t, g)}{\alpha} \Bigg) \cdot \left\lVert v_{\pi^\text{AWR}_h}\big(s_t, g, u, g_s^u\big) - (g_s - z) \right\rVert_2^2 \Bigg].
    \end{aligned}
    \notag
\end{equation}
At deployment, the high-level policy is still defined using Rejection Sampling and the low-level with behaviour cloning:
\begin{equation}
    \begin{aligned}
        \pi_h(g_s \mid s_t, g) &:= \text{argmax}_{g_s \in \{g_s^i \}, g_s^i \sim \pi^{\text{AWR}}_h(\cdot \mid s_t,g)} Q_h(s_t, g, g_s^i),\\
        \pi_l(a_t \mid s_t, g_s) &:=  \pi^{\text{BC}}_l(a_t \mid s_t,g_s).
    \end{aligned}
    \notag
\end{equation}

\paragraph{Abstract Relative Options: HIQL}
HIQL \citep{park2023hiql} shapes the abstract subgoal option space through its single, action-free state-value function $V(s, \phi_\omega(s,g))$, which is trained using Implicit V-Learning (IVL) \citep{park2025ogbench}:
\begin{equation}
    \mathcal{L}_{V, \phi_\omega} = \mathbb{E}_{(s_t,a_t, g) \sim \mathcal{D}, g \sim p^\mathcal{D}(\cdot \mid s, a)} \left[\ell^2_\nu\left(V(s_t,\phi_\omega(s_t,g)) - r(s_t, g) - \gamma \tilde{V}(s_{t+1},\tilde{\phi}_\omega(s_{t+1},g)) \right)\right],
    \notag
\end{equation}
where $\tilde{V}$ and $\tilde{{\phi}}_\omega$ denote the target network and representations and $\ell^2_\nu$ the expectile loss. HIQL extracts high and low-level Gaussian policies using AWR-like objectives:
\begin{equation}
    \begin{aligned}
        \mathcal{L}_{\pi_h} = -\mathbb{E}_{(s_t,s_{t+k}, g) \sim \mathcal{D}, \omega = \phi_\omega(s_t, s_{t+k})} \left[\exp \left({\frac{V(s_{t+k},g)- V(s_t,g)}{\alpha}} \right)\cdot \log \pi_h(\omega\mid s_t, g)\right],\\
        \mathcal{L}_{\pi_l} = -\mathbb{E}_{(s_t,s_{t+k}, g) \sim \mathcal{D}, \omega = \phi_\omega(s_t, s_{t+k})} \left[\exp \left({\frac{V(s_{t+1},s_{t+k})- V(s_{t},s_{t+k})}{\alpha}}\right) \cdot \log \pi_l(a_t\mid s_t, \omega)\right].
    \end{aligned}
    \notag
\end{equation}

\paragraph{Abstract Options: OPAL}
OPAL \citep{ajay2020opal} shapes the abstract subgoal option space using a $\beta$-VAE trained to encode state-action trajectory sequences:
\begin{equation}
    \min_{\pi_l, p_\omega, q_\omega} -\mathbb{E}_{\tau\sim \mathcal{D}, \omega \sim q_\omega(\cdot \mid \tau)}\left[\sum_{t=0}^{k-1} \log\pi^\text{BC}_l(a_t \mid s_t, \omega)\right] + \beta \cdot \text{D}_{\text{KL}}(q_\omega(\omega \mid \tau) \mid \mid p_\omega(\omega \mid s_0)),
    \notag
\end{equation}
where $q_\omega$ and $p_\omega$ respectively denote the option-space encoder and option-space prior. Using this learned option space to encode trajectory sequences, state-value and Q functions are trained using IQL: 
\begin{equation}
    \begin{aligned}
        \mathcal{L}_{V} &= \mathbb{E}_{(\tau, g) \sim \mathcal{D}, \omega \sim q_\omega(\cdot \mid \tau)} \left[\ell^2_\nu\left(V(s_t,g) - \tilde{Q}(s_t, g, \omega) \right)\right], \\
    \mathcal{L}_{Q} &= \mathbb{E}_{(\tau,g) \sim \mathcal{D}, \omega \sim q_\omega(\cdot \mid \tau)} \left[\left(Q(s_t,g,\omega) - \sum_{t'=0}^{k-1}\gamma^{t'}r(s_{t+t'}, g)  - \gamma^k V(s_{t+k}, g)\right)^2\right],
    \notag
    \end{aligned}
\end{equation}
and a high-level flow-matching policy is fitted using the following behaviour-cloning flow-matching loss: 
\begin{equation}
    \begin{aligned}
        \mathcal{L}_{\pi_h^{\mathrm{BC}}} &= \mathbb{E}_{\substack{(\tau, g) \sim \mathcal{D},\; z \sim \mathcal{N}(0, I_{d_\omega}) \\ u \sim \mathrm{Unif}([0,1]),\; \omega \sim q_\omega(\cdot \mid \tau),\; \omega^u = (1-u)z + u\,\omega}} 
        \Bigg[  \left\lVert v_{\pi^\text{BC}_h}\big(s_t, g, u, \omega^u\big) - (\omega - z) \right\rVert_2^2 \Bigg],
    \end{aligned}
    \notag
\end{equation}
At deployment, the high-level policy is defined using Rejection Sampling, and the low-level with behaviour cloning:
\begin{equation}
    \begin{aligned}
        \pi_h(\omega \mid s_t, g) &:= \text{argmax}_{\omega \in \{\omega^i \}, \omega^i \sim \pi^{\text{BC}}_h(\cdot \mid s_t,g)} Q_h(s_t, g, \omega^i),\\
        \pi_l(a_t \mid s_t, \omega) &:= \pi_l^\text{BC}(a_t \mid s_t, \omega).
        \notag
    \end{aligned}
\end{equation}
In practice, the low-level policy is finetuned to output the immediate action. 

\paragraph{Non-hierarchical: IQL} IQL \citep{kostrikov2021offline} trains a value and Q function with the following losses: 
\begin{equation}
    \begin{aligned}
    \mathcal{L}_{V} &= \mathbb{E}_{(s_t,a_t, g) \sim \mathcal{D}} \left[\ell^2_\nu\left(V(s_t,g) - \tilde{Q}(s_t, g, a_t) \right)\right], \\
    \mathcal{L}_{Q} &= \mathbb{E}_{(s_t,a_t, s_{t+1},g) \sim \mathcal{D}} \left[\left(Q(s_t,g,a_t) - r(s_{t}, g) - \gamma V(s_{t+1}, g)\right)^2\right],
    \end{aligned}
    \notag
\end{equation}
We then extract a Gaussian policy by maximising the Deep Deterministic Policy Gradient objective \citep{fujimoto2021minimalist}: 
\begin{equation}
    \mathcal{L}_\pi = - \mathbb{E}_{(s_t,a_t,g) \sim \mathcal{D}} \Bigg[Q\big(s, g, \mu_{\pi}(s,g)\big) + \alpha \log \pi (a \mid s,g) \Bigg], 
    \notag
\end{equation}
where $\mu_\pi$ denotes the mean of the Gaussian policy. 
Following \citet{park2024value}, we use DDPGBC rather than AWR for policy extraction. 

\paragraph{Action Chunking: QC}
QC \citep{li2026reinforcement} trains an action-chunked Q function: 
\begin{equation}
    \begin{aligned}
    \mathcal{L}_{Q} &= \mathbb{E}_{\substack{(s_t,a_{t:t+k},s_{t+k},g) \sim \mathcal{D} \\ a_{t+k: t+2k} \sim \pi(\cdot \mid s_{t+k}, g)}} \Bigg[\Big(Q \big(s_t, g, a_{t:t+k}\big) - \sum_{t'=0}^{k-1}\gamma^{t'}r(s_{t+t'}, g) - \gamma^k Q\big(s_{t+k},g, a_{t+k: t+2k}\big)\Big)^2\Bigg].
    \end{aligned}
    \notag
\end{equation} 
It fits a flow-matching policy using behaviour-cloning:
\begin{equation}
    \mathcal{L}_{\pi^\text{BC}} = \mathbb{E}_{\substack{(s_t, a_{t:t+k}, g) \sim \mathcal{D},\; z \sim \mathcal{N}(0, I_d) \\ u \sim \mathrm{Unif}([0,1]),\; a^u = (1-u)z + u\,a_{t:t+k}}}
    \Bigg[\left\lVert v_{\pi_l}\big(s_t, g, u, a^u\big) - (a_{t:t+k} - z) \right\rVert_2^2 \Bigg].
    \notag
\end{equation}
The policy is defined using Rejection Sampling: 
\begin{equation}
    \pi(a_{t:t+k} \mid s_t, g) := \text{argmax}_{a_{t:t+k} \in \{a_{t:t+k}^i \}, \{a_{t:t+k}\}^i \sim \pi^\text{BC}(\cdot \mid s_t, g)} Q\big(s_t, g, a^i_{t:t+k}\big).
    \notag
\end{equation}

\subsubsection{Dataset Sampling for Non-Hierarchical and Action Chunking Baselines}

The dataset sampling intends to mirror the hierarchical ones as closely as possible, to enable the isolation of the hierarchical option space on performance as opposed to dataset sampling. 

\paragraph{Non-hierarchical: IQL}
The flat baseline has no option structure, so there is no distinction between a near and a distant goal and the option window collapses to the current state. Goals are sampled according to:
\begin{align}
    p_{\mathrm{flat}}(g \mid s_t)
    =
    w^{\mathrm{cur}}\, \delta_{s_t}(g)
    +
    w^{\mathrm{traj}}\, p^{\mathrm{traj}}_{\mathrm{flat}}(g \mid s_t)
    +
    w^{\mathrm{rand}}\, \mathrm{Unif}_{\mathcal{D}}(g),
    \notag
\end{align}
where the second component samples the goal uniformly from the remainder of the trajectory beyond the current state,
\begin{align}
    p^{\mathrm{traj}}_{\mathrm{flat}}(g \mid s_t) = \mathcal{U}\big(\{\,\min(t+1,\,H),\; \dots,\; H\,\}\big).
    \notag
\end{align}
and, like for hierarchical high-level dataset sampling, the third component samples goals uniformly from the entire dataset. We use the same weighting as for the hierarchical high-level dataset sampling as opposed to the low-level one, since the single non-hierarchical policy must cover the full range of state-goal possibilities with one distribution, in contrast to the hierarchical case where the near and far regimes can be split across two levels.

\paragraph{Action Chunking: QC}
The action chunking baseline is flat but temporally extended, so it inherits the high-level goal distribution of Equation \ref{eq: high_level_goals} and additionally requires the length-$k$ action sequence beginning at $s_t$,
\begin{align}
    a_{t:t+k} = \big(a_{\min(t,\,H)},\; \dots,\; a_{\min(t+k-1,\,H)}\big),
    \notag
\end{align}
where indices are clipped to the end of the trajectory. If an agent reaches its target state before the end of the $k$ length option chunk, actions past the target state are masked out and set to zero, so that the critic is not trained on behaviour occurring past termination.

Following original implementations of non-hierarchical methods, we use strict reward labelling. 

\newpage
\subsection{Hyperparameters}
We use the following default hyperparameters for all value networks (both Value and Q functions), flow-matching and Gaussian policies, and Rejection Sampling methods. Default learning rate, batch size, target update rate, layer normalisation, non-linearities, and representation latent dimensions follow \cite{park2025ogbench}. According to \citet{park2023hiql}, we use $128$ for HIQL. Rejection Sampling $N$ follows \citet{park2026horizon}. Default network sizes and actor flow steps were chosen according to available compute time and memory. For Gaussian policies, following \cite{park2023hiql}, we use a network that outputs a mean and constant standard deviation. We use the number of steps from \citet{park2023hiql}, apart from for QC, where we use their suggestion of $5$. 

\begin{table}[H]
\footnotesize
\caption{Hyperparameters.}
\label{tab: hyperparameters}
\begin{center}
\begin{tabular}{ll}
\toprule
\multicolumn{1}{l}{\bf Hyperparameter}  &\multicolumn{1}{l}{\bf Value} \\
\midrule
Optimiser & Adam \citep{kingma2014adam} \\
Learning Rate & 0.0003 \\
Batch Size &  1024 \\
Target Update Rate & 0.005 \\
Representation Latent Dimension & 10 (default), 128 (HIQL)\\
Representation Layer Normalisation & True \\
Representation Non-linearity & GELU \citep{hendrycks2016gaussian} \\ 
Representation MLP & [512, 512, 512] \\
Value Layer Normalisation & True \\
Value Non-linearity & GELU \citep{hendrycks2016gaussian} \\ 
Value MLP &  [512, 512, 512, 512] \\
Value Expectile $\nu$ (IQL) & 0.7 \\
Actor Layer Normalisation & True \\
Actor Non-linearity & GELU \citep{hendrycks2016gaussian} \\ 
Actor MLP &  [512, 512, 512, 512] \\
Actor Flow Steps &  6 \\
Actor Rejection Sampling $N$ &  32 \\
Actor AWR $\alpha$ & 3.0 \\
Actor DDPGBC $\alpha$ & 1.0 (IQL default), 0.3 (IQL ant), 0.1 (IQL humanoid) \\
Subgoal Steps $k$ & 25 (default), 100 (humanoid), 5 (QC) \\
Hierarchical Goal Sampling $(w_h^{\mathrm{opt}}, w_h^{\mathrm{traj}}, w_h^{\mathrm{rand}})$ &  (0.1, 0.5, 0.4)\\
Hierarchical Goal Sampling $(w_l^{\mathrm{cur}}, w_l^{\mathrm{traj}}, w_l^{\mathrm{rand}})$ &  (0.1, 0.85, 0.05)\\
Non-hierarchical Goal Sampling $(w^{\mathrm{cur}}, w^{\mathrm{traj}}, w^{\mathrm{rand}})$ &  (0.1, 0.5, 0.4)\\
Discount $\gamma$ & 0.99 (default), 0.995/0.999 (HIQL \& IQL ant/humanoid) \\ 
\bottomrule \\
\end{tabular}
\end{center}
\end{table}

\subsection{Further Results}

\begin{figure}[H]
    \centering
    \includegraphics[width=1.0\linewidth]{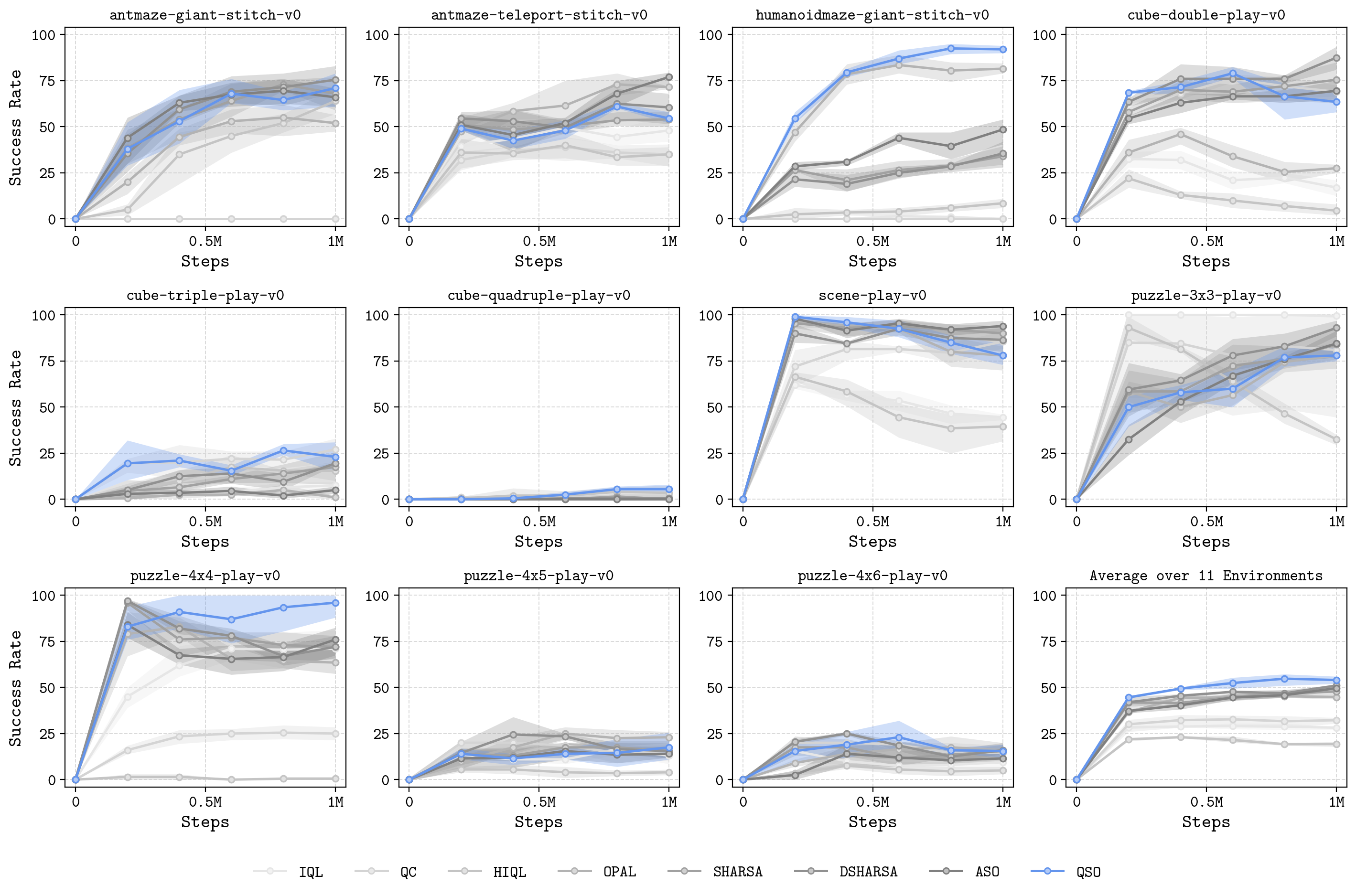}
    \caption{\textbf{Training on selected OGBench environments using the default datasets.} Success rate averaged over 5 evaluation tasks. Mean and 95\% bootstrap confidence interval over 4 seeds.}
    \label{fig: comparison}
\end{figure}

\begin{table}[t]
\centering
\providecommand{\resultcell}[1]{\rule{0pt}{3.2ex}#1\rule[-1.4ex]{0pt}{0pt}}
\renewcommand{\arraystretch}{1.0}
\setlength{\tabcolsep}{4pt}
\resizebox{\textwidth}{!}{\ttfamily
\begin{tabular}{llcc>{\columncolor[HTML]{F2F2F2}}c>{\columncolor[HTML]{F2F2F2}}c>{\columncolor[HTML]{F2F2F2}}c>{\columncolor[HTML]{F2F2F2}}c>{\columncolor[HTML]{F2F2F2}}c>{\columncolor[HTML]{EDF2FE}}c}
\toprule
 &  & \multicolumn{2}{c}{Non-Hierarchical} & \multicolumn{6}{c}{Hierarchical} \\
\cmidrule(lr){3-4} \cmidrule(lr){5-10}
Environment & Task & IQL & QC & HIQL & OPAL & SHARSA & DSHARSA & ASO & \textcolor[HTML]{6495ED}{QSO} \\
\midrule
\multirow{6}{*}{antmaze-giant-stitch-v0} & Task 1 & \resultcell{0 {\footnotesize [0, 0]}} & \resultcell{0 {\footnotesize [0, 0]}} & \resultcell{35 {\footnotesize [29, 39]}} & \resultcell{35 {\footnotesize [33, 37]}} & \resultcell{59 {\footnotesize [42, 71]}} & \resultcell{\textcolor[HTML]{6495ED}{\textbf{72}} {\footnotesize [63, 86]}} & \resultcell{57 {\footnotesize [47, 67]}} & \resultcell{57 {\footnotesize [50, 63]}} \\
 & Task 2 & \resultcell{0 {\footnotesize [0, 0]}} & \resultcell{0 {\footnotesize [0, 0]}} & \resultcell{66 {\footnotesize [60, 70]}} & \resultcell{\textcolor[HTML]{6495ED}{\textbf{83}} {\footnotesize [77, 90]}} & \resultcell{74 {\footnotesize [65, 87]}} & \resultcell{67 {\footnotesize [60, 73]}} & \resultcell{67 {\footnotesize [64, 69]}} & \resultcell{75 {\footnotesize [69, 79]}} \\
 & Task 3 & \resultcell{0 {\footnotesize [0, 0]}} & \resultcell{0 {\footnotesize [0, 0]}} & \resultcell{42 {\footnotesize [38, 45]}} & \resultcell{62 {\footnotesize [52, 71]}} & \resultcell{53 {\footnotesize [39, 66]}} & \resultcell{72 {\footnotesize [66, 78]}} & \resultcell{66 {\footnotesize [58, 72]}} & \resultcell{\textcolor[HTML]{6495ED}{\textbf{73}} {\footnotesize [66, 81]}} \\
 & Task 4 & \resultcell{0 {\footnotesize [0, 0]}} & \resultcell{0 {\footnotesize [0, 0]}} & \resultcell{57 {\footnotesize [50, 62]}} & \resultcell{17 {\footnotesize [12, 22]}} & \resultcell{65 {\footnotesize [57, 72]}} & \resultcell{\textcolor[HTML]{6495ED}{\textbf{67}} {\footnotesize [57, 77]}} & \resultcell{65 {\footnotesize [60, 72]}} & \resultcell{52 {\footnotesize [43, 61]}} \\
 & Task 5 & \resultcell{0 {\footnotesize [0, 0]}} & \resultcell{0 {\footnotesize [0, 0]}} & \resultcell{69 {\footnotesize [53, 86]}} & \resultcell{69 {\footnotesize [65, 74]}} & \resultcell{\textcolor[HTML]{6495ED}{\textbf{90}} {\footnotesize [85, 95]}} & \resultcell{82 {\footnotesize [80, 85]}} & \resultcell{84 {\footnotesize [70, 94]}} & \resultcell{82 {\footnotesize [78, 87]}} \\
\cmidrule(lr){2-10}
 & Overall & \resultcell{0 {\footnotesize [0, 0]}} & \resultcell{0 {\footnotesize [0, 0]}} & \resultcell{54 {\footnotesize [48, 60]}} & \resultcell{53 {\footnotesize [52, 55]}} & \resultcell{68 {\footnotesize [61, 76]}} & \resultcell{\textcolor[HTML]{6495ED}{\textbf{72}} {\footnotesize [67, 77]}} & \resultcell{68 {\footnotesize [66, 70]}} & \resultcell{68 {\footnotesize [66, 70]}} \\
\midrule
\multirow{6}{*}{antmaze-teleport-stitch-v0} & Task 1 & \resultcell{35 {\footnotesize [27, 44]}} & \resultcell{27 {\footnotesize [18, 33]}} & \resultcell{44 {\footnotesize [40, 48]}} & \resultcell{\textcolor[HTML]{6495ED}{\textbf{61}} {\footnotesize [50, 72]}} & \resultcell{18 {\footnotesize [12, 28]}} & \resultcell{28 {\footnotesize [17, 38]}} & \resultcell{54 {\footnotesize [46, 65]}} & \resultcell{34 {\footnotesize [28, 40]}} \\
 & Task 2 & \resultcell{47 {\footnotesize [42, 52]}} & \resultcell{38 {\footnotesize [32, 43]}} & \resultcell{44 {\footnotesize [41, 48]}} & \resultcell{85 {\footnotesize [81, 91]}} & \resultcell{77 {\footnotesize [70, 83]}} & \resultcell{\textcolor[HTML]{6495ED}{\textbf{87}} {\footnotesize [83, 93]}} & \resultcell{75 {\footnotesize [70, 80]}} & \resultcell{75 {\footnotesize [73, 78]}} \\
 & Task 3 & \resultcell{57 {\footnotesize [52, 62]}} & \resultcell{40 {\footnotesize [33, 47]}} & \resultcell{34 {\footnotesize [27, 42]}} & \resultcell{\textcolor[HTML]{6495ED}{\textbf{87}} {\footnotesize [82, 90]}} & \resultcell{80 {\footnotesize [76, 83]}} & \resultcell{74 {\footnotesize [71, 78]}} & \resultcell{82 {\footnotesize [78, 88]}} & \resultcell{68 {\footnotesize [53, 79]}} \\
 & Task 4 & \resultcell{51 {\footnotesize [40, 60]}} & \resultcell{42 {\footnotesize [36, 50]}} & \resultcell{26 {\footnotesize [13, 41]}} & \resultcell{50 {\footnotesize [47, 54]}} & \resultcell{55 {\footnotesize [51, 61]}} & \resultcell{51 {\footnotesize [45, 58]}} & \resultcell{\textcolor[HTML]{6495ED}{\textbf{59}} {\footnotesize [51, 65]}} & \resultcell{50 {\footnotesize [45, 57]}} \\
 & Task 5 & \resultcell{48 {\footnotesize [38, 57]}} & \resultcell{37 {\footnotesize [27, 48]}} & \resultcell{32 {\footnotesize [26, 38]}} & \resultcell{\textcolor[HTML]{6495ED}{\textbf{61}} {\footnotesize [52, 71]}} & \resultcell{33 {\footnotesize [26, 42]}} & \resultcell{48 {\footnotesize [38, 58]}} & \resultcell{58 {\footnotesize [50, 67]}} & \resultcell{46 {\footnotesize [43, 48]}} \\
\cmidrule(lr){2-10}
 & Overall & \resultcell{47 {\footnotesize [44, 51]}} & \resultcell{37 {\footnotesize [32, 40]}} & \resultcell{36 {\footnotesize [34, 39]}} & \resultcell{\textcolor[HTML]{6495ED}{\textbf{69}} {\footnotesize [64, 74]}} & \resultcell{53 {\footnotesize [52, 54]}} & \resultcell{58 {\footnotesize [55, 62]}} & \resultcell{66 {\footnotesize [62, 69]}} & \resultcell{54 {\footnotesize [52, 57]}} \\
\midrule
\multirow{6}{*}{humanoidmaze-giant-stitch-v0} & Task 1 & \resultcell{0 {\footnotesize [0, 0]}} & \resultcell{0 {\footnotesize [0, 0]}} & \resultcell{5 {\footnotesize [0, 10]}} & \resultcell{73 {\footnotesize [63, 83]}} & \resultcell{12 {\footnotesize [8, 18]}} & \resultcell{17 {\footnotesize [12, 22]}} & \resultcell{22 {\footnotesize [18, 28]}} & \resultcell{\textcolor[HTML]{6495ED}{\textbf{82}} {\footnotesize [77, 89]}} \\
 & Task 2 & \resultcell{3 {\footnotesize [1, 6]}} & \resultcell{0 {\footnotesize [0, 0]}} & \resultcell{18 {\footnotesize [8, 27]}} & \resultcell{78 {\footnotesize [65, 89]}} & \resultcell{41 {\footnotesize [33, 51]}} & \resultcell{47 {\footnotesize [32, 69]}} & \resultcell{32 {\footnotesize [18, 43]}} & \resultcell{\textcolor[HTML]{6495ED}{\textbf{98}} {\footnotesize [95, 100]}} \\
 & Task 3 & \resultcell{1 {\footnotesize [0, 2]}} & \resultcell{0 {\footnotesize [0, 0]}} & \resultcell{6 {\footnotesize [4, 7]}} & \resultcell{85 {\footnotesize [72, 92]}} & \resultcell{42 {\footnotesize [28, 61]}} & \resultcell{61 {\footnotesize [58, 65]}} & \resultcell{55 {\footnotesize [48, 62]}} & \resultcell{\textcolor[HTML]{6495ED}{\textbf{91}} {\footnotesize [88, 93]}} \\
 & Task 4 & \resultcell{0 {\footnotesize [0, 0]}} & \resultcell{0 {\footnotesize [0, 0]}} & \resultcell{2 {\footnotesize [0, 3]}} & \resultcell{78 {\footnotesize [73, 83]}} & \resultcell{11 {\footnotesize [10, 12]}} & \resultcell{8 {\footnotesize [3, 12]}} & \resultcell{19 {\footnotesize [17, 22]}} & \resultcell{\textcolor[HTML]{6495ED}{\textbf{85}} {\footnotesize [82, 87]}} \\
 & Task 5 & \resultcell{0 {\footnotesize [0, 0]}} & \resultcell{0 {\footnotesize [0, 0]}} & \resultcell{1 {\footnotesize [0, 2]}} & \resultcell{95 {\footnotesize [92, 98]}} & \resultcell{43 {\footnotesize [32, 61]}} & \resultcell{17 {\footnotesize [5, 28]}} & \resultcell{92 {\footnotesize [84, 98]}} & \resultcell{\textcolor[HTML]{6495ED}{\textbf{97}} {\footnotesize [94, 99]}} \\
\cmidrule(lr){2-10}
 & Overall & \resultcell{1 {\footnotesize [1, 1]}} & \resultcell{0 {\footnotesize [0, 0]}} & \resultcell{6 {\footnotesize [4, 8]}} & \resultcell{82 {\footnotesize [78, 85]}} & \resultcell{30 {\footnotesize [27, 33]}} & \resultcell{30 {\footnotesize [27, 32]}} & \resultcell{44 {\footnotesize [38, 48]}} & \resultcell{\textcolor[HTML]{6495ED}{\textbf{90}} {\footnotesize [88, 92]}} \\
\bottomrule
\end{tabular}}
\caption{\textbf{Full results on selected OGBench locomotion environments using default datasets.} Success rate (\%) on each evaluation task after 1M offline training steps. Mean and 95\% bootstrap confidence interval over 4 seeds.}
\label{tab: results_appendix_locomotion}
\end{table}

\begin{table}[H]
\centering
\providecommand{\resultcell}[1]{\rule{0pt}{3.2ex}#1\rule[-1.4ex]{0pt}{0pt}}
\renewcommand{\arraystretch}{1.0}
\setlength{\tabcolsep}{4pt}
\resizebox{\textwidth}{!}{\ttfamily
\begin{tabular}{llcc>{\columncolor[HTML]{F2F2F2}}c>{\columncolor[HTML]{F2F2F2}}c>{\columncolor[HTML]{F2F2F2}}c>{\columncolor[HTML]{F2F2F2}}c>{\columncolor[HTML]{F2F2F2}}c>{\columncolor[HTML]{EDF2FE}}c}
\toprule
 &  & \multicolumn{2}{c}{Non-Hierarchical} & \multicolumn{6}{c}{Hierarchical} \\
\cmidrule(lr){3-4} \cmidrule(lr){5-10}
Environment & Task & IQL & QC & HIQL & OPAL & SHARSA & DSHARSA & ASO & \textcolor[HTML]{6495ED}{QSO} \\
\midrule
\multirow{6}{*}{cube-double-play-v0} & Task 1 & \resultcell{62 {\footnotesize [53, 70]}} & \resultcell{90 {\footnotesize [86, 93]}} & \resultcell{26 {\footnotesize [20, 32]}} & \resultcell{82 {\footnotesize [78, 87]}} & \resultcell{95 {\footnotesize [92, 98]}} & \resultcell{98 {\footnotesize [97, 100]}} & \resultcell{96 {\footnotesize [92, 99]}} & \resultcell{\textcolor[HTML]{6495ED}{\textbf{99}} {\footnotesize [98, 100]}} \\
 & Task 2 & \resultcell{11 {\footnotesize [3, 18]}} & \resultcell{88 {\footnotesize [83, 96]}} & \resultcell{2 {\footnotesize [0, 5]}} & \resultcell{12 {\footnotesize [4, 18]}} & \resultcell{82 {\footnotesize [78, 86]}} & \resultcell{\textcolor[HTML]{6495ED}{\textbf{93}} {\footnotesize [91, 96]}} & \resultcell{88 {\footnotesize [87, 90]}} & \resultcell{68 {\footnotesize [53, 78]}} \\
 & Task 3 & \resultcell{20 {\footnotesize [16, 23]}} & \resultcell{88 {\footnotesize [85, 90]}} & \resultcell{4 {\footnotesize [1, 8]}} & \resultcell{11 {\footnotesize [6, 15]}} & \resultcell{89 {\footnotesize [87, 92]}} & \resultcell{\textcolor[HTML]{6495ED}{\textbf{91}} {\footnotesize [86, 95]}} & \resultcell{85 {\footnotesize [83, 88]}} & \resultcell{68 {\footnotesize [62, 76]}} \\
 & Task 4 & \resultcell{2 {\footnotesize [0, 5]}} & \resultcell{42 {\footnotesize [38, 45]}} & \resultcell{2 {\footnotesize [0, 3]}} & \resultcell{15 {\footnotesize [8, 22]}} & \resultcell{41 {\footnotesize [33, 48]}} & \resultcell{\textcolor[HTML]{6495ED}{\textbf{59}} {\footnotesize [48, 68]}} & \resultcell{20 {\footnotesize [13, 27]}} & \resultcell{53 {\footnotesize [50, 57]}} \\
 & Task 5 & \resultcell{6 {\footnotesize [3, 8]}} & \resultcell{54 {\footnotesize [45, 63]}} & \resultcell{2 {\footnotesize [0, 3]}} & \resultcell{25 {\footnotesize [18, 32]}} & \resultcell{53 {\footnotesize [43, 67]}} & \resultcell{58 {\footnotesize [50, 64]}} & \resultcell{48 {\footnotesize [40, 57]}} & \resultcell{\textcolor[HTML]{6495ED}{\textbf{61}} {\footnotesize [55, 69]}} \\
\cmidrule(lr){2-10}
 & Overall & \resultcell{20 {\footnotesize [17, 23]}} & \resultcell{72 {\footnotesize [70, 74]}} & \resultcell{7 {\footnotesize [5, 9]}} & \resultcell{29 {\footnotesize [26, 33]}} & \resultcell{72 {\footnotesize [68, 77]}} & \resultcell{\textcolor[HTML]{6495ED}{\textbf{80}} {\footnotesize [75, 84]}} & \resultcell{68 {\footnotesize [65, 70]}} & \resultcell{70 {\footnotesize [65, 74]}} \\
\midrule
\multirow{6}{*}{cube-triple-play-v0} & Task 1 & \resultcell{29 {\footnotesize [27, 32]}} & \resultcell{57 {\footnotesize [48, 68]}} & \resultcell{\textcolor[HTML]{6495ED}{\textbf{58}} {\footnotesize [51, 65]}} & \resultcell{8 {\footnotesize [5, 12]}} & \resultcell{45 {\footnotesize [29, 58]}} & \resultcell{44 {\footnotesize [40, 48]}} & \resultcell{16 {\footnotesize [13, 18]}} & \resultcell{45 {\footnotesize [42, 48]}} \\
 & Task 2 & \resultcell{1 {\footnotesize [0, 2]}} & \resultcell{\textcolor[HTML]{6495ED}{\textbf{31}} {\footnotesize [18, 42]}} & \resultcell{7 {\footnotesize [7, 7]}} & \resultcell{1 {\footnotesize [0, 2]}} & \resultcell{15 {\footnotesize [11, 21]}} & \resultcell{16 {\footnotesize [11, 23]}} & \resultcell{2 {\footnotesize [1, 3]}} & \resultcell{28 {\footnotesize [27, 30]}} \\
 & Task 3 & \resultcell{0 {\footnotesize [0, 0]}} & \resultcell{\textcolor[HTML]{6495ED}{\textbf{20}} {\footnotesize [17, 23]}} & \resultcell{12 {\footnotesize [10, 13]}} & \resultcell{2 {\footnotesize [1, 3]}} & \resultcell{5 {\footnotesize [3, 7]}} & \resultcell{5 {\footnotesize [0, 10]}} & \resultcell{1 {\footnotesize [0, 2]}} & \resultcell{17 {\footnotesize [12, 22]}} \\
 & Task 4 & \resultcell{0 {\footnotesize [0, 0]}} & \resultcell{4 {\footnotesize [2, 7]}} & \resultcell{3 {\footnotesize [0, 8]}} & \resultcell{2 {\footnotesize [0, 5]}} & \resultcell{3 {\footnotesize [1, 6]}} & \resultcell{6 {\footnotesize [2, 9]}} & \resultcell{0 {\footnotesize [0, 0]}} & \resultcell{\textcolor[HTML]{6495ED}{\textbf{15}} {\footnotesize [12, 18]}} \\
 & Task 5 & \resultcell{0 {\footnotesize [0, 0]}} & \resultcell{\textcolor[HTML]{6495ED}{\textbf{6}} {\footnotesize [4, 7]}} & \resultcell{1 {\footnotesize [0, 2]}} & \resultcell{1 {\footnotesize [0, 2]}} & \resultcell{2 {\footnotesize [0, 5]}} & \resultcell{1 {\footnotesize [0, 2]}} & \resultcell{0 {\footnotesize [0, 0]}} & \resultcell{3 {\footnotesize [3, 3]}} \\
\cmidrule(lr){2-10}
 & Overall & \resultcell{6 {\footnotesize [6, 6]}} & \resultcell{\textcolor[HTML]{6495ED}{\textbf{24}} {\footnotesize [19, 27]}} & \resultcell{16 {\footnotesize [14, 18]}} & \resultcell{3 {\footnotesize [2, 4]}} & \resultcell{14 {\footnotesize [10, 18]}} & \resultcell{14 {\footnotesize [13, 16]}} & \resultcell{4 {\footnotesize [4, 4]}} & \resultcell{22 {\footnotesize [20, 23]}} \\
\midrule
\multirow{6}{*}{cube-quadruple-play-v0} & Task 1 & \resultcell{0 {\footnotesize [0, 0]}} & \resultcell{4 {\footnotesize [3, 6]}} & \resultcell{\textcolor[HTML]{6495ED}{\textbf{11}} {\footnotesize [5, 19]}} & \resultcell{0 {\footnotesize [0, 0]}} & \resultcell{2 {\footnotesize [0, 3]}} & \resultcell{2 {\footnotesize [0, 3]}} & \resultcell{0 {\footnotesize [0, 0]}} & \resultcell{9 {\footnotesize [8, 10]}} \\
 & Task 2 & \resultcell{0 {\footnotesize [0, 0]}} & \resultcell{4 {\footnotesize [3, 6]}} & \resultcell{4 {\footnotesize [2, 7]}} & \resultcell{0 {\footnotesize [0, 0]}} & \resultcell{1 {\footnotesize [0, 2]}} & \resultcell{1 {\footnotesize [0, 2]}} & \resultcell{0 {\footnotesize [0, 0]}} & \resultcell{\textcolor[HTML]{6495ED}{\textbf{10}} {\footnotesize [7, 13]}} \\
 & Task 3 & \resultcell{0 {\footnotesize [0, 0]}} & \resultcell{0 {\footnotesize [0, 0]}} & \resultcell{\textcolor[HTML]{6495ED}{\textbf{1}} {\footnotesize [0, 2]}} & \resultcell{0 {\footnotesize [0, 0]}} & \resultcell{0 {\footnotesize [0, 0]}} & \resultcell{0 {\footnotesize [0, 0]}} & \resultcell{0 {\footnotesize [0, 0]}} & \resultcell{\textcolor[HTML]{6495ED}{\textbf{1}} {\footnotesize [0, 2]}} \\
 & Task 4 & \resultcell{0 {\footnotesize [0, 0]}} & \resultcell{0 {\footnotesize [0, 0]}} & \resultcell{0 {\footnotesize [0, 0]}} & \resultcell{0 {\footnotesize [0, 0]}} & \resultcell{0 {\footnotesize [0, 0]}} & \resultcell{0 {\footnotesize [0, 0]}} & \resultcell{0 {\footnotesize [0, 0]}} & \resultcell{\textcolor[HTML]{6495ED}{\textbf{2}} {\footnotesize [0, 5]}} \\
 & Task 5 & \resultcell{0 {\footnotesize [0, 0]}} & \resultcell{0 {\footnotesize [0, 0]}} & \resultcell{0 {\footnotesize [0, 0]}} & \resultcell{0 {\footnotesize [0, 0]}} & \resultcell{0 {\footnotesize [0, 0]}} & \resultcell{0 {\footnotesize [0, 0]}} & \resultcell{0 {\footnotesize [0, 0]}} & \resultcell{\textcolor[HTML]{6495ED}{\textbf{1}} {\footnotesize [0, 2]}} \\
\cmidrule(lr){2-10}
 & Overall & \resultcell{0 {\footnotesize [0, 0]}} & \resultcell{2 {\footnotesize [1, 2]}} & \resultcell{3 {\footnotesize [1, 5]}} & \resultcell{0 {\footnotesize [0, 0]}} & \resultcell{0 {\footnotesize [0, 1]}} & \resultcell{0 {\footnotesize [0, 1]}} & \resultcell{0 {\footnotesize [0, 0]}} & \resultcell{\textcolor[HTML]{6495ED}{\textbf{4}} {\footnotesize [4, 5]}} \\
\midrule
\multirow{6}{*}{scene-play-v0} & Task 1 & \resultcell{86 {\footnotesize [78, 93]}} & \resultcell{\textcolor[HTML]{6495ED}{\textbf{100}} {\footnotesize [100, 100]}} & \resultcell{55 {\footnotesize [40, 65]}} & \resultcell{99 {\footnotesize [98, 100]}} & \resultcell{\textcolor[HTML]{6495ED}{\textbf{100}} {\footnotesize [100, 100]}} & \resultcell{\textcolor[HTML]{6495ED}{\textbf{100}} {\footnotesize [100, 100]}} & \resultcell{\textcolor[HTML]{6495ED}{\textbf{100}} {\footnotesize [100, 100]}} & \resultcell{\textcolor[HTML]{6495ED}{\textbf{100}} {\footnotesize [100, 100]}} \\
 & Task 2 & \resultcell{54 {\footnotesize [49, 58]}} & \resultcell{98 {\footnotesize [97, 99]}} & \resultcell{59 {\footnotesize [52, 71]}} & \resultcell{92 {\footnotesize [88, 98]}} & \resultcell{98 {\footnotesize [97, 100]}} & \resultcell{96 {\footnotesize [93, 98]}} & \resultcell{98 {\footnotesize [95, 100]}} & \resultcell{\textcolor[HTML]{6495ED}{\textbf{99}} {\footnotesize [98, 100]}} \\
 & Task 3 & \resultcell{64 {\footnotesize [57, 72]}} & \resultcell{97 {\footnotesize [94, 99]}} & \resultcell{22 {\footnotesize [20, 23]}} & \resultcell{80 {\footnotesize [67, 88]}} & \resultcell{98 {\footnotesize [97, 100]}} & \resultcell{\textcolor[HTML]{6495ED}{\textbf{100}} {\footnotesize [100, 100]}} & \resultcell{\textcolor[HTML]{6495ED}{\textbf{100}} {\footnotesize [100, 100]}} & \resultcell{82 {\footnotesize [78, 88]}} \\
 & Task 4 & \resultcell{33 {\footnotesize [23, 43]}} & \resultcell{65 {\footnotesize [60, 70]}} & \resultcell{47 {\footnotesize [38, 55]}} & \resultcell{76 {\footnotesize [68, 80]}} & \resultcell{85 {\footnotesize [75, 92]}} & \resultcell{84 {\footnotesize [82, 87]}} & \resultcell{\textcolor[HTML]{6495ED}{\textbf{86}} {\footnotesize [78, 93]}} & \resultcell{75 {\footnotesize [65, 82]}} \\
 & Task 5 & \resultcell{3 {\footnotesize [1, 6]}} & \resultcell{52 {\footnotesize [46, 56]}} & \resultcell{22 {\footnotesize [13, 30]}} & \resultcell{70 {\footnotesize [59, 78]}} & \resultcell{78 {\footnotesize [72, 85]}} & \resultcell{64 {\footnotesize [52, 77]}} & \resultcell{\textcolor[HTML]{6495ED}{\textbf{85}} {\footnotesize [81, 91]}} & \resultcell{69 {\footnotesize [60, 80]}} \\
\cmidrule(lr){2-10}
 & Overall & \resultcell{48 {\footnotesize [45, 51]}} & \resultcell{82 {\footnotesize [81, 83]}} & \resultcell{41 {\footnotesize [35, 46]}} & \resultcell{84 {\footnotesize [78, 87]}} & \resultcell{92 {\footnotesize [90, 93]}} & \resultcell{89 {\footnotesize [87, 91]}} & \resultcell{\textcolor[HTML]{6495ED}{\textbf{94}} {\footnotesize [91, 95]}} & \resultcell{85 {\footnotesize [82, 90]}} \\
\midrule
\multirow{6}{*}{puzzle-3x3-play-v0} & Task 1 & \resultcell{\textcolor[HTML]{6495ED}{\textbf{100}} {\footnotesize [100, 100]}} & \resultcell{94 {\footnotesize [82, 100]}} & \resultcell{63 {\footnotesize [57, 68]}} & \resultcell{94 {\footnotesize [90, 98]}} & \resultcell{\textcolor[HTML]{6495ED}{\textbf{100}} {\footnotesize [100, 100]}} & \resultcell{\textcolor[HTML]{6495ED}{\textbf{100}} {\footnotesize [100, 100]}} & \resultcell{99 {\footnotesize [98, 100]}} & \resultcell{\textcolor[HTML]{6495ED}{\textbf{100}} {\footnotesize [100, 100]}} \\
 & Task 2 & \resultcell{\textcolor[HTML]{6495ED}{\textbf{100}} {\footnotesize [100, 100]}} & \resultcell{91 {\footnotesize [75, 100]}} & \resultcell{58 {\footnotesize [42, 73]}} & \resultcell{72 {\footnotesize [59, 81]}} & \resultcell{74 {\footnotesize [55, 93]}} & \resultcell{85 {\footnotesize [79, 91]}} & \resultcell{78 {\footnotesize [70, 87]}} & \resultcell{81 {\footnotesize [73, 88]}} \\
 & Task 3 & \resultcell{\textcolor[HTML]{6495ED}{\textbf{100}} {\footnotesize [100, 100]}} & \resultcell{74 {\footnotesize [25, 100]}} & \resultcell{32 {\footnotesize [30, 35]}} & \resultcell{58 {\footnotesize [52, 62]}} & \resultcell{67 {\footnotesize [58, 76]}} & \resultcell{78 {\footnotesize [70, 87]}} & \resultcell{70 {\footnotesize [68, 72]}} & \resultcell{66 {\footnotesize [59, 73]}} \\
 & Task 4 & \resultcell{\textcolor[HTML]{6495ED}{\textbf{99}} {\footnotesize [98, 100]}} & \resultcell{72 {\footnotesize [24, 98]}} & \resultcell{36 {\footnotesize [32, 41]}} & \resultcell{59 {\footnotesize [57, 64]}} & \resultcell{69 {\footnotesize [58, 80]}} & \resultcell{76 {\footnotesize [67, 86]}} & \resultcell{52 {\footnotesize [40, 62]}} & \resultcell{48 {\footnotesize [38, 58]}} \\
 & Task 5 & \resultcell{\textcolor[HTML]{6495ED}{\textbf{100}} {\footnotesize [100, 100]}} & \resultcell{72 {\footnotesize [25, 100]}} & \resultcell{49 {\footnotesize [43, 55]}} & \resultcell{65 {\footnotesize [55, 75]}} & \resultcell{78 {\footnotesize [68, 90]}} & \resultcell{84 {\footnotesize [78, 90]}} & \resultcell{79 {\footnotesize [75, 84]}} & \resultcell{64 {\footnotesize [62, 67]}} \\
\cmidrule(lr){2-10}
 & Overall & \resultcell{\textcolor[HTML]{6495ED}{\textbf{100}} {\footnotesize [100, 100]}} & \resultcell{81 {\footnotesize [46, 99]}} & \resultcell{48 {\footnotesize [46, 49]}} & \resultcell{70 {\footnotesize [65, 74]}} & \resultcell{78 {\footnotesize [71, 87]}} & \resultcell{85 {\footnotesize [81, 90]}} & \resultcell{76 {\footnotesize [72, 80]}} & \resultcell{72 {\footnotesize [69, 75]}} \\
\midrule
\multirow{6}{*}{puzzle-4x4-play-v0} & Task 1 & \resultcell{85 {\footnotesize [77, 92]}} & \resultcell{94 {\footnotesize [90, 98]}} & \resultcell{0 {\footnotesize [0, 0]}} & \resultcell{70 {\footnotesize [63, 81]}} & \resultcell{85 {\footnotesize [82, 88]}} & \resultcell{76 {\footnotesize [68, 83]}} & \resultcell{81 {\footnotesize [77, 85]}} & \resultcell{\textcolor[HTML]{6495ED}{\textbf{96}} {\footnotesize [92, 100]}} \\
 & Task 2 & \resultcell{47 {\footnotesize [33, 60]}} & \resultcell{0 {\footnotesize [0, 0]}} & \resultcell{1 {\footnotesize [0, 2]}} & \resultcell{63 {\footnotesize [48, 78]}} & \resultcell{72 {\footnotesize [66, 78]}} & \resultcell{67 {\footnotesize [62, 70]}} & \resultcell{68 {\footnotesize [61, 75]}} & \resultcell{\textcolor[HTML]{6495ED}{\textbf{92}} {\footnotesize [80, 100]}} \\
 & Task 3 & \resultcell{78 {\footnotesize [64, 90]}} & \resultcell{32 {\footnotesize [20, 43]}} & \resultcell{0 {\footnotesize [0, 0]}} & \resultcell{65 {\footnotesize [56, 78]}} & \resultcell{74 {\footnotesize [63, 83]}} & \resultcell{78 {\footnotesize [75, 80]}} & \resultcell{60 {\footnotesize [51, 76]}} & \resultcell{\textcolor[HTML]{6495ED}{\textbf{92}} {\footnotesize [80, 100]}} \\
 & Task 4 & \resultcell{72 {\footnotesize [64, 78]}} & \resultcell{0 {\footnotesize [0, 0]}} & \resultcell{1 {\footnotesize [0, 2]}} & \resultcell{52 {\footnotesize [50, 53]}} & \resultcell{71 {\footnotesize [62, 80]}} & \resultcell{68 {\footnotesize [57, 78]}} & \resultcell{71 {\footnotesize [68, 75]}} & \resultcell{\textcolor[HTML]{6495ED}{\textbf{89}} {\footnotesize [75, 100]}} \\
 & Task 5 & \resultcell{72 {\footnotesize [56, 88]}} & \resultcell{0 {\footnotesize [0, 0]}} & \resultcell{0 {\footnotesize [0, 0]}} & \resultcell{72 {\footnotesize [67, 78]}} & \resultcell{68 {\footnotesize [53, 82]}} & \resultcell{73 {\footnotesize [68, 82]}} & \resultcell{67 {\footnotesize [62, 73]}} & \resultcell{\textcolor[HTML]{6495ED}{\textbf{92}} {\footnotesize [78, 100]}} \\
\cmidrule(lr){2-10}
 & Overall & \resultcell{71 {\footnotesize [61, 78]}} & \resultcell{25 {\footnotesize [23, 27]}} & \resultcell{0 {\footnotesize [0, 1]}} & \resultcell{64 {\footnotesize [61, 68]}} & \resultcell{74 {\footnotesize [67, 79]}} & \resultcell{72 {\footnotesize [69, 75]}} & \resultcell{69 {\footnotesize [65, 75]}} & \resultcell{\textcolor[HTML]{6495ED}{\textbf{92}} {\footnotesize [81, 100]}} \\
\midrule
\multirow{6}{*}{puzzle-4x5-play-v0} & Task 1 & \resultcell{68 {\footnotesize [53, 76]}} & \resultcell{\textcolor[HTML]{6495ED}{\textbf{81}} {\footnotesize [63, 97]}} & \resultcell{18 {\footnotesize [13, 23]}} & \resultcell{43 {\footnotesize [39, 47]}} & \resultcell{46 {\footnotesize [38, 55]}} & \resultcell{45 {\footnotesize [40, 52]}} & \resultcell{46 {\footnotesize [36, 60]}} & \resultcell{36 {\footnotesize [29, 42]}} \\
 & Task 2 & \resultcell{0 {\footnotesize [0, 0]}} & \resultcell{0 {\footnotesize [0, 0]}} & \resultcell{0 {\footnotesize [0, 0]}} & \resultcell{\textcolor[HTML]{6495ED}{\textbf{36}} {\footnotesize [27, 46]}} & \resultcell{17 {\footnotesize [10, 23]}} & \resultcell{27 {\footnotesize [18, 35]}} & \resultcell{19 {\footnotesize [7, 32]}} & \resultcell{20 {\footnotesize [15, 25]}} \\
 & Task 3 & \resultcell{0 {\footnotesize [0, 0]}} & \resultcell{0 {\footnotesize [0, 0]}} & \resultcell{0 {\footnotesize [0, 0]}} & \resultcell{7 {\footnotesize [2, 12]}} & \resultcell{20 {\footnotesize [12, 27]}} & \resultcell{\textcolor[HTML]{6495ED}{\textbf{21}} {\footnotesize [18, 23]}} & \resultcell{6 {\footnotesize [3, 8]}} & \resultcell{3 {\footnotesize [0, 10]}} \\
 & Task 4 & \resultcell{0 {\footnotesize [0, 0]}} & \resultcell{0 {\footnotesize [0, 0]}} & \resultcell{1 {\footnotesize [0, 2]}} & \resultcell{\textcolor[HTML]{6495ED}{\textbf{30}} {\footnotesize [23, 37]}} & \resultcell{1 {\footnotesize [0, 2]}} & \resultcell{0 {\footnotesize [0, 0]}} & \resultcell{1 {\footnotesize [0, 2]}} & \resultcell{17 {\footnotesize [5, 28]}} \\
 & Task 5 & \resultcell{0 {\footnotesize [0, 0]}} & \resultcell{0 {\footnotesize [0, 0]}} & \resultcell{0 {\footnotesize [0, 0]}} & \resultcell{\textcolor[HTML]{6495ED}{\textbf{2}} {\footnotesize [0, 3]}} & \resultcell{1 {\footnotesize [0, 2]}} & \resultcell{0 {\footnotesize [0, 0]}} & \resultcell{0 {\footnotesize [0, 0]}} & \resultcell{1 {\footnotesize [0, 2]}} \\
\cmidrule(lr){2-10}
 & Overall & \resultcell{14 {\footnotesize [11, 15]}} & \resultcell{16 {\footnotesize [13, 19]}} & \resultcell{4 {\footnotesize [3, 5]}} & \resultcell{\textcolor[HTML]{6495ED}{\textbf{24}} {\footnotesize [22, 25]}} & \resultcell{17 {\footnotesize [15, 20]}} & \resultcell{18 {\footnotesize [18, 19]}} & \resultcell{14 {\footnotesize [11, 18]}} & \resultcell{15 {\footnotesize [11, 20]}} \\
\midrule
\multirow{6}{*}{puzzle-4x6-play-v0} & Task 1 & \resultcell{32 {\footnotesize [20, 45]}} & \resultcell{\textcolor[HTML]{6495ED}{\textbf{80}} {\footnotesize [68, 92]}} & \resultcell{20 {\footnotesize [18, 22]}} & \resultcell{39 {\footnotesize [33, 43]}} & \resultcell{39 {\footnotesize [30, 49]}} & \resultcell{33 {\footnotesize [27, 38]}} & \resultcell{38 {\footnotesize [33, 43]}} & \resultcell{45 {\footnotesize [41, 51]}} \\
 & Task 2 & \resultcell{10 {\footnotesize [3, 17]}} & \resultcell{0 {\footnotesize [0, 0]}} & \resultcell{5 {\footnotesize [2, 8]}} & \resultcell{34 {\footnotesize [31, 38]}} & \resultcell{21 {\footnotesize [15, 28]}} & \resultcell{30 {\footnotesize [27, 34]}} & \resultcell{14 {\footnotesize [5, 24]}} & \resultcell{\textcolor[HTML]{6495ED}{\textbf{37}} {\footnotesize [28, 50]}} \\
 & Task 3 & \resultcell{0 {\footnotesize [0, 0]}} & \resultcell{0 {\footnotesize [0, 0]}} & \resultcell{0 {\footnotesize [0, 0]}} & \resultcell{8 {\footnotesize [5, 10]}} & \resultcell{7 {\footnotesize [3, 11]}} & \resultcell{\textcolor[HTML]{6495ED}{\textbf{14}} {\footnotesize [7, 22]}} & \resultcell{4 {\footnotesize [2, 7]}} & \resultcell{8 {\footnotesize [3, 13]}} \\
 & Task 4 & \resultcell{0 {\footnotesize [0, 0]}} & \resultcell{0 {\footnotesize [0, 0]}} & \resultcell{0 {\footnotesize [0, 0]}} & \resultcell{\textcolor[HTML]{6495ED}{\textbf{2}} {\footnotesize [0, 5]}} & \resultcell{1 {\footnotesize [0, 2]}} & \resultcell{1 {\footnotesize [0, 2]}} & \resultcell{0 {\footnotesize [0, 0]}} & \resultcell{1 {\footnotesize [0, 2]}} \\
 & Task 5 & \resultcell{0 {\footnotesize [0, 0]}} & \resultcell{0 {\footnotesize [0, 0]}} & \resultcell{0 {\footnotesize [0, 0]}} & \resultcell{0 {\footnotesize [0, 0]}} & \resultcell{0 {\footnotesize [0, 0]}} & \resultcell{0 {\footnotesize [0, 0]}} & \resultcell{0 {\footnotesize [0, 0]}} & \resultcell{0 {\footnotesize [0, 0]}} \\
\cmidrule(lr){2-10}
 & Overall & \resultcell{8 {\footnotesize [5, 10]}} & \resultcell{16 {\footnotesize [14, 18]}} & \resultcell{5 {\footnotesize [4, 6]}} & \resultcell{17 {\footnotesize [15, 18]}} & \resultcell{14 {\footnotesize [10, 16]}} & \resultcell{16 {\footnotesize [14, 17]}} & \resultcell{11 {\footnotesize [8, 15]}} & \resultcell{\textcolor[HTML]{6495ED}{\textbf{18}} {\footnotesize [15, 22]}} \\
\midrule
\multicolumn{2}{l}{Average over 55 Tasks} & \resultcell{29 {\footnotesize [27, 30]}} & \resultcell{32 {\footnotesize [29, 34]}} & \cellcolor[HTML]{E6E6E6}\resultcell{20 {\footnotesize [20, 20]}} & \cellcolor[HTML]{E6E6E6}\resultcell{45 {\footnotesize [44, 46]}} & \cellcolor[HTML]{E6E6E6}\resultcell{46 {\footnotesize [44, 49]}} & \cellcolor[HTML]{E6E6E6}\resultcell{49 {\footnotesize [48, 49]}} & \cellcolor[HTML]{E6E6E6}\resultcell{47 {\footnotesize [46, 48]}} & \cellcolor[HTML]{DCE6FC}\resultcell{\textcolor[HTML]{6495ED}{\textbf{54}} {\footnotesize [51, 56]}} \\
\bottomrule
\end{tabular}}
\caption{\textbf{Full results on selected OGBench manipulation environments using default datasets.} Success rate (\%) on each evaluation task after 1M offline training steps. Mean and 95\% bootstrap confidence interval over 4 seeds. Last row averages each seed over all 55 tasks, locomotion and manipulation (Table~\ref{tab: results_appendix_locomotion}).}
\label{tab: results_appendix_manipulation}
\end{table}

\newpage
\subsection{Analysis and Ablations}

\begin{figure}[H]
    \centering
    \includegraphics[width=\linewidth]{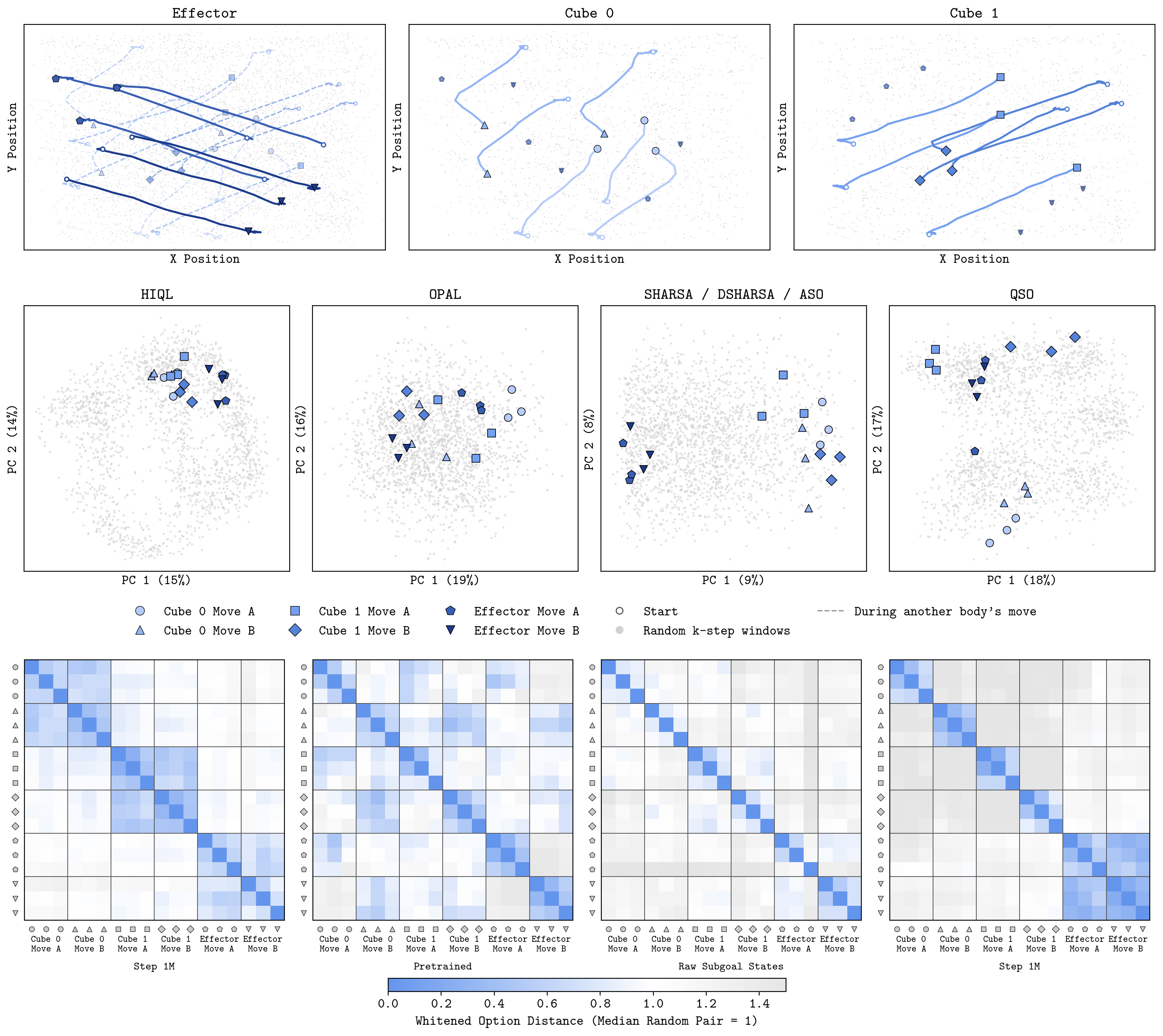}
    \caption{\textbf{PCA} (\textbf{top}) of the action abstraction for cube-double fitted on random segments, showing different groups of segments with semantically similar behaviour (randomly selected from the validation dataset).  \textbf{L2 distances} (\textbf{bottom}) in units of median distance between two random segments. QSO is the only algorithm that consistently groups functionally equivalent behaviours (\textbf{Desideratum 3}) while separating behaviour groups (\textbf{Desideratum 2}).}
    \label{fig: cube_double_option_embeddings}
\end{figure}

\begin{figure}[H]
    \centering
    \includegraphics[width=\linewidth]{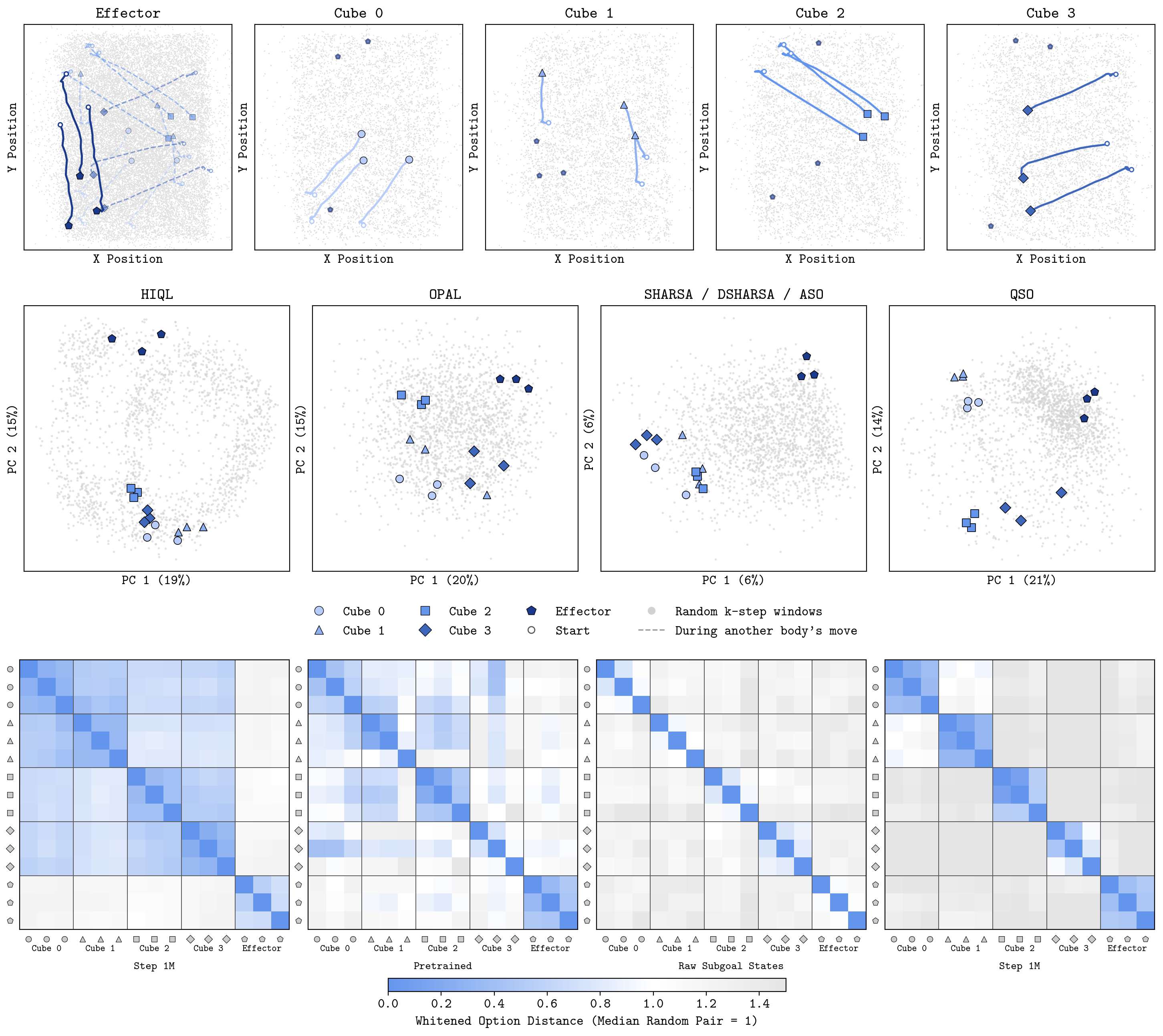}
    \caption{\textbf{PCA} (\textbf{top}) of the action abstraction for cube-quadruple fitted on random segments, showing different groups of segments with semantically similar behaviour (randomly selected from the validation dataset).  \textbf{L2 distances} (\textbf{bottom}) in units of median distance between two random segments. QSO is the only algorithm that consistently groups functionally equivalent behaviours (\textbf{Desideratum 3}) while separating behaviour groups (\textbf{Desideratum 2}).}
    \label{fig: cube_quadruple_option_embeddings}
\end{figure}

\begin{table}[H]
\centering
\providecommand{\resultcell}[1]{\rule{0pt}{3.2ex}#1\rule[-1.4ex]{0pt}{0pt}}
\renewcommand{\arraystretch}{1.0}
\setlength{\tabcolsep}{4pt}
\resizebox{1.0\textwidth}{!}{\ttfamily
\begin{tabular}{lll>{\columncolor[HTML]{F2F2F2}}c>{\columncolor[HTML]{F2F2F2}}c>{\columncolor[HTML]{F2F2F2}}cccc>{\columncolor[HTML]{F2F2F2}}c>{\columncolor[HTML]{F2F2F2}}c>{\columncolor[HTML]{F2F2F2}}cccc}
\toprule
 & \multicolumn{2}{c}{Reward Labelling} & \multicolumn{3}{c}{High-level Policy} & \multicolumn{3}{c}{High-level Value} & \multicolumn{3}{c}{Option Encoder} & \multicolumn{3}{c}{Low-level Policy} \\
\cmidrule(lr){2-3} \cmidrule(lr){4-6} \cmidrule(lr){7-9} \cmidrule(lr){10-12} \cmidrule(lr){13-15}
Agent & High & Low & Fast & Slow & \shortstack{Interaction/\\Noise} & Fast & Slow & \shortstack{Interaction} & Fast & Slow & \shortstack{Interaction} & Fast & Slow & \shortstack{Interaction/\\Noise} \\
\midrule
\multicolumn{15}{l}{\rule{0pt}{2.6ex}\textit{antmaze-giant-stitch-v0}} \\
\multirow{4}{*}{\textcolor[HTML]{6495ED}{QSO}} & E & S & \resultcell{35 {\footnotesize [31, 39]}} & \resultcell{14 {\footnotesize [13, 16]}} & \resultcell{50 {\footnotesize [47, 54]}} & \resultcell{6 {\footnotesize [4, 7]}} & \resultcell{88 {\footnotesize [86, 90]}} & \resultcell{7 {\footnotesize [6, 7]}} & \resultcell{45 {\footnotesize [42, 47]}} & \resultcell{45 {\footnotesize [42, 47]}} & \resultcell{11 {\footnotesize [10, 11]}} & \resultcell{59 {\footnotesize [57, 60]}} & \resultcell{4 {\footnotesize [4, 4]}} & \resultcell{37 {\footnotesize [36, 38]}} \\
 & E & E & \resultcell{28 {\footnotesize [26, 30]}} & \resultcell{18 {\footnotesize [17, 19]}} & \resultcell{54 {\footnotesize [52, 55]}} & \resultcell{6 {\footnotesize [4, 7]}} & \resultcell{88 {\footnotesize [86, 90]}} & \resultcell{6 {\footnotesize [6, 7]}} & \resultcell{23 {\footnotesize [21, 25]}} & \resultcell{72 {\footnotesize [70, 74]}} & \resultcell{5 {\footnotesize [5, 6]}} & \resultcell{55 {\footnotesize [54, 57]}} & \resultcell{5 {\footnotesize [4, 5]}} & \resultcell{40 {\footnotesize [39, 41]}} \\
 & S & S & \resultcell{34 {\footnotesize [30, 38]}} & \resultcell{14 {\footnotesize [13, 15]}} & \resultcell{52 {\footnotesize [49, 56]}} & \resultcell{8 {\footnotesize [6, 10]}} & \resultcell{84 {\footnotesize [81, 87]}} & \resultcell{8 {\footnotesize [7, 9]}} & \resultcell{46 {\footnotesize [43, 48]}} & \resultcell{45 {\footnotesize [43, 48]}} & \resultcell{9 {\footnotesize [9, 9]}} & \resultcell{54 {\footnotesize [53, 56]}} & \resultcell{5 {\footnotesize [4, 5]}} & \resultcell{41 {\footnotesize [40, 42]}} \\
 & S & E & \resultcell{29 {\footnotesize [26, 32]}} & \resultcell{19 {\footnotesize [18, 20]}} & \resultcell{52 {\footnotesize [51, 54]}} & \resultcell{8 {\footnotesize [6, 11]}} & \resultcell{84 {\footnotesize [81, 87]}} & \resultcell{8 {\footnotesize [7, 8]}} & \resultcell{28 {\footnotesize [25, 30]}} & \resultcell{68 {\footnotesize [65, 70]}} & \resultcell{5 {\footnotesize [5, 5]}} & \resultcell{56 {\footnotesize [54, 57]}} & \resultcell{4 {\footnotesize [4, 5]}} & \resultcell{40 {\footnotesize [39, 41]}} \\
\cmidrule(lr){1-15}
\multirow{4}{*}{ASO} & E & S & \resultcell{21 {\footnotesize [18, 24]}} & \resultcell{13 {\footnotesize [12, 14]}} & \resultcell{67 {\footnotesize [64, 69]}} & \resultcell{5 {\footnotesize [4, 6]}} & \resultcell{88 {\footnotesize [86, 90]}} & \resultcell{7 {\footnotesize [7, 8]}} & \resultcell{--} & \resultcell{--} & \resultcell{--} & \resultcell{53 {\footnotesize [51, 54]}} & \resultcell{5 {\footnotesize [5, 5]}} & \resultcell{42 {\footnotesize [41, 44]}} \\
 & E & E & \resultcell{21 {\footnotesize [18, 24]}} & \resultcell{13 {\footnotesize [12, 14]}} & \resultcell{67 {\footnotesize [64, 69]}} & \resultcell{5 {\footnotesize [4, 6]}} & \resultcell{88 {\footnotesize [86, 90]}} & \resultcell{7 {\footnotesize [7, 8]}} & \resultcell{--} & \resultcell{--} & \resultcell{--} & \resultcell{53 {\footnotesize [51, 54]}} & \resultcell{5 {\footnotesize [5, 5]}} & \resultcell{42 {\footnotesize [41, 44]}} \\
 & S & S & \resultcell{19 {\footnotesize [16, 22]}} & \resultcell{13 {\footnotesize [12, 14]}} & \resultcell{68 {\footnotesize [66, 70]}} & \resultcell{8 {\footnotesize [6, 10]}} & \resultcell{83 {\footnotesize [80, 85]}} & \resultcell{9 {\footnotesize [8, 10]}} & \resultcell{--} & \resultcell{--} & \resultcell{--} & \resultcell{54 {\footnotesize [53, 55]}} & \resultcell{5 {\footnotesize [4, 5]}} & \resultcell{41 {\footnotesize [40, 43]}} \\
 & S & E & \resultcell{19 {\footnotesize [16, 22]}} & \resultcell{13 {\footnotesize [12, 14]}} & \resultcell{68 {\footnotesize [66, 70]}} & \resultcell{8 {\footnotesize [6, 10]}} & \resultcell{83 {\footnotesize [80, 85]}} & \resultcell{9 {\footnotesize [8, 10]}} & \resultcell{--} & \resultcell{--} & \resultcell{--} & \resultcell{54 {\footnotesize [53, 55]}} & \resultcell{5 {\footnotesize [4, 5]}} & \resultcell{41 {\footnotesize [40, 43]}} \\
\midrule
\multicolumn{15}{l}{\rule{0pt}{2.6ex}\textit{cube-double-play-v0}} \\
\multirow{4}{*}{\textcolor[HTML]{6495ED}{QSO}} & E & S & \resultcell{73 {\footnotesize [71, 74]}} & \resultcell{5 {\footnotesize [5, 5]}} & \resultcell{22 {\footnotesize [21, 24]}} & \resultcell{3 {\footnotesize [2, 4]}} & \resultcell{74 {\footnotesize [70, 77]}} & \resultcell{23 {\footnotesize [20, 26]}} & \resultcell{79 {\footnotesize [77, 80]}} & \resultcell{15 {\footnotesize [14, 16]}} & \resultcell{6 {\footnotesize [6, 7]}} & \resultcell{77 {\footnotesize [75, 78]}} & \resultcell{4 {\footnotesize [4, 5]}} & \resultcell{19 {\footnotesize [18, 20]}} \\
 & E & E & \resultcell{25 {\footnotesize [23, 27]}} & \resultcell{23 {\footnotesize [21, 24]}} & \resultcell{52 {\footnotesize [51, 53]}} & \resultcell{3 {\footnotesize [3, 4]}} & \resultcell{75 {\footnotesize [72, 78]}} & \resultcell{21 {\footnotesize [19, 24]}} & \resultcell{8 {\footnotesize [8, 9]}} & \resultcell{84 {\footnotesize [83, 85]}} & \resultcell{8 {\footnotesize [7, 8]}} & \resultcell{71 {\footnotesize [69, 72]}} & \resultcell{8 {\footnotesize [7, 8]}} & \resultcell{22 {\footnotesize [21, 23]}} \\
 & S & S & \resultcell{71 {\footnotesize [70, 72]}} & \resultcell{7 {\footnotesize [6, 7]}} & \resultcell{22 {\footnotesize [21, 23]}} & \resultcell{4 {\footnotesize [3, 5]}} & \resultcell{70 {\footnotesize [67, 72]}} & \resultcell{26 {\footnotesize [24, 28]}} & \resultcell{76 {\footnotesize [75, 78]}} & \resultcell{15 {\footnotesize [14, 16]}} & \resultcell{9 {\footnotesize [8, 9]}} & \resultcell{81 {\footnotesize [80, 83]}} & \resultcell{5 {\footnotesize [4, 5]}} & \resultcell{14 {\footnotesize [13, 15]}} \\
 & S & E & \resultcell{18 {\footnotesize [17, 20]}} & \resultcell{39 {\footnotesize [37, 41]}} & \resultcell{42 {\footnotesize [41, 44]}} & \resultcell{4 {\footnotesize [3, 5]}} & \resultcell{73 {\footnotesize [71, 76]}} & \resultcell{23 {\footnotesize [21, 25]}} & \resultcell{8 {\footnotesize [7, 9]}} & \resultcell{83 {\footnotesize [82, 84]}} & \resultcell{9 {\footnotesize [9, 9]}} & \resultcell{75 {\footnotesize [74, 77]}} & \resultcell{5 {\footnotesize [4, 5]}} & \resultcell{20 {\footnotesize [19, 21]}} \\
\cmidrule(lr){1-15}
\multirow{4}{*}{ASO} & E & S & \resultcell{26 {\footnotesize [25, 27]}} & \resultcell{52 {\footnotesize [51, 53]}} & \resultcell{22 {\footnotesize [21, 23]}} & \resultcell{3 {\footnotesize [3, 4]}} & \resultcell{75 {\footnotesize [71, 78]}} & \resultcell{22 {\footnotesize [19, 25]}} & \resultcell{--} & \resultcell{--} & \resultcell{--} & \resultcell{88 {\footnotesize [87, 89]}} & \resultcell{2 {\footnotesize [2, 2]}} & \resultcell{10 {\footnotesize [9, 11]}} \\
 & E & E & \resultcell{26 {\footnotesize [25, 26]}} & \resultcell{52 {\footnotesize [51, 53]}} & \resultcell{23 {\footnotesize [22, 24]}} & \resultcell{3 {\footnotesize [3, 4]}} & \resultcell{74 {\footnotesize [70, 78]}} & \resultcell{22 {\footnotesize [19, 26]}} & \resultcell{--} & \resultcell{--} & \resultcell{--} & \resultcell{88 {\footnotesize [87, 89]}} & \resultcell{2 {\footnotesize [2, 2]}} & \resultcell{10 {\footnotesize [9, 11]}} \\
 & S & S & \resultcell{26 {\footnotesize [25, 27]}} & \resultcell{52 {\footnotesize [51, 53]}} & \resultcell{22 {\footnotesize [21, 23]}} & \resultcell{4 {\footnotesize [3, 5]}} & \resultcell{66 {\footnotesize [63, 69]}} & \resultcell{30 {\footnotesize [28, 32]}} & \resultcell{--} & \resultcell{--} & \resultcell{--} & \resultcell{88 {\footnotesize [87, 89]}} & \resultcell{3 {\footnotesize [2, 3]}} & \resultcell{9 {\footnotesize [8, 10]}} \\
 & S & E & \resultcell{26 {\footnotesize [25, 27]}} & \resultcell{52 {\footnotesize [51, 53]}} & \resultcell{22 {\footnotesize [21, 23]}} & \resultcell{4 {\footnotesize [3, 5]}} & \resultcell{66 {\footnotesize [63, 69]}} & \resultcell{30 {\footnotesize [28, 32]}} & \resultcell{--} & \resultcell{--} & \resultcell{--} & \resultcell{88 {\footnotesize [87, 89]}} & \resultcell{3 {\footnotesize [2, 3]}} & \resultcell{9 {\footnotesize [8, 10]}} \\
\bottomrule
\end{tabular}}
\caption{\textbf{QSO and ASO high and low-level decision process state abstractions.} Percentage share of the variance of each output explained by slow-moving dimensions (state and goal XY coordinates for antmaze-giant; state and goal cube positions and orientations for cube-double) and by the fast-moving dimensions (respectively state and goal XY height, orientation, joints and velocities; and state and goal effector, gripper, joints and velocities). The reward labelling columns indicate whether the high and low-level decision processes are trained using environment (E) or strict (S) reward labelling. Inputs are built on grids that combine the slow-moving dimensions of 16 random validation state-goal (or state-target) pairs with the fast-moving dimensions of 16 others, each under 8 flow-noise samples. For QSO, the state-target pairs are encoded before passing them to the low-level policy; the option encoder columns give the variance of these encoded options (ASO has no encoder). For ASO's high-level policy and both ASO's and QSO's low-level policies, each dimension of the output is divided by the larger of its standard deviations over the dataset and over the sampled absolute subgoal options or actions. The variance is attributed exactly using two-way ANOVA, so that the attributions sum to 100. Mean and 95\% bootstrap confidence interval (2000 resamples of the 64 grids) for one seed after 1M training steps.}
\label{tab: state_abstraction}
\end{table}

\begin{figure}[H]
    \centering
    \includegraphics[width=0.6\linewidth]{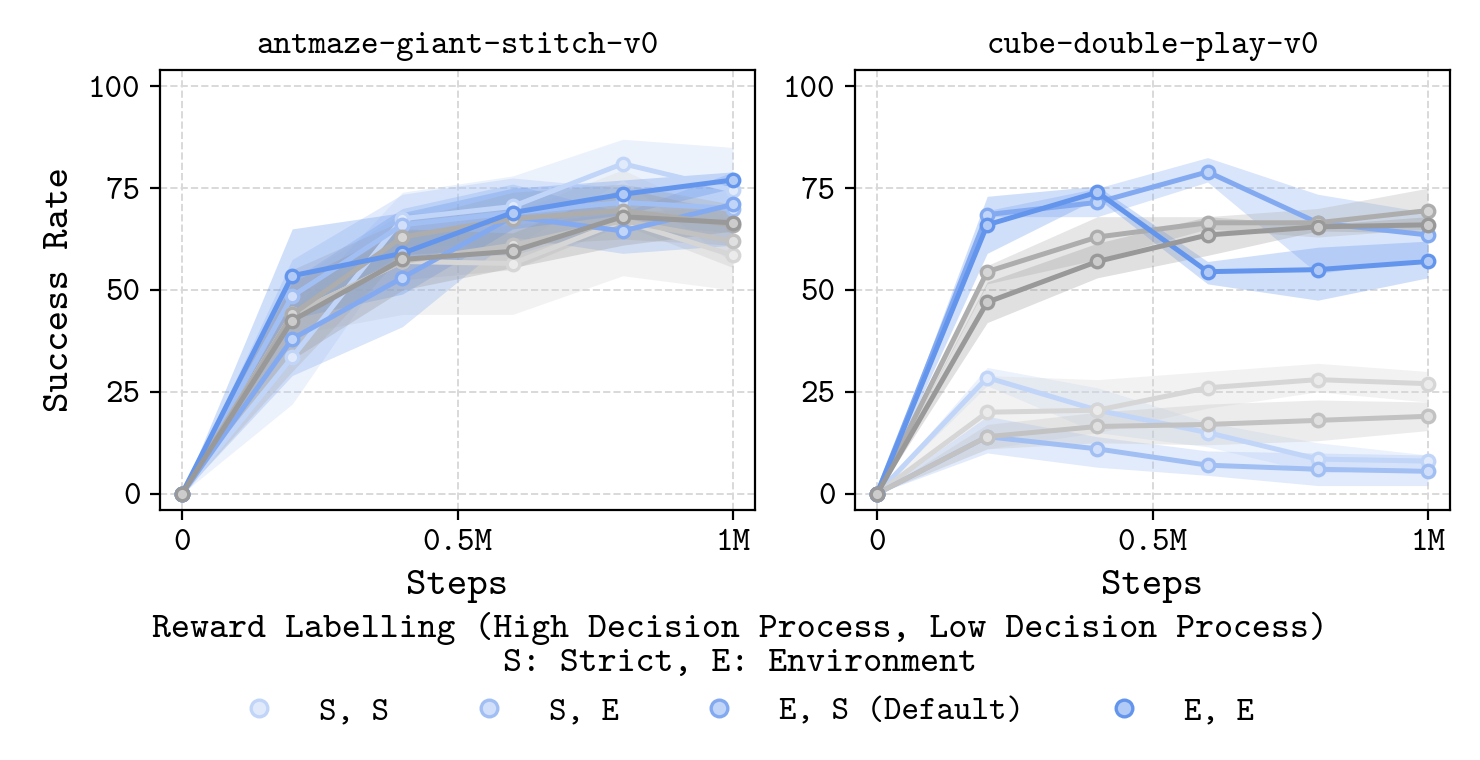}
    \caption{Strict versus environment dataset \textbf{reward labelling ablation} for QSO (\textbf{blue}) and ASO (\textbf{grey}).}
    \label{fig: ablation_reward_label}
\end{figure}

\paragraph{Further discussion on the coherence of state abstractions.} The high-level policy remains dependent on fast-moving dimensions (with 35\% of the output variance being explained by fast-moving dimensions in antmaze-giant and 73\% in cube-double). We attribute this to our method of Q-shaping: because the option is shaped to retain information about the required immediate action, if the target's fast-moving dimensions form part of the representation that the option encodes, the high-level policy inherits this dependence. The following ablation confirms this hypothesis. We ablate using the environment success criterion (which only focusses on slow-moving dimensions) rather than the stricter success criterion to also label the reward in the dataset for the low-level decision process. Doing so makes the option encoder and hence also high-level policy substantially less dependent on fast-moving dimensions: percentage shares for the option encoder drop from 45\% to 23\% and from 79\% to 8\% respectively, while for the high-level policy, they drop from 35\% to 28\% and 73\% to 25\%. Because we see no loss in downstream performance (Figure \ref{fig: ablation_reward_label}) and although we use the stricter success criterion for labelling reward in the low-level decision process across all experiments, we hypothesise that the environment success criterion might be preferable at both levels. 

\begin{figure}[H]
    \centering
    \includegraphics[width=0.6\linewidth]{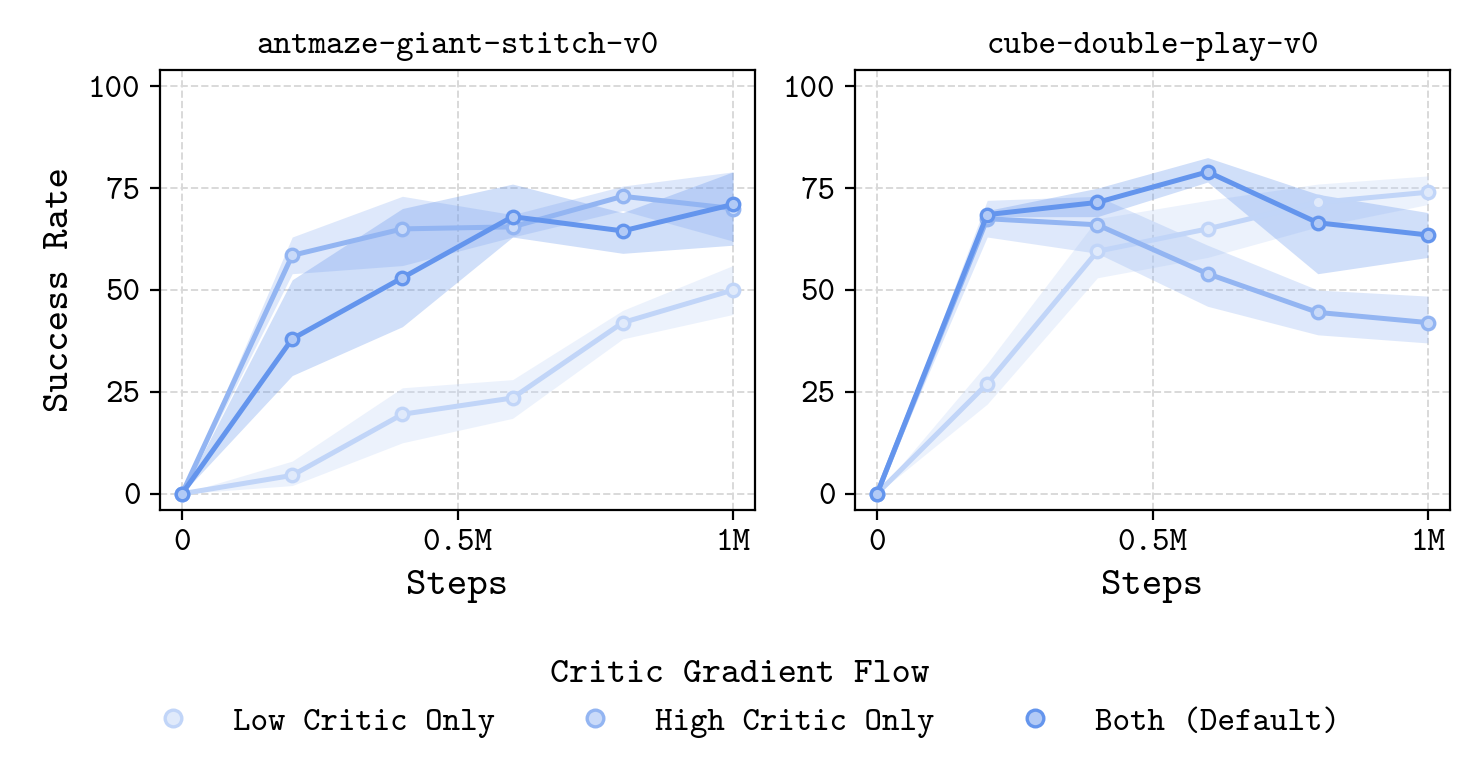}
    \caption{High and low-level critic \textbf{Q-function shaping ablation}.}
    \label{fig: ablation_gradient_flow}
\end{figure}

\begin{figure}[H]
    \centering
    \includegraphics[width=\linewidth]{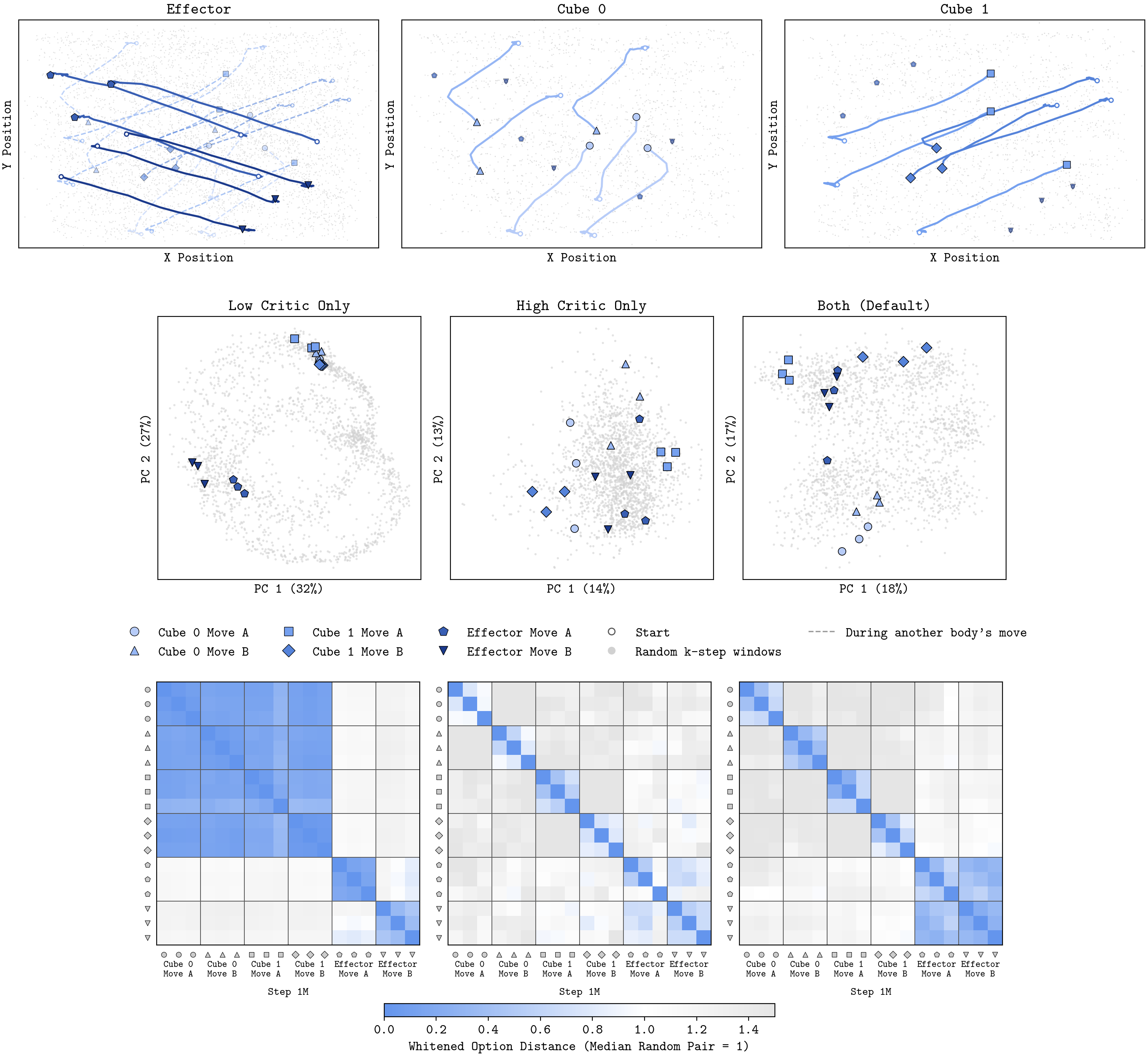}
    \caption{\textbf{PCA} (\textbf{top}) of QSO's action abstractions for gradient flow ablations for cube-double fitted on random trajectory sequences, showing different groups of three k-step trajectory sequences with semantically similar behaviour (randomly selected from the validation dataset).  \textbf{L2 distances} (\textbf{bottom}) for those groups in units of median distance between two random trajectory sequences. Both high and low-level gradient flows are required to group functionally equivalent behaviours while separating the behaviour groups.}
    \label{fig: cube_critic_option_embeddings}
\end{figure}

\begin{figure}[H]
    \centering
    \includegraphics[width=0.6\linewidth]{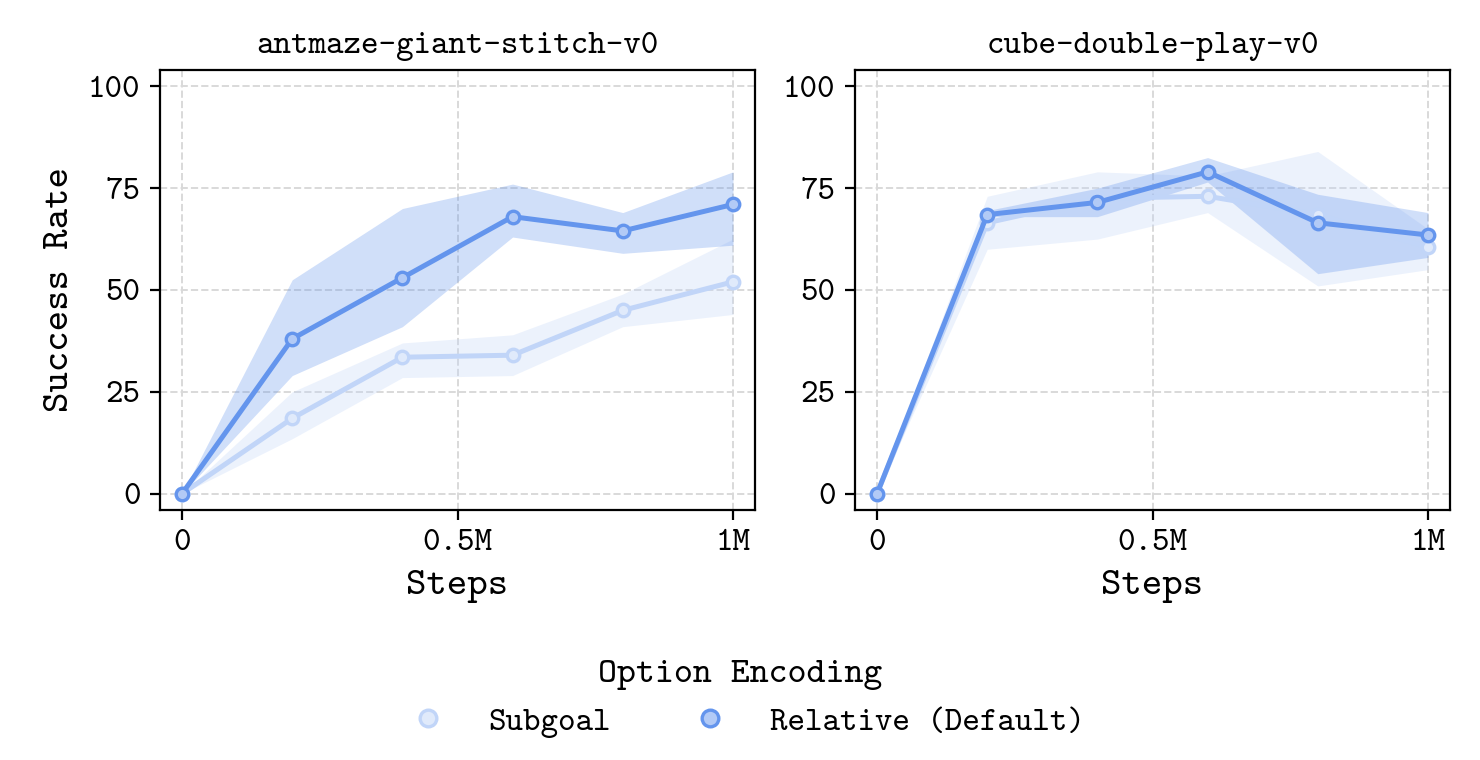}
    \caption{$\phi_\omega(s_{t+k})$ rather than $\phi_\omega(s_t,s_{t+k})$ \textbf{option encoding ablation}.}
    \label{fig: ablation_rep_encoding}
\end{figure}

\newpage
\textbf{How sensitive is QSO to option representation input?} Although our reasoning in Section \ref{sec: method} motivates representing options as relative achievements, we test whether this design choice improves downstream performance. Figure \ref{fig: ablation_rep_encoding} compares performance with absolute achievement $\phi_\omega(s_{t+k})$ with relative achievement encodings $\phi_\omega(s_t,s_{t+k})$. Relative encoding substantially improves performance in antmaze-giant, but has little effect in cube-double. We hypothesise this to be because relative encoding provides the largest benefit when many absolute state-goal pairs $(s, g_s)$ can be collapsed into the same context-dependent achievement, which is the case for the locomotion task, but not necessarily for the manipulation one, where the above state-abstraction analysis suggests that the option retains much more fine-grained information for immediate control.

\begin{figure}[H]
    \centering
    \includegraphics[width=0.6\linewidth]{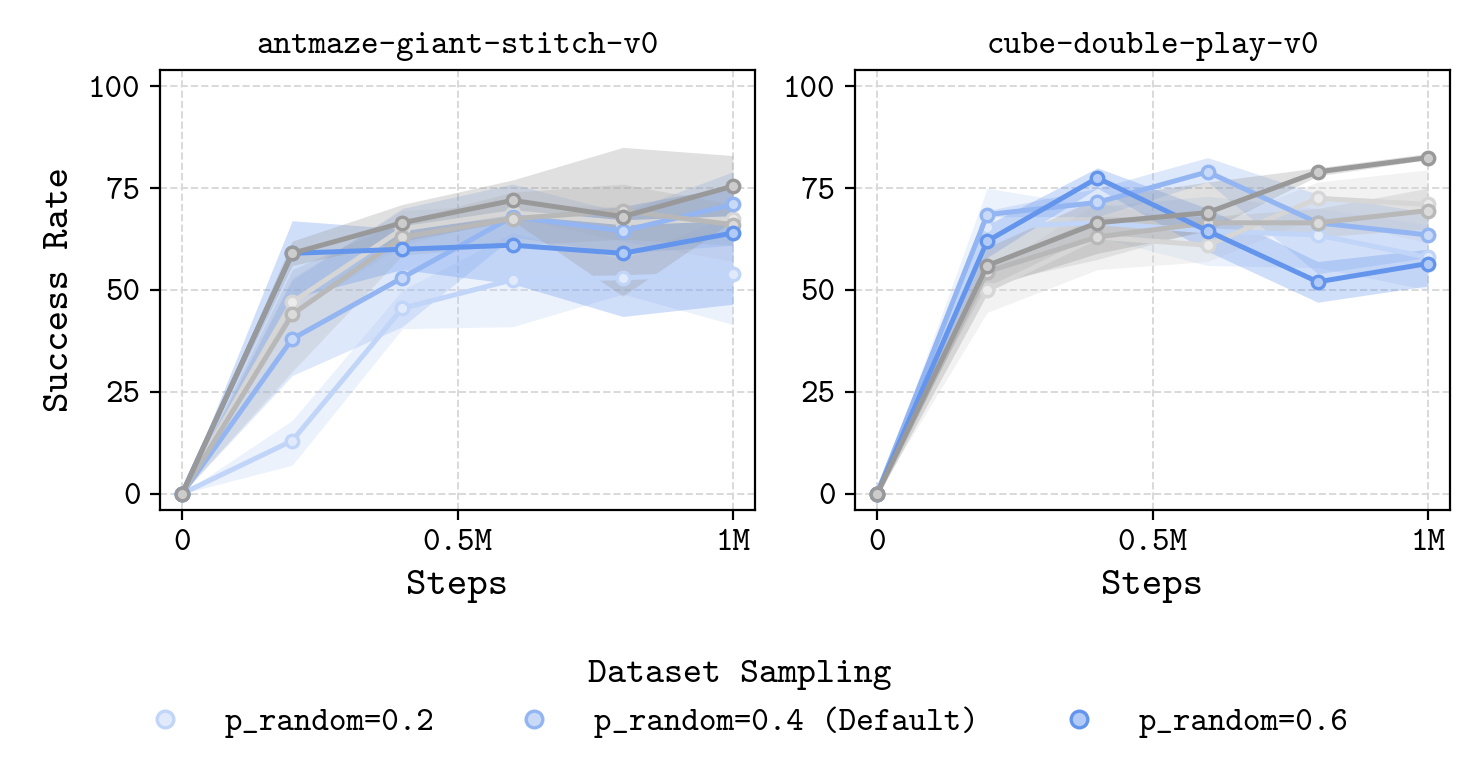}
    \caption{Random \textbf{goal sampling ablation} for the high-level decision process for QSO (\textbf{blue}) and ASO (\textbf{grey}).}
    \label{fig: ablation_dataset_sampling}
\end{figure}

\textbf{How sensitive is QSO to goal sampling?} Both QSO and ASO are relatively robust to the proportion of random high-level goals (Figure \ref{fig: ablation_dataset_sampling}). In both antmaze-giant and cube-double, all three sampling ratios converge to similar performance in both algorithms. 

\newpage
\paragraph{How sensitive is QSO to reward labelling?} Environment reward labelling for the high-level decision process substantially improves performance for both QSO and ASO in cube-double, while on antmaze-giant the effect is small (Figure \ref{fig: ablation_reward_label}). For ASO, this gain comes from the high-level critic: ranking samples from the environment-labelled high-level policy using the strict-labelled high-level critic recovers the full performance (of ~70\% rather than ~20\%), while the reverse swap results in a corresponding drop in performance. We hypothesise that the environment reward labelling enables the high-level decision process to form better state abstractions: consistently, the percentage share of the output variance for slow-moving dimensions in the high-level state-value function increases by almost 10\% (66\% to 75\%) for cube-double (Table \ref{tab: state_abstraction}). This is not the case for antmaze-giant, where even strict labelling yields a value dominated by the slow-moving dimensions. Note that we analyse percentage shares of output variance for the state-value function rather than the critic, which would add a confounding factor of how to sample the subgoal/option for the crossed state-goal pairs. 

\begin{figure}[H]
    \centering
    \includegraphics[width=\linewidth]{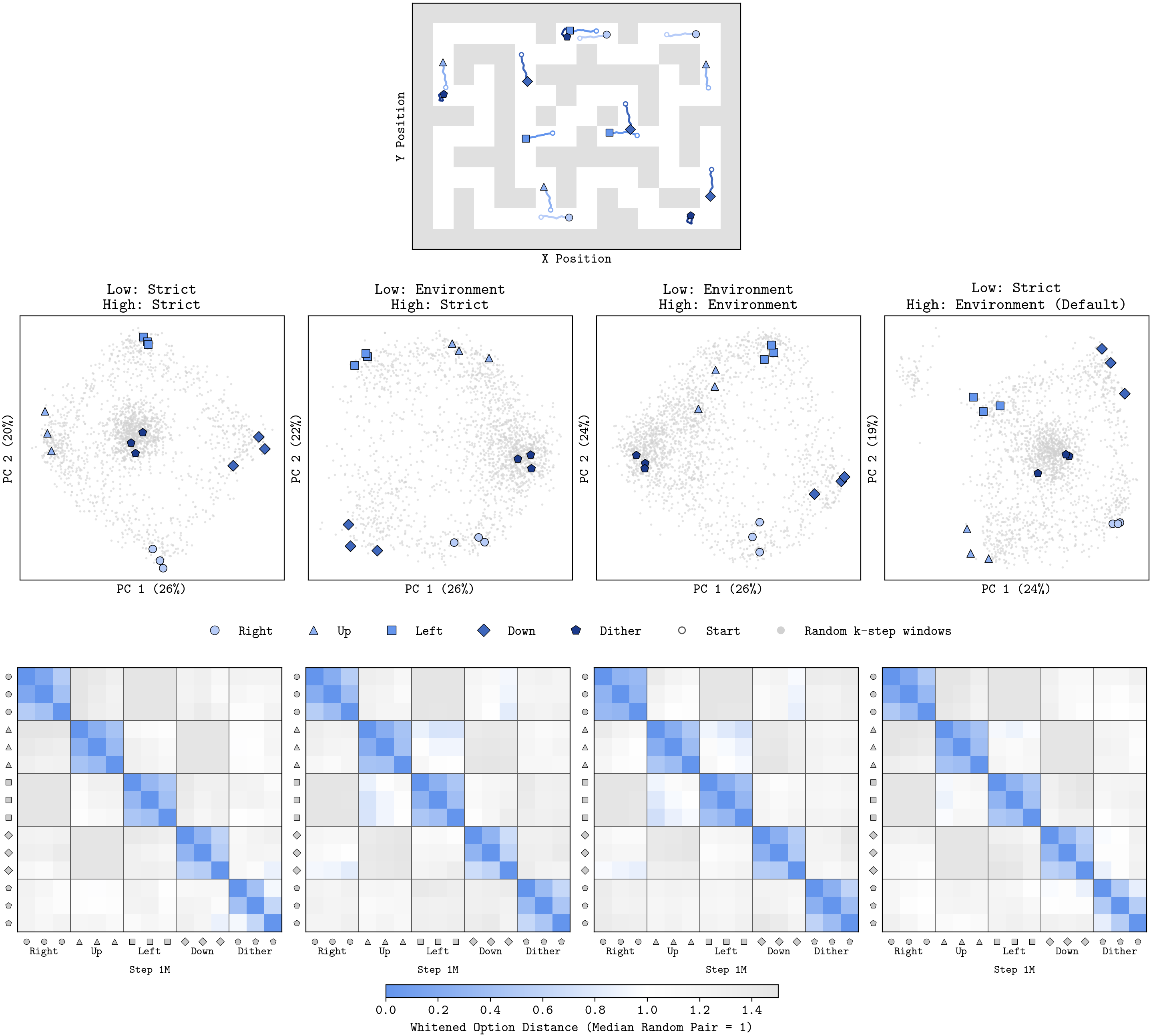}
    \caption{\textbf{PCA} (\textbf{top}) of QSO's action abstractions for reward labelling ablations for antmaze-giant fitted on random trajectory sequences, showing different groups of three k-step trajectory sequences with semantically similar behaviour (randomly selected from the validation dataset).  \textbf{L2 distances} (\textbf{bottom}) for those groups in units of median distance between two random trajectory sequences. All reward labelling seems to encourage grouping functionally equivalent behaviours while separating the behaviour groups.}
    \label{fig: ant_reward_option_embeddings}
\end{figure}

\begin{figure}[H]
    \centering
    \includegraphics[width=\linewidth]{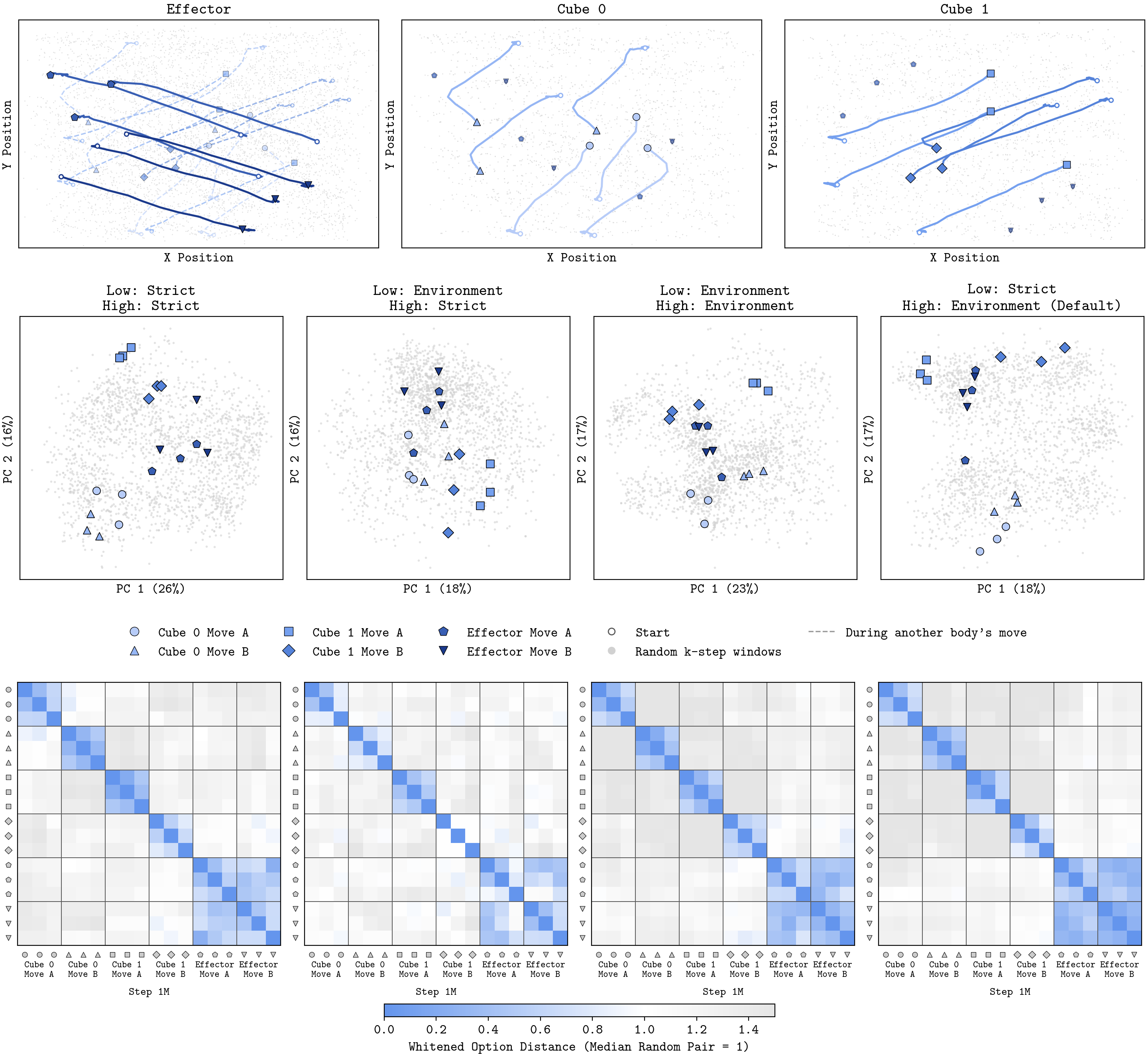}
    \caption{\textbf{PCA} (\textbf{top}) of QSO's action abstractions for reward labelling ablations for cube-double fitted on random trajectory sequences, showing different groups of three k-step trajectory sequences with semantically similar behaviour (randomly selected from the validation dataset).  \textbf{L2 distances} (\textbf{bottom}) for those groups in units of median distance between two random trajectory sequences. All reward labelling seems to encourage grouping functionally equivalent behaviours while separating the behaviour groups.}
    \label{fig: cube_reward_option_embeddings}
\end{figure}

For QSO, reward labelling also shapes the option space itself, as the option encoder is trained through the critics. Downstream performance is most sensitive to using high-level reward labelling, which, like for ASO, we attribute to the high-level critic. Interestingly, however, low-level environment labelling makes the encoder markedly less sensitive to fast-moving dimensions, dropping from 79\% to 8\% for cube-double and 45\% to 23\% for antmaze-giant; the high-level policy becomes correspondingly more dependent on slow-moving dimensions, respectively increasing from  5\% to 23\% and 14\% to 18\%. Hence, while the option space remains similarly clean under both variants (Figures \ref{fig: ant_reward_option_embeddings} and \ref{fig: cube_reward_option_embeddings}), using low-level environment reward labelling could be considered better, as it retains less information about fast-moving state dimensions without sacrificing downstream performance.
\begin{figure}[H]
    \centering
    \includegraphics[width=0.6\linewidth]{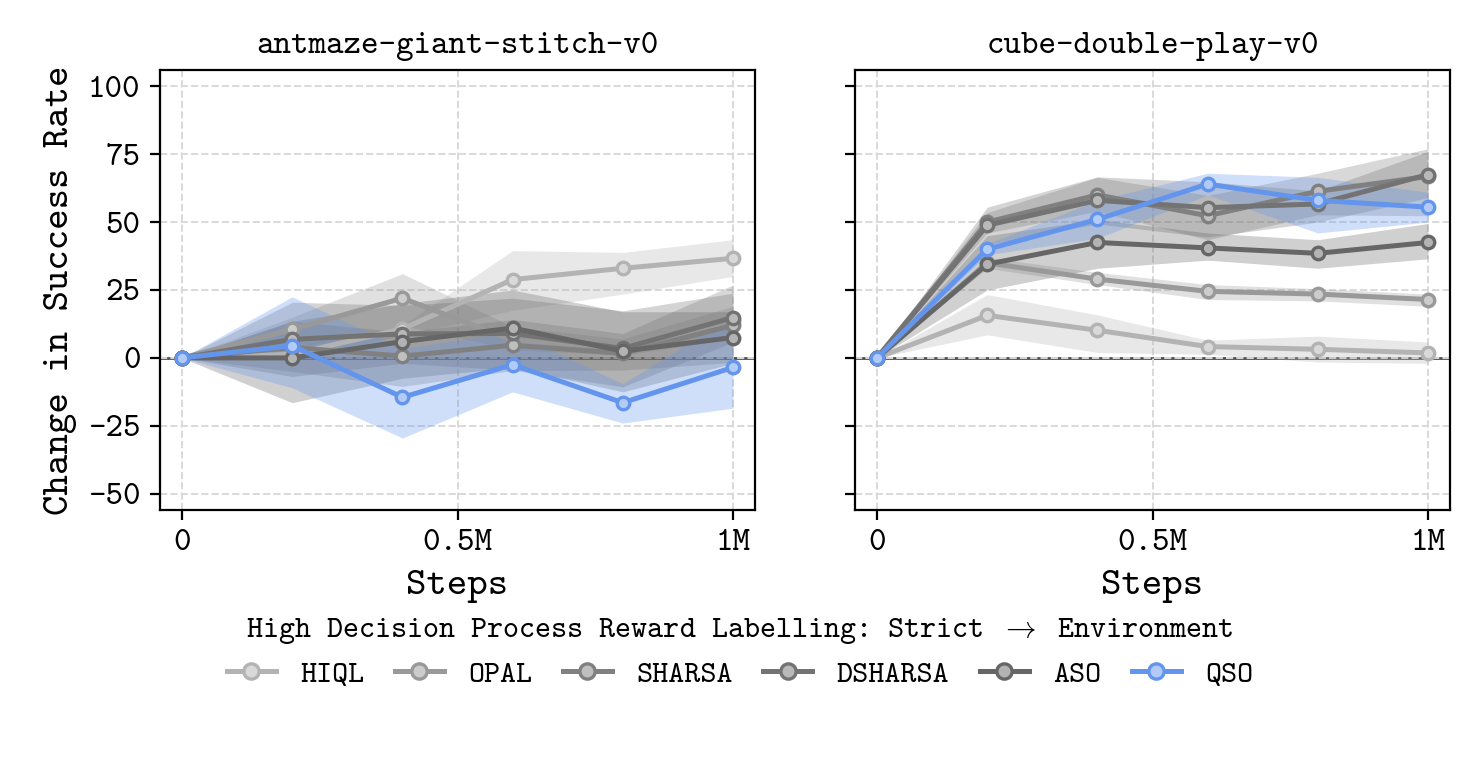}
    \caption{Change in performance when using the \textbf{environment success criterion rather than the stricter success criterion used by prior work} for reward labelling in the high-level decision process. Performance either does not change significantly, or improves for all algorithms.}
    \label{fig: ablation_reward_label_all}
\end{figure}
\begin{figure}[H]
    \centering
    \includegraphics[width=0.6\linewidth]{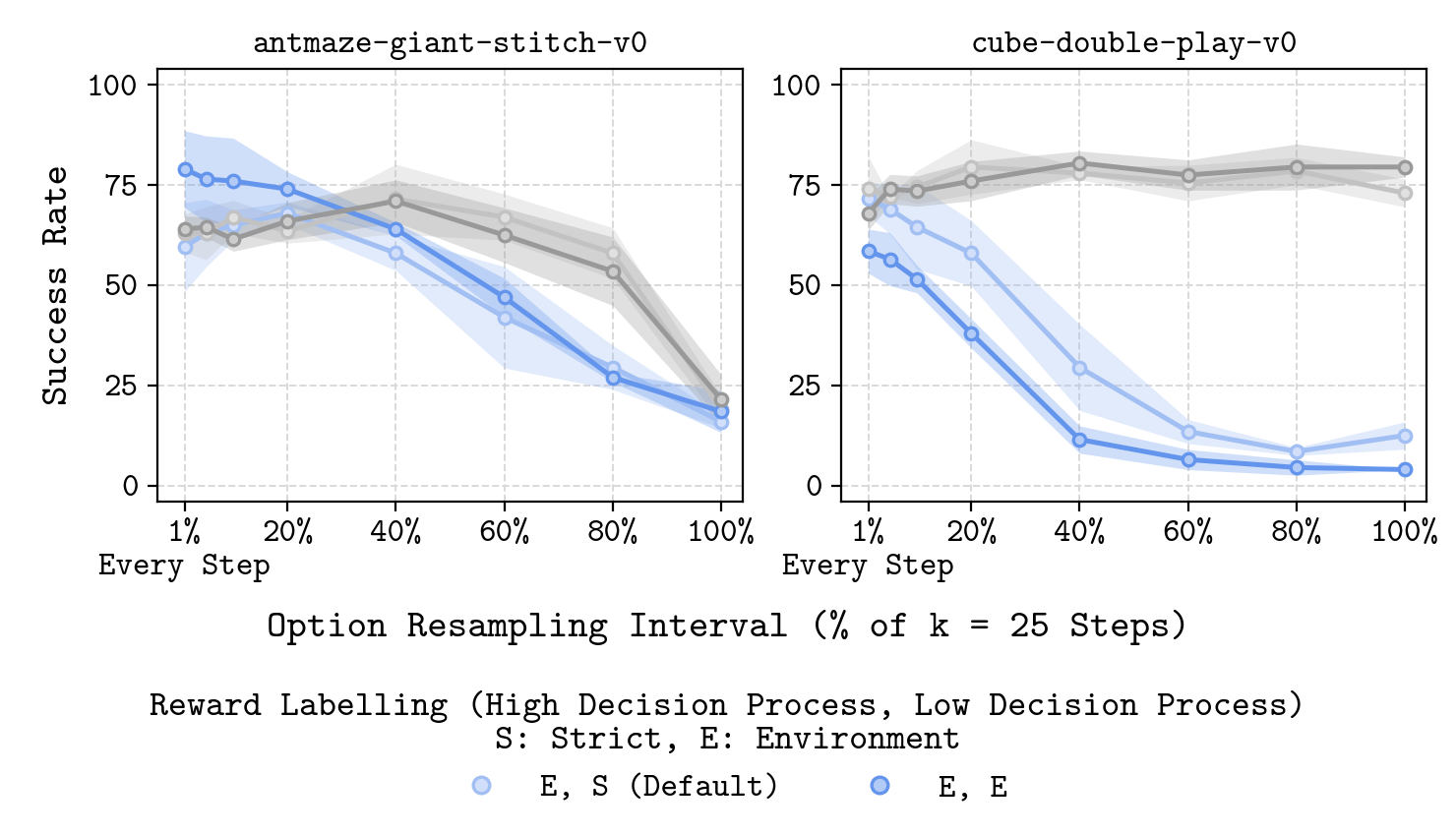}
    \caption{Performance with different \textbf{option resampling frequencies} for \textbf{low-level dataset reward labelling ablations} for QSO (\textbf{blue}) and ASO (\textbf{grey}).}
    \label{fig: option_resampling}
\end{figure}

\begin{figure}[H]
    \centering
    \includegraphics[width=0.6\linewidth]{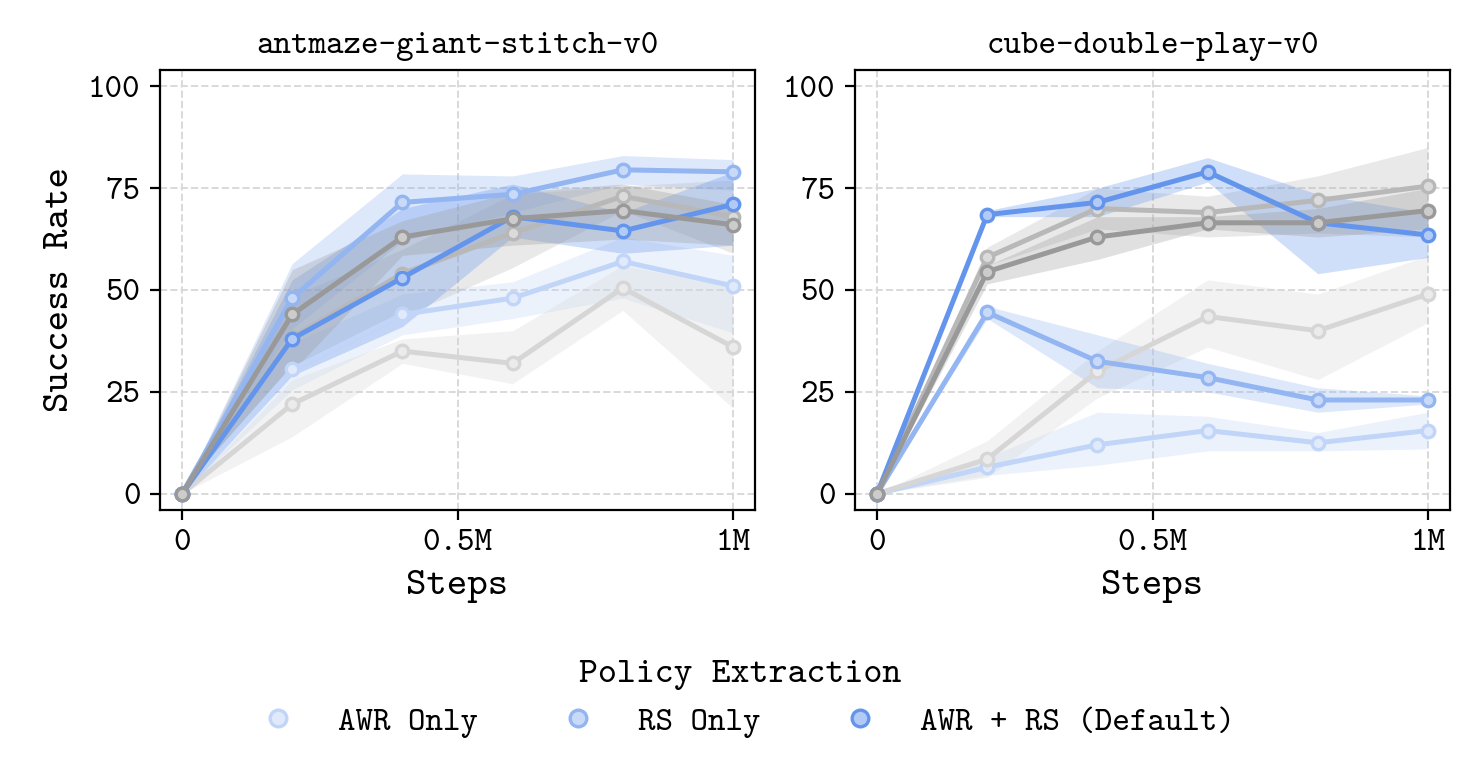}
    \caption{High-level \textbf{policy extraction ablation} for QSO (\textbf{blue}) and ASO (\textbf{grey}).}
    \label{fig: ablation_policy_extraction}
\end{figure}

\newpage
\textbf{How sensitive is QSO to policy extraction?} Both advantage weighting and RS lead to more consistent performance for ASO and QSO (Figure \ref{fig: ablation_policy_extraction}). As motivated, advantage weighting increases the high-level policy's dependence on the goal: compared with BC, the goal's share of output variance increases from 1\% to 14\% for QSO and 1\% to 5\% for ASO in antmaze-giant, and, respectively, from 0\% to 1\% and 1\% to 3\% in cube-double. Note that RS on top of this further increases goal-dependency (Table \ref{tab: goal_sensitivity}). 

\begin{table}[H]
\centering
\providecommand{\resultcell}[1]{\rule{0pt}{3.2ex}#1\rule[-1.4ex]{0pt}{0pt}}
\renewcommand{\arraystretch}{1.0}
\setlength{\tabcolsep}{4pt}
\resizebox{0.78\textwidth}{!}{\ttfamily
\begin{tabular}{ll>{\columncolor[HTML]{F2F2F2}}c>{\columncolor[HTML]{F2F2F2}}c>{\columncolor[HTML]{F2F2F2}}cccc}
\toprule
 & & \multicolumn{3}{c}{Flow Policy} & \multicolumn{3}{c}{Deployed Policy (best of 32)} \\
\cmidrule(lr){3-5} \cmidrule(lr){6-8}
Agent & Policy Extraction & State & Goal & Noise & State & Goal & Noise \\
\midrule
\multicolumn{8}{l}{\rule{0pt}{2.6ex}\textit{antmaze-giant-stitch-v0}} \\
\multirow{2}{*}{\textcolor[HTML]{6495ED}{QSO}} & BC & \resultcell{72 {\footnotesize [69, 74]}} & \resultcell{1 {\footnotesize [1, 2]}} & \resultcell{27 {\footnotesize [25, 30]}} & \resultcell{60 {\footnotesize [57, 62]}} & \resultcell{24 {\footnotesize [23, 26]}} & \resultcell{16 {\footnotesize [15, 17]}} \\
 & \textcolor[HTML]{6495ED}{AWR} & \resultcell{64 {\footnotesize [61, 67]}} & \resultcell{\textcolor[HTML]{6495ED}{\textbf{14}} {\footnotesize [13, 15]}} & \resultcell{22 {\footnotesize [21, 24]}} & \resultcell{54 {\footnotesize [52, 57]}} & \resultcell{\textcolor[HTML]{6495ED}{\textbf{34}} {\footnotesize [32, 36]}} & \resultcell{12 {\footnotesize [11, 12]}} \\
\cmidrule(lr){1-8}
SHARSA & BC & \resultcell{38 {\footnotesize [35, 41]}} & \resultcell{1 {\footnotesize [1, 2]}} & \resultcell{61 {\footnotesize [58, 64]}} & \resultcell{34 {\footnotesize [31, 36]}} & \resultcell{15 {\footnotesize [15, 16]}} & \resultcell{51 {\footnotesize [49, 53]}} \\
ASO & AWR & \resultcell{35 {\footnotesize [32, 38]}} & \resultcell{\textcolor[HTML]{6495ED}{\textbf{5}} {\footnotesize [5, 5]}} & \resultcell{60 {\footnotesize [57, 62]}} & \resultcell{33 {\footnotesize [31, 36]}} & \resultcell{\textcolor[HTML]{6495ED}{\textbf{19}} {\footnotesize [18, 20]}} & \resultcell{48 {\footnotesize [46, 50]}} \\
\midrule
\multicolumn{8}{l}{\rule{0pt}{2.6ex}\textit{cube-double-play-v0}} \\
\multirow{2}{*}{\textcolor[HTML]{6495ED}{QSO}} & BC & \resultcell{98 {\footnotesize [98, 98]}} & \resultcell{0 {\footnotesize [0, 0]}} & \resultcell{2 {\footnotesize [2, 2]}} & \resultcell{97 {\footnotesize [96, 97]}} & \resultcell{2 {\footnotesize [2, 2]}} & \resultcell{2 {\footnotesize [1, 2]}} \\
 & \textcolor[HTML]{6495ED}{AWR} & \resultcell{93 {\footnotesize [92, 93]}} & \resultcell{\textcolor[HTML]{6495ED}{\textbf{1}} {\footnotesize [1, 1]}} & \resultcell{6 {\footnotesize [6, 7]}} & \resultcell{88 {\footnotesize [88, 89]}} & \resultcell{\textcolor[HTML]{6495ED}{\textbf{6}} {\footnotesize [6, 7]}} & \resultcell{5 {\footnotesize [5, 6]}} \\
\cmidrule(lr){1-8}
SHARSA & BC & \resultcell{85 {\footnotesize [85, 86]}} & \resultcell{1 {\footnotesize [1, 1]}} & \resultcell{14 {\footnotesize [13, 15]}} & \resultcell{80 {\footnotesize [79, 81]}} & \resultcell{9 {\footnotesize [8, 9]}} & \resultcell{11 {\footnotesize [10, 13]}} \\
ASO & AWR & \resultcell{84 {\footnotesize [83, 85]}} & \resultcell{\textcolor[HTML]{6495ED}{\textbf{3}} {\footnotesize [3, 4]}} & \resultcell{13 {\footnotesize [12, 13]}} & \resultcell{79 {\footnotesize [78, 79]}} & \resultcell{\textcolor[HTML]{6495ED}{\textbf{12}} {\footnotesize [11, 12]}} & \resultcell{10 {\footnotesize [9, 10]}} \\
\bottomrule
\end{tabular}}
\caption{\textbf{QSO, SHARSA and ASO high-level goal sensitivity with AWR and BC policy extraction.} Percentage share of the variance of the high-level policy's options explained by the state and goal. Inputs are built on grids that combine 16 random validation states with 16 random validation goals, each under 8 flow-noise samples. For SHARSA and ASO, each dimension of the output is divided by the larger of its standard deviations over the dataset and over the sampled absolute subgoal options. The variance is attributed exactly using two-way ANOVA, so that the attributions sum to 100 (``Goal'' includes the state-goal interaction). The flow policy returns a single flow sample; the deployed policy returns the best of 32 flow samples under the high-level critic. Mean and 95\% bootstrap confidence interval (2000 resamples of the 64 grids) for one seed after 1M training steps.}
\label{tab: goal_sensitivity}
\end{table}

\begin{figure}[H]
    \centering
    \includegraphics[width=0.6\linewidth]{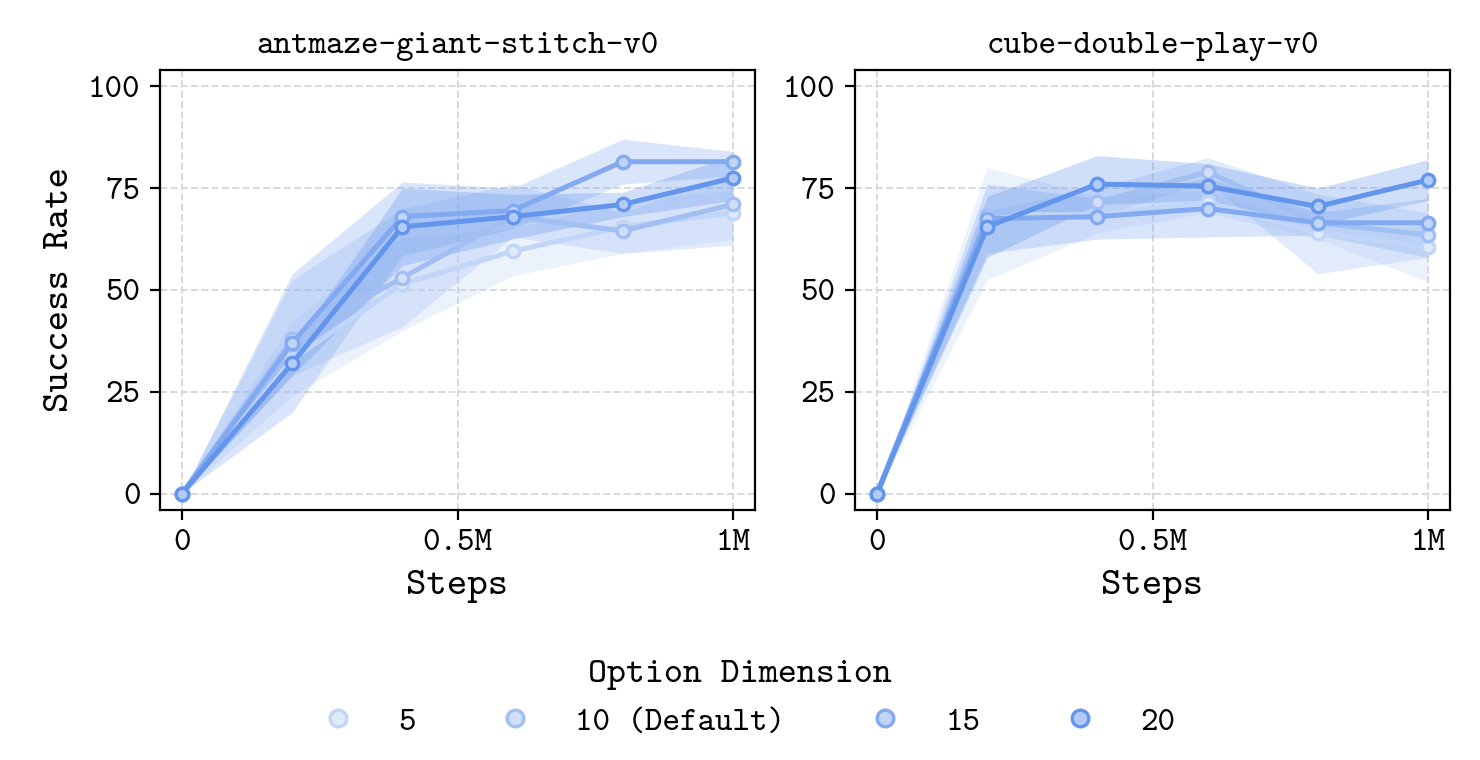}
    \caption{$d$ \textbf{option-space dimensionality ablation}.}
    \label{fig: ablation_rep_dim}
\end{figure}

\newpage
\textbf{How sensitive is QSO to option dimension?} QSO is largely insensitive to option dimension from 5 to 20, which all lead to similar downstream performance (Figure \ref{fig: ablation_rep_dim}). We use a fixed default of 10.

\begin{figure}[H]
    \centering
    \includegraphics[width=0.6\linewidth]{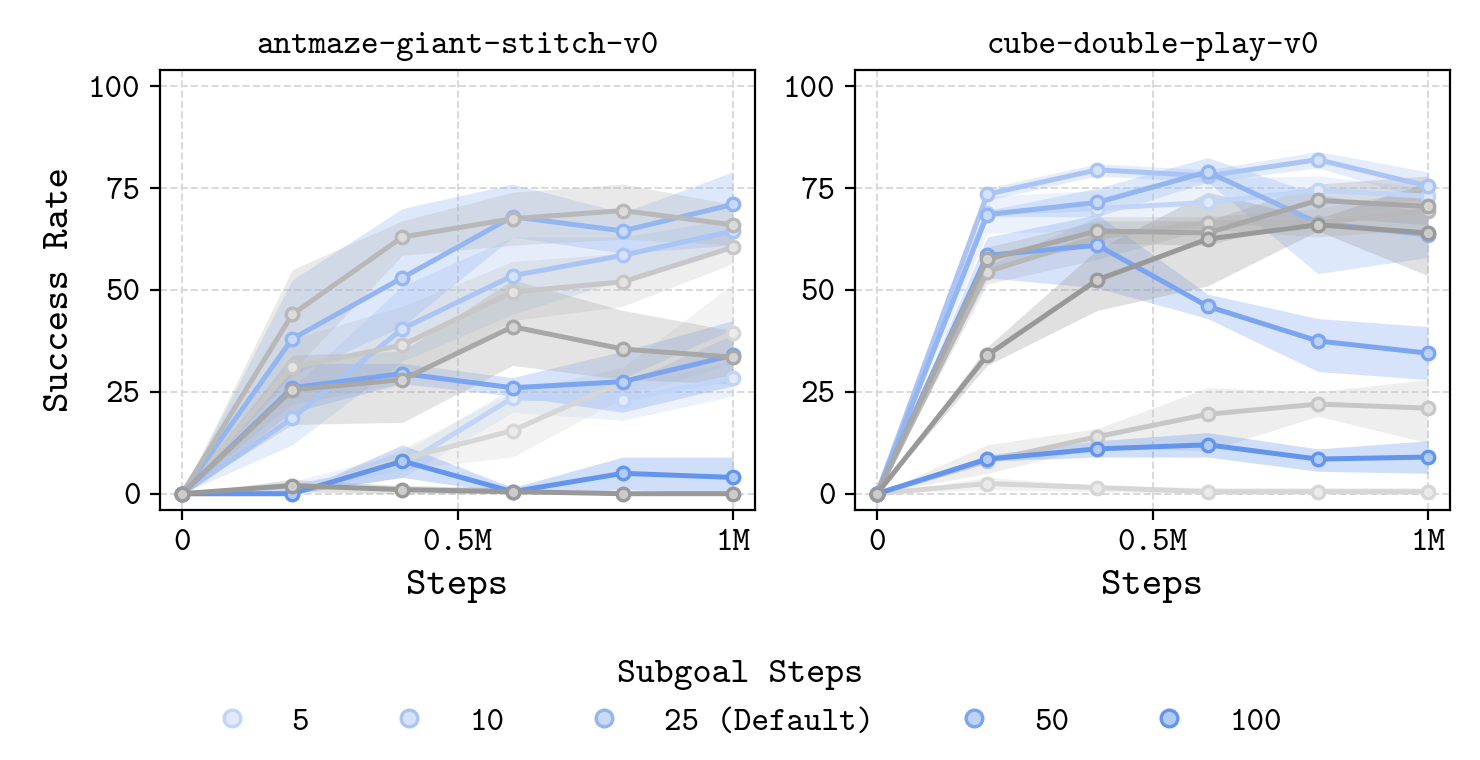}
    \caption{$k$ \textbf{temporal horizon ablation} for QSO (\textbf{blue}) and ASO (\textbf{grey}).}
    \label{fig: ablation_subgoal_steps}
\end{figure}

\textbf{How sensitive is QSO to temporal horizon?} Like ASO and prior work \citep{park2026horizon, ajay2020opal, park2023hiql, li2026reinforcement}, QSO is sensitive to temporal abstraction horizon (Figure \ref{fig: ablation_subgoal_steps}). For example, option horizons of 100 steps substantially degrades performance for both QSO and ASO in antmaze-giant; this is unsurprising, however, considering that the maximum number of steps in the antmaze-giant dataset is 200. 

\end{document}